\documentclass{article} %
\usepackage[table]{xcolor}
\usepackage{arxiv}

\usepackage{amsmath,amsfonts,bm}

\def\eqref#1{equation~\ref{#1}}

\def\1{\bm{1}}

\def\vtheta{{\bm{\theta}}}
\def\va{{\bm{a}}}

\def\vg{{\bm{g}}}

\DeclareMathAlphabet{\mathsfit}{\encodingdefault}{\sfdefault}{m}{sl}
\SetMathAlphabet{\mathsfit}{bold}{\encodingdefault}{\sfdefault}{bx}{n}

\usepackage{hyperref}
\usepackage{url}
\usepackage[pdftex]{graphicx}
\graphicspath{ {./images/} }

\usepackage{natbib}

\usepackage{todonotes}
\usepackage{multirow}
\usepackage{subcaption}
\usepackage{booktabs}
\usepackage{collcell}
\newcommand{\highlight}[1]{\colorbox{yellow}{$\displaystyle #1$}}
\newcommand{\highlighttwo}[1]{\colorbox{orange}{$\displaystyle #1$}}
 
\usepackage{enumitem}

\def\highlightifnumstyle{\bfseries}
\newcommand\highlightifnum[3]{%
  \ifdim#1#2#3%
    \highlightifnumstyle #3%
  \else%
    #3%
  \fi%
}

\newcolumntype{H}[3]{%
  >{%
    \def\wrapper{\highlightifnum{#1}{#2}}%
    \collectcell\wrapper%
  }%
  #3%
  <{\endcollectcell}%
}
\usepackage{authblk}

\title{Keep the Gradients Flowing: Using Gradient Flow to Study Sparse Network Optimization}

\author[1,3]{\textbf{Kale-ab Tessera}}
\author[2]{\textbf{Sara Hooker}}
\author[1]{\textbf{Benjamin Rosman}}

\affil[1]{School of Computer Science and Applied Mathematics, University of the Witwatersrand}
\affil[2]{Google Brain}
\affil[3]{InstaDeep}
\affil[ ]{\texttt{kaleabtessera@gmail.com} , \texttt{shooker@google.com} , \texttt{Benjamin.Rosman1@wits.ac.za}}

\begin{document}
\maketitle

\begin{abstract}
Training sparse networks to converge to the same performance as dense neural architectures has proved to be elusive. Recent work suggests that initialization is the key. However, while this direction of research has had some success, focusing on initialization alone appears to be inadequate. In this paper, we take a broader view of training sparse networks and consider the role of regularization, optimization, and architecture choices on sparse models. We propose a simple experimental framework, \textit{Same Capacity Sparse vs Dense Comparison} (\texttt{SC-SDC}), that allows for a fair comparison of sparse and dense networks. Furthermore, we propose a new measure of gradient flow, \textit{Effective Gradient Flow} (\texttt{EGF}), that better correlates to performance in sparse networks. Using top-line metrics, \texttt{SC-SDC} and \texttt{EGF}, we show that default choices of optimizers, activation functions and regularizers used for dense networks can disadvantage sparse networks. Based upon these findings, we show that gradient flow in sparse networks can be improved by reconsidering aspects of the architecture design and the training regime. Our work suggests that initialization is only one piece of the puzzle and taking a wider view of tailoring optimization to sparse networks yields promising results.  
\end{abstract}

\section{Introduction}
Over the last decade, a “bigger is better” race in the number of model parameters has gripped the field of machine learning \citep{2018Amodei,2020arXiv200705558T}, primarily driven by overparameterized deep neural networks (DNNs). Additional parameters improve top-line metrics, but drive up the cost of training \citep{2014Horowitz,strubelL2019energy} and increase the latency and memory footprint at inference time \citep{warden2019tinyml, Samala_2018, 8364435}. Moreover, overparameterized networks are more prone to memorization \citep{zhang2016understanding,2019shooker}.

To address some of these limitations, there has been a renewed focus on compression techniques that preserve top-line performance, while improving efficiency. A considerable amount of research focus has centred on pruning, where weights estimated to be unnecessary are removed from the network at the end of training \citep{louizos2017learning, wen2016learning, Cun90optimalbrain, 1993optimalbrain, Strom97sparseconnection, Hassibi93secondorder, zhu2017prune, 2016abigail, 2017Narang}. Pruning has shown a remarkable ability to preserve top-line performance metrics, even when removing the majority of weights \citep{2019arXiv190209574G,frankle2018the}. Furthermore, pruned networks also have faster training \citep{dettmers2019sparse,luo2017thinet} and inference \citep{2016Molchanov,luo2017thinet} times, while also being more robust to noise \citep{ahmad2019can}. Even with the substantial benefits of pruning, most pruning techniques still require training a large, overparameterized model \emph{before} pruning a subset of weights.

Due to the drawbacks of starting dense prior to introducing sparsity, there has been a recent focus on methods that allow networks that \emph{start} sparse at initialization, to converge to similar performance as dense networks \citep{frankle2018the,frankle2019stabilizing,liu2018rethinking}. These efforts have focused disproportionately on understanding the properties of initial sparse weight distributions that allow for convergence. However, while this work has had some success, focusing on initialization alone has proved to be inadequate \citep{2020FrankleMissingMark,evci2019difficulty}. Furthermore, certain aspects of sparse networks are poorly understood, such as their sensitivity to learning rates, notably higher ones \citep{frankle2018the,liu2018rethinking,frankle2019stabilizing}. This hints to optimization in sparse networks not being well understood, even though it appears to be critical to designing well-performing sparse networks. 

In this work, we take a broader view of why training sparse networks to converge to the same performance as dense networks has proved to be elusive. We reconsider many of the basic building blocks of the training process and ask whether they disadvantage sparse networks or not. Our work focuses on the behaviour of networks with random, fixed sparsity at initialization and we aim to gain further intuition into how these networks learn. Furthermore, we provide tooling tailored to the analysis of these networks. 

To study sparse network optimization in a controlled environment, we propose an experimental framework, \textit{Same Capacity Sparse vs Dense Comparison} (\texttt{SC-SDC}). Contrary to most prior work comparing sparse to dense networks, where overparameterized dense networks are compared to smaller sparse networks, \texttt{SC-SDC} compares sparse networks to their equivalent capacity dense networks (same number of active connections and depth). This ensures that the results are a direct result of sparse connections themselves and not due to having more or fewer weights (as is the case when comparing large, dense networks to smaller, sparse networks). 

Historically, exploding and vanishing gradients were a common problem in neural networks \citep{hochreiter2001gradient,hochreiter1991untersuchungen,bengio1994learning,glorot2010understanding,goodfellow2016deep}. Recent work has suggested that poor gradient flow is an exacerbated issue in sparse networks \citep{wang2020picking,evci2020gradient}. To this end, we go beyond simply comparing top-line metrics by also measuring the impact on gradient flow of each intervention. To accurately measure gradient flow in sparse networks, we propose a normalized measure of gradient flow, which we term \textit{Effective Gradient Flow} (\texttt{EGF}) -- this measure normalizes gradient flow by the number of active weights and is thus better suited to studying the training dynamics of sparse networks. We use this measure in conjunction with \texttt{SC-SDC} to see where sparse optimization fails, and consider where this failure could be due to poor gradient flow. 

\textbf{Contributions} Our contributions are enumerated as follows:

\begin{enumerate}[leftmargin=*,align=left]
  \item \textbf{New Tooling to Study Sparse Network Optimization}  We conduct large-scale experiments to evaluate the role of regularization, optimization and architecture choices on sparse models. We first introduce a new experimental framework --- \textit{Same Capacity Sparse vs Dense Comparison} (\texttt{SC-SDC}) --- that fairly compares sparse networks to their equivalent capacity dense networks (same number of active connections, depth, and weight initialization). We also propose a new measure of gradient flow, \textit{Effective Gradient Flow} (\texttt{EGF}), that we show to be a stronger predictor, in sparse networks, of top-line metrics such as accuracy and loss than current gradient flow formulations.
  \item \textbf{Batch Normalization Plays a Disproportionate Role in Stabilizing Sparse Networks} We show that batch normalization (BatchNorm) \citep{Ioffe2015} is statistically more critical for sparse networks than for dense networks, suggesting that gradient instability is a key obstacle to starting sparse.
  \item \textbf{Not All Optimizers and Regularizers Are Created Equal} We show that optimizers that use an exponentially weighted moving average (EWMA) to obtain an estimate of the variance of the gradient, such as Adam \citep{kingma2014adam} and RMSProp \citep{hinton2012neural}, are sensitive to higher gradient flow. This results in these methods, at times, having poor performance when used with $L2$ regularization or data augmentation.
  \item \textbf{Changing Activation Functions Can Benefit Sparse Networks} We benchmark a wide set of activation functions, specifically ReLU \citep{nair2010rectified} and non-sparse activation functions, such as PReLU \citep{he2015delving}, Swish \citep{ramachandran2017swish}, ELU \citep{clevert2015fast}, SReLU \citep{jin2015deep} and Sigmoid \citep{neal1992connectionist}. Our results show that Swish is a promising activation function when using adaptive optimization methods, while when using stochastic gradient descent (SGD), PReLU and Swish both show promise. We also show that due to Swish's non-monotonic formulation, it leads to better gradient flow, which helps performance in the sparse regime.
\end{enumerate}

\textbf{Implications} Our work is timely as sparse training dynamics are poorly understood. Most training algorithms and methods have been developed to suit training dense networks. Our work provides insight into the nature of sparse optimization. It suggests the need for a broader viewpoint beyond initialization to achieve better performing sparse networks. Our proposed approach provides a more accurate measurement of the training dynamics of sparse networks. This can be used to inform future work on the design of networks and optimization techniques that are tailored explicitly to sparsity.

\section{Methodology} \label{sec:methodology}
In this paper, we study sparse network optimization and measure which architecture and optimization choices favour sparse networks relative to dense networks. To this end, in the following sub-sections we introduce \textit{Same Capacity Sparse vs Dense Comparison} (\texttt{SC-SDC}) , a framework that we use to fairly compare sparse and dense networks (Section \ref{sec:sdsdc}) and \textit{Effective Gradient Flow} (\texttt{EGF}), a better measure of gradient flow in sparse networks to study network optimization (Section \ref{sec:Grad_Flow}). We use both \texttt{SC-SDC} and \texttt{EGF} in conjunction with test loss and accuracy  to study the behaviour of these networks and better understand their training dynamics (Section \ref{sec:results}). 

\subsection{Same Capacity Sparse vs Dense Comparison (\texttt{SC-SDC})} \label{sec:sdsdc}
\begin{figure}[ht]
    \begin{center}
    \includegraphics[width=\linewidth]{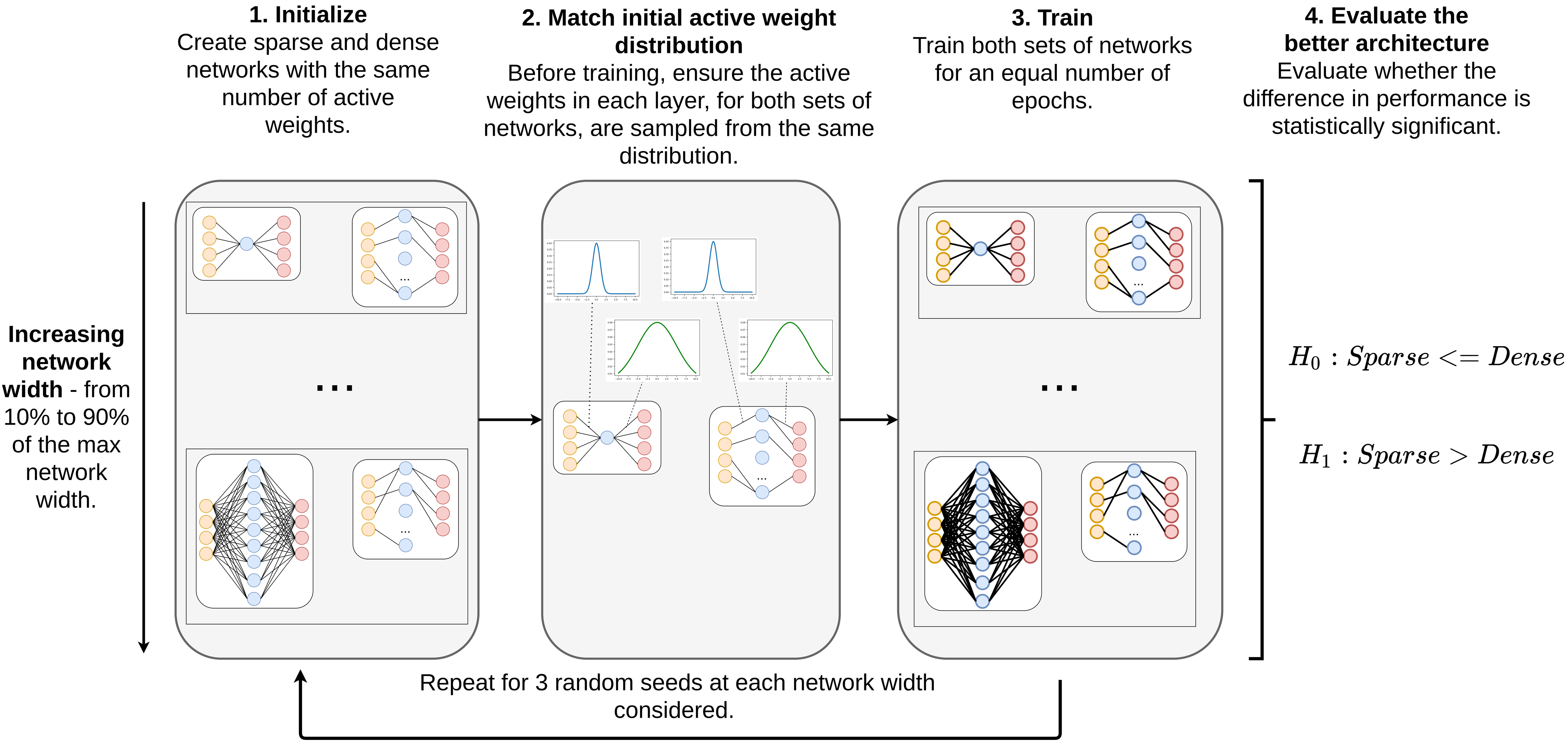}
    \end{center}
    \caption{\textbf{Same Capacity Sparse vs Dense Comparison (SC-SDC).} \texttt{SC-SDC} is a simple framework that fairly compares sparse and dense networks. This is done by ensuring that the compared sparse and dense networks have the same number of active (nonzero) weights in each layer, and that these active weights are initially sampled from the same distribution.}
    \label{fig:sdsdc}
\end{figure}

To fairly compare sparse and dense networks, we propose \textit{Same Capacity Sparse vs Dense Comparison} (\texttt{SC-SDC}), a simple framework that allows us to study sparse network optimization and identify which training configurations are not well suited for sparse networks. 

\texttt{SC-SDC} can be summarized as follows (see Figure \ref{fig:sdsdc} for an overview): 

\textbf{1. Initialize} For a chosen network depth (number of layers) $L$ and a maximum network width $N_{MW}$, we compare sparse network $S$ and dense network $D$ at various widths, while ensuring they have the same parameter count.
    Initially, we mask the weights $\vtheta_{S}$ of sparse network $S$:
    \begin{equation}
        \displaystyle \va_{S}^{l} = \vtheta_{S}^{l} \odot m^{l}  \quad , \quad  \va_{D}^{l} = \vtheta_{D}^{l}  , \quad \text { for } \quad l=1, \ldots, L  \quad,
    \end{equation}
where $\displaystyle \vtheta_{S}^{l} \odot m^{l}$ denotes an element-wise product of the weights $\vtheta_{S}$ of layer $l$ and the random binary matrix (mask) for layer $l$, $\displaystyle m^{l}$, $\displaystyle \va_{S}^{l}$ is the nonzero weights in layer $l$ of sparse network $S$, and $\displaystyle \va_{D}^{l}$ is the nonzero weights in layer $l$ of dense network $D$ (all the weights since no masking occurs).

For a fair comparison, we need to ensure the same number of nonzero weights for sparse network $S$ and dense network $D$ , across each layer $L$.

\begin{equation}\label{sdsdc:initialize}
     \displaystyle ||\va_{S}^{l}||_0  = ||\va_{D}^{l}||_0 , \quad \text { for } \quad l=1, \ldots, L
\end{equation}

We provide more implementation details of how we achieve this in Appendix \ref{sdsdc:implementation}.

\textbf{2. Match active weight distributions} Following prior work \citep{liu2018rethinking,2019arXiv190209574G}, we ensure that the nonzero weights at initialization of the sparse and dense networks are sampled from the same distribution at each layer as follows: 
\begin{equation}\label{sdsdc:match_weight_dist}
\displaystyle \va_{S}^{l} \sim P^{l} \quad,\quad  \va_{D}^{l} \sim P^{l} , \quad \text { for } \quad l=1, \ldots, L  \quad,
\end{equation}
where $P^l$ refers to the initial weight distribution at layer $l$. This ensures that both sets of active weights (sparse and dense) are initially sampled from the same distribution.

\textbf{3. Train} We then train the sparse and dense networks for the same number of epochs,  allowing for convergence ($500$ epochs for Fashion-MNIST, $1000$ epochs for CIFAR-10 and CIFAR-100). 

\textbf{4. Evaluate the better architecture}  We gather the results across the different widths and conduct a paired, one-tail Wilcoxon signed-rank test \citep{wilcoxon1945individual} to evaluate the better architecture. We have a sample size of $30$ networks for each test configuration. This comprises of five different network widths $\times$ three independent training runs (different random seeds) $\times$ two architectures (sparse and dense variant).

Our null hypothesis ($H_0$) is that sparse networks have similar or worse test accuracy than dense networks (lower or the same median), while our alternative hypothesis ($H_1$) is that sparse networks have better test accuracy performance than dense networks of the same capacity (higher median). This can be formulated as:
    \begin{equation}\label{txt:wil_test}
        H_0: Sparse <= Dense \quad , \quad  H_1: Sparse >  Dense
    \end{equation}

\subsection{Measuring Gradient Flow}\label{sec:Grad_Flow}

Gradient flow (GF) is used to study optimization dynamics and is typically approximated by taking the norm of the gradients of the network \citep{pascanu2013difficulty,nocedaL2002behavior,chen2018gradnorm,wang2020picking,evci2020gradient}. Looking at gradient flow is particularly relevant in the context of sparse networks, since they are sensitive to poor gradient flow \citep{wang2020picking,evci2020gradient}. Hence this would be a useful analysis tool for their optimization.

We consider a feedforward neural network $\displaystyle f: \mathbb{R}^{D} \rightarrow \mathbb{R}$, with function inputs $\displaystyle \mathbf{x} \in \mathbb{R}^{D}$ and network weights $\vtheta$. The gradient norm is usually computed by concatenating all the gradients of a network into a single vector, 
   $\vg = \frac{\partial \mathcal{C}}{\partial \vtheta}$ , 
where $C$ is our cost function. Then the vector norm is taken as follows: 
\begin{equation} \label{eq:g}
    gf_p = || \vg ||_p \quad ,
\end{equation}
where $p$ denotes the pth-norm and $gf_p$ is the gradient flow. Traditional gradient flow measures take the $L1$ or $L2$ norm of all the gradients \citep{chen2018gradnorm,pascanu2013difficulty,evci2020gradient}. Unless gradients are masked before calculating the gradient flow, the standard gradient flow formulation could result in gradients of masked weights, which do not influence the forward pass, being included in the formulation. Furthermore, computing the $L1$ or $L2$ gradient norm by concatenating all the gradients into a single vector gives disproportionate influence to layers with more weights.

\textbf{\textbf{Effective Gradient Flow}} To address these shortcomings with the standard $gf_p$  measures, we propose a simple modification of Equation \ref{eq:g}, which we term \textit{Effective Gradient Flow} (\texttt{EGF}), that computes the average, masked gradient (only gradients of active weights) norm across all layers. 

We calculate \texttt{EGF} as follows:
\begin{gather}  \label{eq:EGF}
  \vg =  (\frac{\partial \mathcal{C}}{\partial \vtheta^{1}}\odot m^{1},\frac{\partial \mathcal{C}}{\partial \vtheta^{2}}\odot m^{2}, \dots, \frac{\partial \mathcal{C}}{\partial \vtheta^{L}}\odot m^{L} ) \quad,  \\
  EGF_p = \dfrac{\sum\limits_{n=1}^{L}{||\vg_n ||_p}}{L} \quad ,
\end{gather}
where $L$ is the number of layers. For every layer $l$, $\displaystyle \frac{\partial \mathcal{C}}{\partial \theta^{l}} \odot m^{l}$ denotes an element-wise product of the gradients of layer $l$, $\displaystyle \frac{\partial \mathcal{C}}{\partial \theta^{l}}$ , and the mask $\displaystyle m^{l}$ applied to the weights of layer $l$. For a fully dense network, $m^{l}$ is a matrix of all ones, since no gradients are masked.

\textbf{EGF has the following favourable properties:}
\begin{itemize}[leftmargin=*,align=left]
  \item \textbf{Gradient Flow Is Evenly Distributed Across Layers} \texttt{EGF} distributes the gradient norm across the layers equally. This prevents layers with many weights from dominating the measure, and prevents layers with vanishing gradients from being hidden in the formulation, as is the case with equation \ref{eq:g} (when all gradients are appended together).
  \item \textbf{Only Gradients of Active Weights Are Used} \texttt{EGF}  ensures that for sparse networks, only gradients of active weights are used. Even though weights are masked, their gradients are not necessarily zero since the partial derivative of the weight w.r.t. the loss is influenced by other weights and activations. Therefore, even though the weight is zero, its gradient can be nonzero.
  \item \textbf{Possibility for Application in Gradient-based Pruning Methods} \cite{tanaka2020pruning} showed that gradient-based pruning methods like GRASP \citep{wang2020picking} and SNIP \citep{lee2018snip}, disproportionately prune large layers and are susceptible to layer-collapse, which is when an algorithm prunes all the weights in a specific layer. Since \texttt{EGF} is evenly distributed across layers, maintaining \texttt{EGF} (as opposed to standard gradient norm) could be used as pruning criteria. Furthermore, current approaches measuring or approximating the change in gradient flow during pruning in sparse networks \citep{wang2020picking,evci2020gradient,lubana2021a}, could benefit from this new formulation.

\end{itemize}

To evaluate \texttt{EGF} against other standard gradient norm measures, such as the $L1$ and $L2$ norm, we empirically compare these measures and their correlation to test loss and accuracy.  We take the average of the absolute Kendall Rank correlation \citep{kendall1938new}, across the different experiment configurations. We follow a similiar approach to \cite{jiang2019fantastic}, but unlike their work which has focused on correlating network complexity measures to the generalization gap, we measure the correlation of gradient flow to performance (accuracy and loss). We measure gradient flow at points evenly spaced throughout training.

Our results from Table \ref{tbl:gradflow_corr_gen} show that in sparse networks, \texttt{EGF} consistently has a higher average absolute correlation to both test loss and accuracy. We see that \texttt{EGF} has similar correlation to standard gradient flow measures in dense networks. This shows that the benefits of using \texttt{EGF} are more apparent when used in conjunction with sparse networks, since \texttt{EGF} only considers the gradients of nonzero weights. Due to the comparative benefits of \texttt{EGF} in sparse networks, we use it for the remainder of the paper to measure the impact of interventions. We include experimental results using other gradient flow measures in Appendix \ref{sec:grad_flow_Appendix} for completeness. 

\begin{table}[t]
\small
\centering
\resizebox{0.7\textwidth}{!}{
 \begin{tabular}{cllllll}
\toprule
&\multicolumn{1}{c}{\bf Measure}  &\multicolumn{2}{c}{\bf Sparse } &\multicolumn{2}{c}{\bf Dense} \\
& &      Test Loss & Test Accuracy & Test Loss & Test Accuracy \\
\midrule
\multirow{4}{*}{{\rotatebox[origin=c]{90}{FMNIST}}}
& $|| \vg ||_1$ & 0.355          & 0.316          & \textbf{0.365} & \textbf{0.354} \\
& $|| \vg ||_2$ & 0.282          & 0.292          & 0.285          & 0.329          \\
& $EGF_1$   &    \textbf{0.419} & \textbf{0.373} & \textbf{0.365} & \textbf{0.354} \\
& $EGF_2$ & 0.360          & 0.323          & 0.298          & 0.320          \\
\midrule
 \multirow{4}{*}{{\rotatebox[origin=c]{90}{CIFAR-10}}} 
 & $|| \vg ||_1$ &    0.440          & 0.327          & \textbf{0.380} & 0.251          \\
 & $|| \vg ||_2$ &    0.447          & 0.308          & 0.355          & \textbf{0.290} \\
 & $EGF_1$       &    0.371          & 0.300          & \textbf{0.380} & 0.252          \\
 & $EGF_2$       &   \textbf{0.451} & \textbf{0.332} & 0.363          & 0.287          \\
 \midrule
\multirow{5}{*}{{\rotatebox[origin=c]{90}{CIFAR-100}}} 
 & $|| \vg ||_1$ &    0.355          & 0.385          & 0.325          & 0.319          \\
 & $|| \vg ||_2$ &    0.373          & 0.393          & 0.357          & \textbf{0.385} \\
 & $EGF_1$       &    0.358          & 0.320          & 0.325          & 0.319          \\
 & $EGF_2$       &    \textbf{0.402} & \textbf{0.396} & \textbf{0.359} & 0.382          \\ \\
 \bottomrule
\end{tabular}
}
\vspace{1ex}
 \caption{\textbf{The Average Correlation Between Gradient Flow Measures and Generalization Performance.} We compare the average, absolute Kendall Rank correlation \citep{kendall1938new} between different formulations of gradient flow and generalization (test loss and test accuracy). The subscripts ($1$ or $2$) denote the $p$-norm ($l1$ or $l2$ norm).  We see that for sparse networks, our proposed measure, \texttt{EGF} (Equation \ref{eq:EGF}), consistently has higher absolute correlation to performance compared to standard gradient flow measures ($|| \vg ||_1$ and $|| \vg ||_2$, Equation \ref{eq:g}). For each dataset, we highlight the measure with the highest correlation to performance in bold. 
 }
 \label{tbl:gradflow_corr_gen}
\end{table}

\begin{table}[hb]
\small
\centering
\resizebox{0.8\linewidth}{!}{
\begin{tabular}{ | m{4.2cm} | m{7.5cm} | }
\multicolumn{1}{c}{\bf Configuration}  &\multicolumn{1}{c}{\bf Variants} \\ 
\midrule 
Optimizers                          & Adagrad, Adam, RMSProp, SGD and SGD with mom (0.9).                                                         \\ \hline   Regularization/Normalization & No Regularization (NR), Weight Decay (L2), Data Augmentation (DA), Skip Connections (SC) and BatchNorm (BN). \\   \hline 
Number of hidden layers             & 1, 2  and 4.                                                                                               \\   \hline 
Dense Width                         & 308, 923, 1538, 2153 and 2768.                                                                              \\  \hline 
Activation functions                & ReLU, PReLU, ELU, Swish, SReLU and Sigmoid.                                                                 \\  \hline 
Learning rate                       & 0.001 and 0.1.                                                                                              \\   \hline 
Datasets                            & Fashion-MNIST, CIFAR-10 and CIFAR-100.  \\           \bottomrule                                                        
\end{tabular}
}
\vspace{1ex}
\caption{\textbf{Network Configurations.} Different network configurations for Sparse and Dense Comparisons.}
\label{tbl:key_variants}
\end{table}

\section{Empirical Setting} 
\subsection{Average \texttt{EGF}}
To measure gradient flow, we use the Average \texttt{EGF} calculated at the end of 11 epochs, evenly spread throughout the training. This is done because it would be computationally costly to calculate gradient norms at the end of every epoch.  

\subsection{Architecture, Normalization, Regularization and Optimizer Variants}

We briefly describe our key experiment variants below and also include for completeness all unique variants in Table \ref{tbl:key_variants}. 

\begin{enumerate}%
\item \textbf{Activation Functions} ReLU networks \citep{nair2010rectified} are known to be more resilient to vanishing gradients than networks that use Sigmoid or Tanh activations, since they only result in vanishing gradients when the input is less than zero, while on active paths, due to ReLU's linearity, the gradients flow uninhibited \citep{glorot2011deep}. 
Although most experiments are run on ReLU networks, we also explore different activation functions, namely  PReLU \citep{he2015delving}, ELU \citep{clevert2015fast}, Swish \citep{ramachandran2017swish}, SReLU \citep{jin2015deep} and Sigmoid \citep{neal1992connectionist}.

\item \textbf{Batch Normalization and Skip Connections} We empirically explore the relative benefits of BatchNorm \citep{Ioffe2015} and skip connections \citep{srivastava2015highway,he2016deep} across dense and sparse networks.

\item \textbf{Regularization Techniques} We evaluate various popular regularization methods: weight decay/$L2$ regularization ($\lambda$ = 0.0001) \citep{krogh1992simple,hanson1989comparing} and data augmentation (random crops and random horizontal flipping \citep{krizhevsky2012imagenet}).

\item \textbf{Optimization Techniques} We benchmark the impact of the most widely used optimizers such as minibatch stochastic gradient descent (SGD) \citep{robbins1951stochastic}, minibatch stochastic gradient descent with momentum (momentum term - $0.9$)  \citep{sutskever2013importance,polyak1964some}, Adam \citep{kingma2014adam}, Adagrad \citep{duchi2011adaptive} and RMSProp \citep{hinton2012neural}. 
\end{enumerate}

\subsection{SC-SDC MLP Setting} 

We firstly use the \texttt{SC-SDC} empirical setting (Section \ref{sec:sdsdc}) to evaluate which choices of optimizer, regularization and architecture choices result in a statistically significant performance difference between sparse and dense networks. We train $600$ MLPs for 500 epochs on Fashion-MNIST (FMNIST) \citep{xiao2017fashion} and over $10\,000$ MLPs for 1000 epochs on CIFAR-10 and CIFAR-100 \citep{krizhevsky2009cifar}. We compare sparse and dense networks across various widths, depths, learning rates, regularization and optimization methods as shown in Table \ref{tbl:key_variants}. 

\textbf{Dense Width}\label{subsec:dense_width} Following from \texttt{SC-SDC}, these networks are compared at various network widths, specifically a width of  308, 923, 1538, 2153, 2768 (10\%, 30\%, 50\%, 70\% and 90\%  of our maximum width $N_{MW}$(3076) when using CIFAR datasets) as shown in Table \ref{tbl:key_variants}. We use the term \textbf{dense width} to refer to the width of a network if that network was dense. For example, when comparing sparse and dense networks at a Dense Width of 308, this means the dense network has a width of 308, while the sparse network has a width of  $N_{MW}$ (3076), but has the same number of active connections as its dense counterpart. We provide more details on Dense Width and other components of the \texttt{SC-SDC} implementation in Appendix \ref{sdsdc:implementation}.

\subsection{Extended CNN Setting} We also extend our experiments to CNNs and train $200$ Wide ResNet-50 (WRN) models, specifically the WRN-28-10 variant \citep{Zagoruyko2016} for $200$ epochs on CIFAR-100. We use the same network training parameters used in the original paper (an initial learning rate of $0.1$, dropped by $0.2$ at epochs 60,120 and 160, weight decay $\lambda$ set to $0.0005$ and a momentum term of $0.9$).

\section{Results and Discussion} \label{sec:results}
We present our results studying sparse network optimization in MLPs, using  \texttt{SC-SDC} and \texttt{EGF}. Furthermore, we extend our results outside of \texttt{SC-SDC}, from MLPs to CNNs and from Random Pruning to Magnitude Pruning. 

For brevity, we use a shorthand notation for our different regularization/normalization methods --- no regularization (\textit{NR}), weight decay (\textit{L2}), data augmentation (\textit{DA}),  skip connections (\textit{SC})  and  BatchNorm (\textit{BN}). When multiple of these methods are combined, we chain their abbreviations such as \textit{DA\_BN}, which represents data augmentation and BatchNorm combined. Furthermore, for our discussion, we make a distinction between optimization methods that use exponential weighted moving averages, also known as leaky averaging, for estimates of the variance of their gradients (\textit{EWMA optimizers}) - specifically Adam and RMSProp, and methods which do not - Adagrad, SGD and SGD with momentum. We provide details on this distinction in Appendix \ref{app_sec:adap}.

\subsection{Comparison of Dense and Sparse Interventions Using SC-SDC} \label{sec:sdsdc_results}
In this section, we use the results from \texttt{SC-SDC} to identify which optimization choices are currently well suited for sparse networks and which are not. Most of the results discussed are achieved using four hidden layers on CIFAR-100, while we provide the full set of results for Fashion-MNIST, CIFAR-10 and CIFAR-100 in Appendix \ref{appen:detailed_sdsdc}.  

\subsubsection{Batch Normalization plays a disproportionate role in stabilizing Sparse Networks} BatchNorm ensures that the distribution of the nonlinearity inputs remains stable as the network trains, which was hypothesized to help stabilize gradient propagation (gradients do not explode or vanish) \citep{Ioffe2015}. 

\textbf{In Non-EWMA Optimizers, BatchNorm Favours Sparse Networks}\label{subsubsec:B1}
From Table \ref{tbl:c100_reg_low_lr} and \ref{tbl:c100_reg_high_lr}, we see that in non-EWMA optimizers (Adagrad, SGD and SGD with momentum), BatchNorm is statistically more significant for sparse network performance than it is for dense networks. Without BatchNorm (configurations \textit{NR}, \textit{DA}, \textit{L2} and \textit{SC}), we see that sparse networks do not outperform their dense counterparts in test accuracy. With the addition of BatchNorm, across high and low learning rates and at most configurations (except in some cases where $L2$ is used), there is strong evidence that sparse networks outperform dense networks.

\begin{figure}[ht]
  \centering
  \begin{subfigure}{0.475\textwidth} \centering
    \caption{Test Accuracy - Low learning rate (0.001)}\label{fig:c100_low_lr_acc}
    \includegraphics[width=\textwidth]{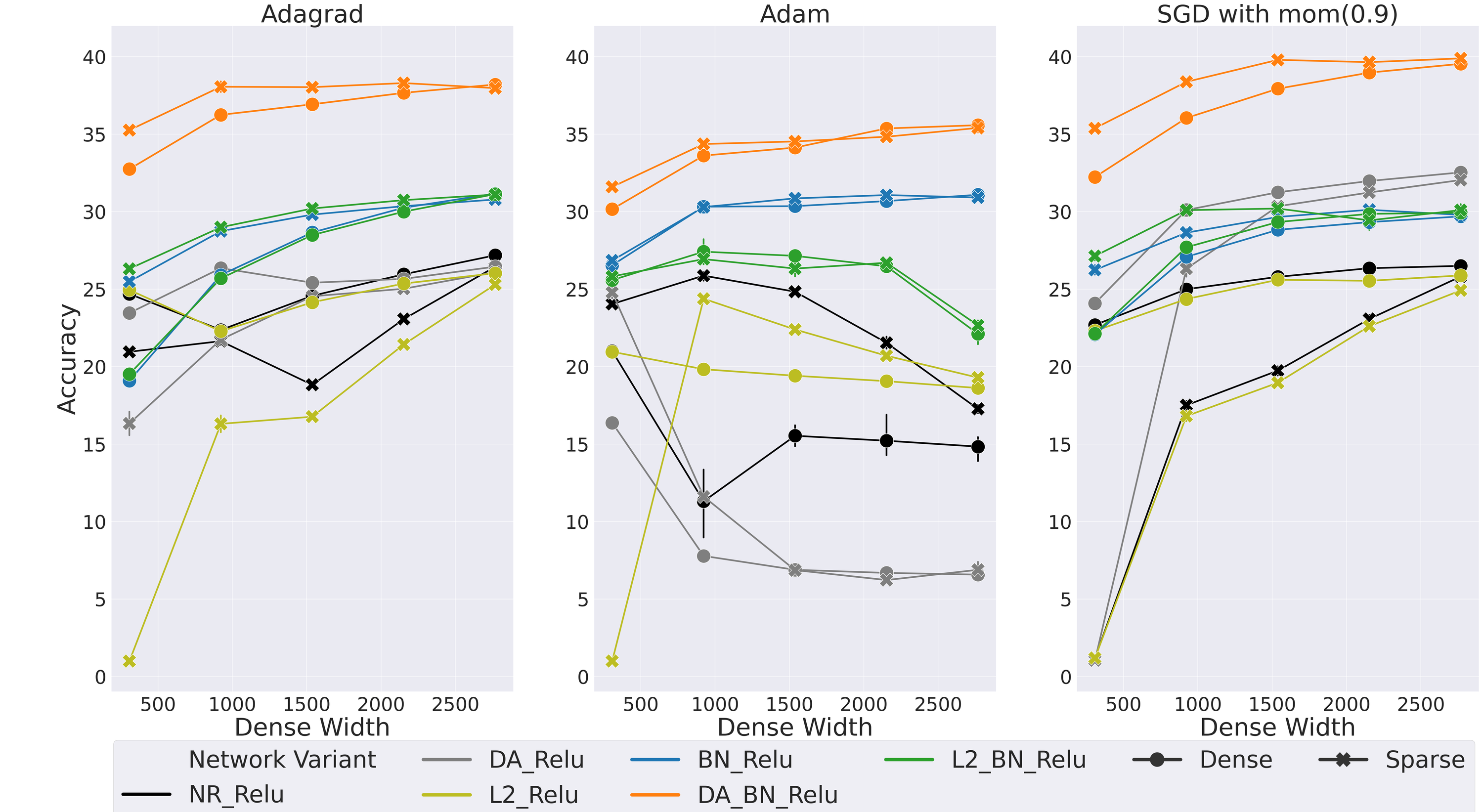}
  \end{subfigure}
  \hfill
  \begin{subfigure}{0.475\textwidth} \centering
    \caption{Gradient Flow - Low learning rate (0.001)}\label{fig:c100_low_lr_gf}
    \includegraphics[width=\textwidth]{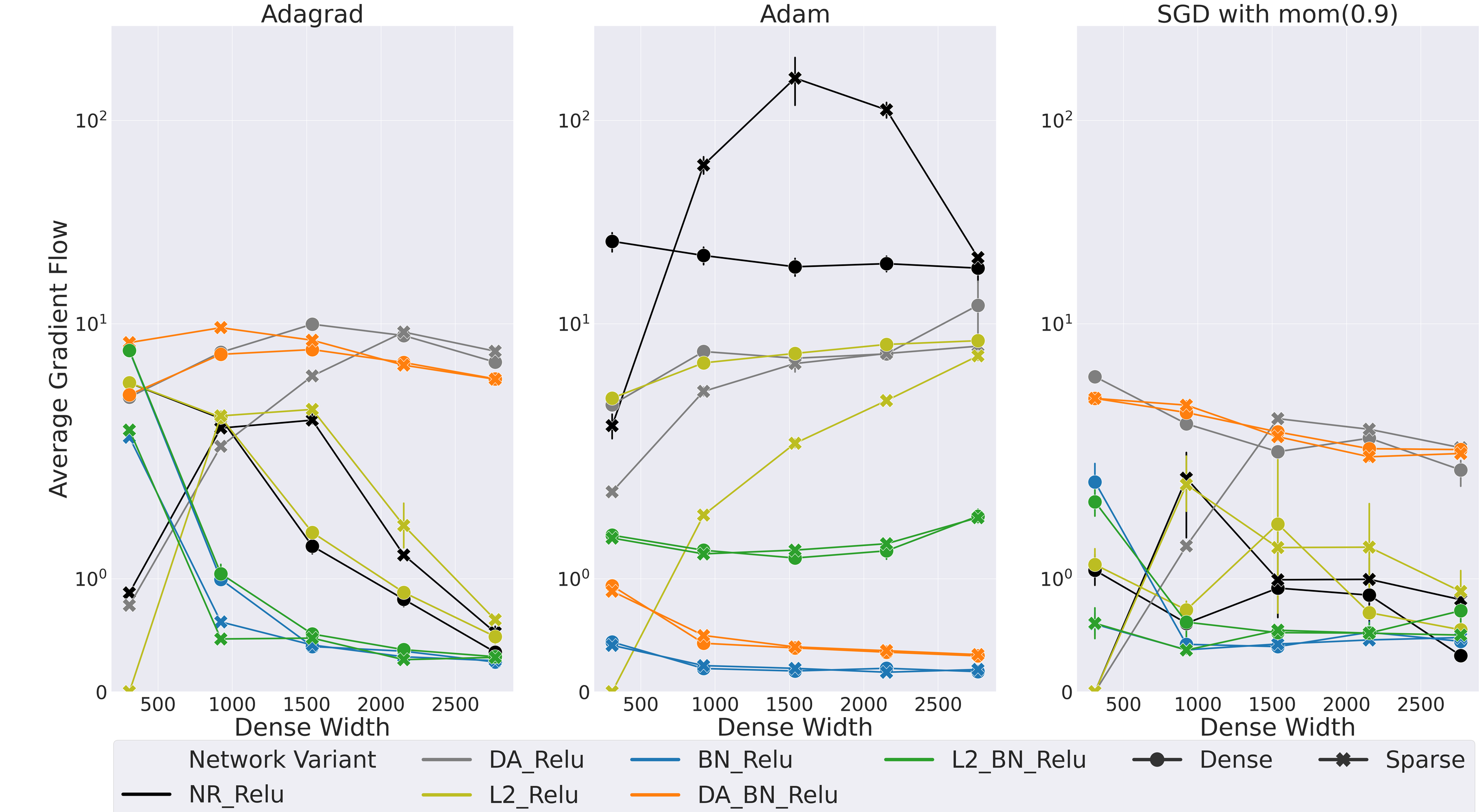}
  \end{subfigure}
  \vskip\baselineskip
  \begin{subfigure}{0.475\textwidth} \centering
    \caption{Test Accuracy - High learning rate (0.1)}\label{fig:c100_high_lr_acc}
    \includegraphics[width=\textwidth]{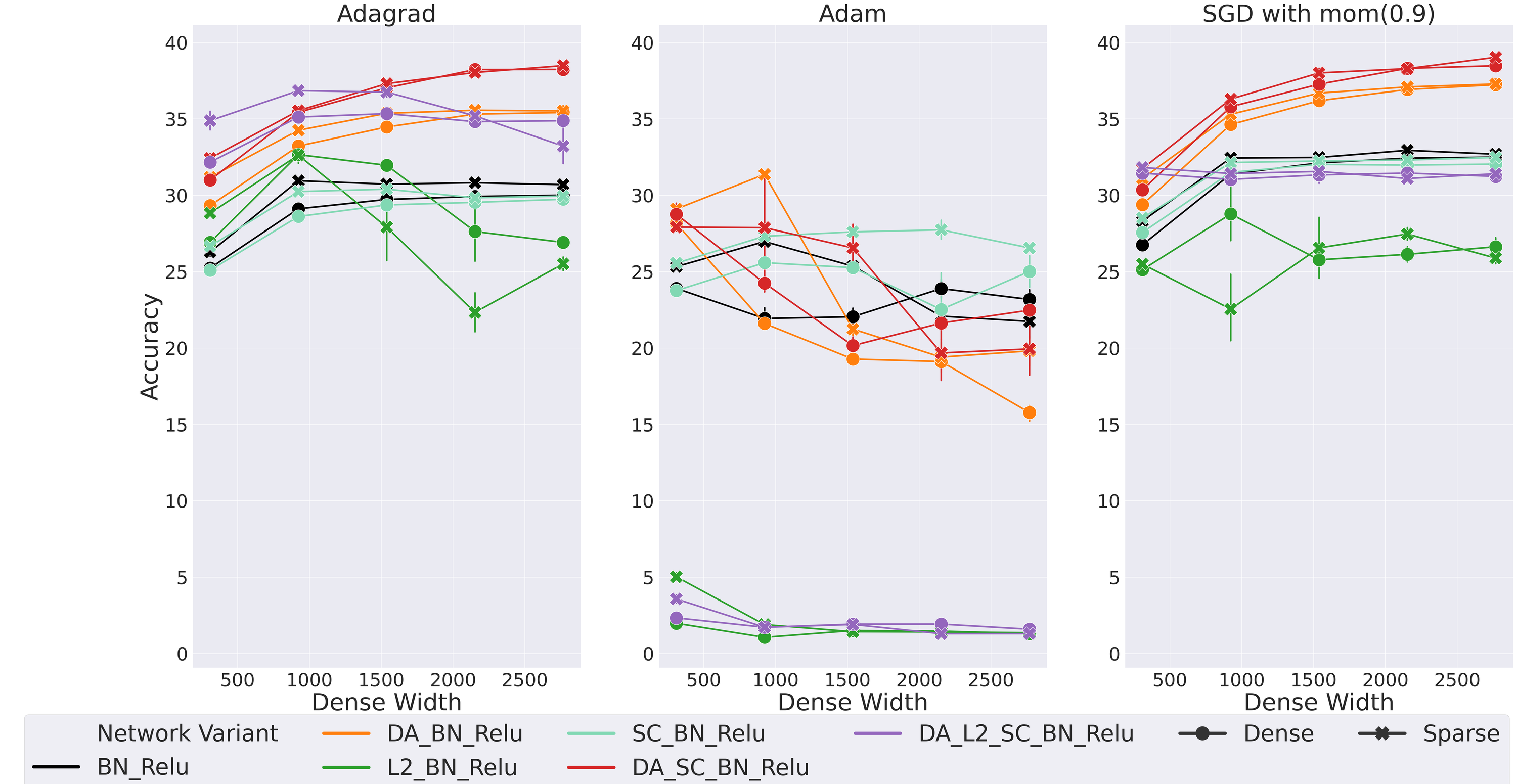}
  \end{subfigure}
  \hfill
  \begin{subfigure}{0.475\textwidth}\centering
    \caption{Gradient Flow - High learning rate (0.1)}\label{fig:c100_high_lr_gf}
    \includegraphics[width=\textwidth]{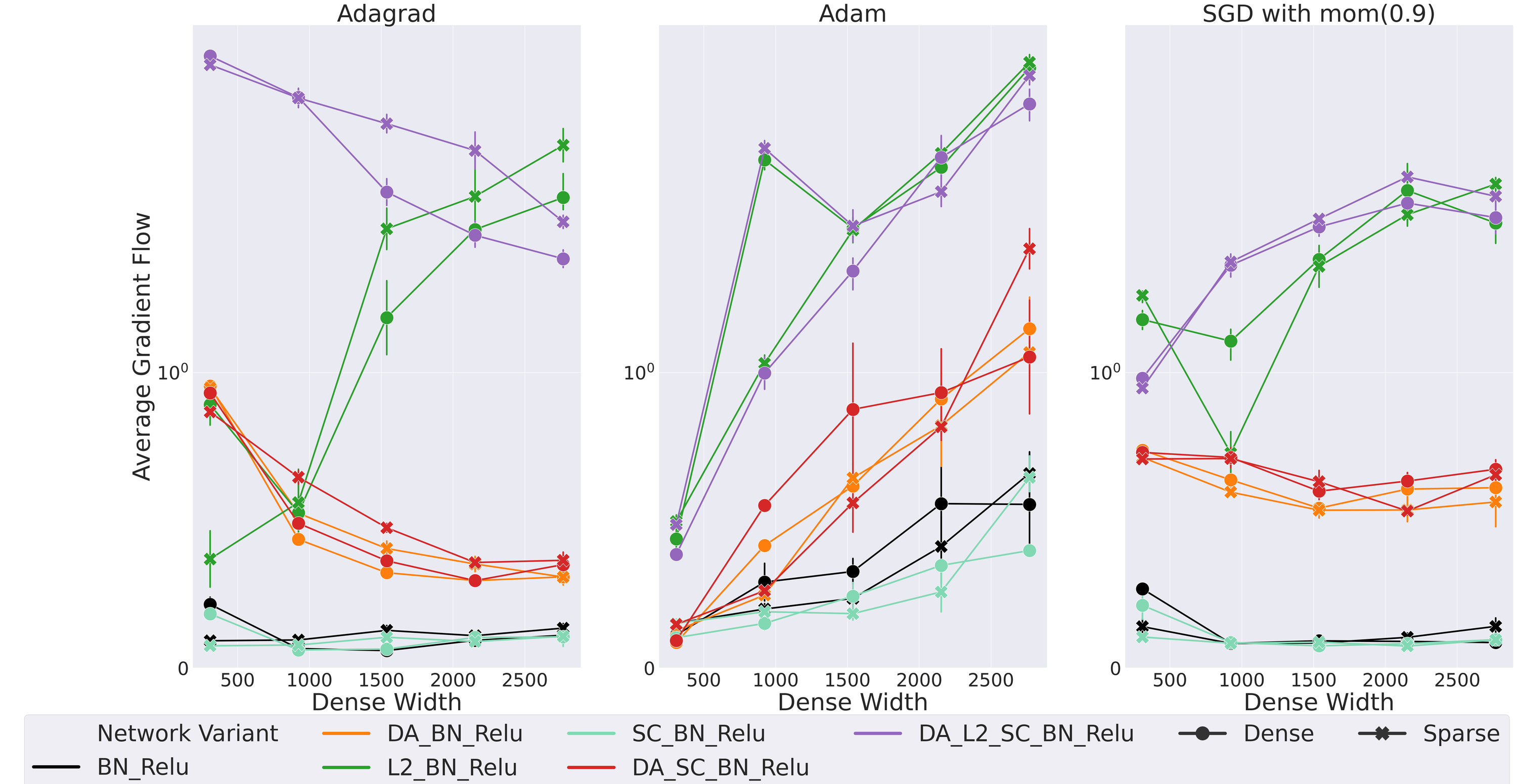}
  \end{subfigure}
  \vskip\baselineskip
  {\textit{NR - No Regularization, BN - Batchnorm, SC - Skip Connections, DA - Data Augmentation and L2- weight decay.}}
  \\
  \caption{\textbf{Test Accuracy and Gradient Flow in Sparse and Dense MLPs.} We study the effect of different regularization and optimization methods on test accuracy and average gradient flow, across different learning rates. We see that for Adam, a higher gradient flow tends to correlate to poor performance. The results for all optimizers can be found in Figures \ref{fig:c100_diff_reg_all_optims_low_lr_all} and \ref{fig:c100_diff_reg_all_optims_high_lr_with_batchnorm}.
  }
  \label{fig:c100_acc_gf}
\end{figure}

\textbf{In EWMA Optimizers, BatchNorm Brings Sparse and Dense Networks Closer in Performance}
When Adam or RMSProp are used as optimization algorithms, we see that sparse networks outperform dense networks without BatchNorm (\textit{NR} from Table \ref{tbl:c100_reg_low_lr} and \ref{tbl:c100_reg_high_lr}). When BatchNorm is added to these methods (\textit{BN}), we see that sparse networks lose their advantage and they perform similarly or slightly better than their dense counterparts in test accuracy and gradient flow, as can be seen from Figure \ref{fig:c100_acc_gf}.

\textbf{At Shallow Depths, BatchNorm is More Critical to Performance than Skip Connections}
Figures \ref{fig:c100_low_lr_acc} and \ref{fig:c100_high_lr_acc} show that BatchNorm (\textit{BN}) is more critical in terms of test accuracy than skip connections (\textit{SC}). Furthermore, we see that BatchNorm (\textit{BN}) performs similarly, in terms of test accuracy, to skip connections and BatchNorm combined (\textit{SC\_BN}). This suggests that skip connections do not provide a performance benefit when applied with BatchNorm. We believe this is due to the shallow depth of these networks (four hidden layers), as skip connections have proved to be critical for deeper networks, even with BatchNorm \citep{balduzzi2017shattered,yang2019mean, labatie2019characterizing}.

\textbf{BatchNorm Stabilizes Gradient Flow}
Across all optimizers and learning rates, we see that when BatchNorm is used, it results in a lower, more stable \texttt{EGF} (Figures \ref{fig:c100_low_lr_gf} and \ref{fig:c100_high_lr_gf}), compared to the original network without BatchNorm (comparing \textit{NR} to \textit{BN}, \textit{L2} to \textit{L2\_BN}, \textit{DA} to \textit{DA\_BN} and \textit{SC} to \textit{SC\_BN}). This further emphasizes the importance of BatchNorm in stabilizing gradient flow.

\begin{table}[ht]
  \small
  \begin{subtable}{\linewidth}\centering
    \begin{center}
      \caption{Different Regularization Methods - Low learning rate (0.001)}
      
\begin{tabular}{lrrrrrrrr}
\toprule
& \multicolumn{1}{l}{NR}                 & \multicolumn{1}{l}{DA}                 & \multicolumn{1}{l}{L2}        & \multicolumn{1}{l}{SC}                 & \multicolumn{1}{l}{BN}                 & \multicolumn{1}{l}{DA\_BN}             & \multicolumn{1}{l}{L2\_BN}             & \multicolumn{1}{l}{SC\_BN}             \\
\midrule
Adagrad   & \cellcolor[HTML]{E67C73}1.000          & \cellcolor[HTML]{E67C73}1.000          & \cellcolor[HTML]{E77D72}0.998 & \cellcolor[HTML]{A7C779}0.239          & \cellcolor[HTML]{58BB8A}\textbf{0.006} & \cellcolor[HTML]{57BB8A}\textbf{0.002} & \cellcolor[HTML]{57BB8A}\textbf{0.001} & \cellcolor[HTML]{57BB8A}\textbf{0.003} \\
Adam      & \cellcolor[HTML]{57BB8A}\textbf{0.000} & \cellcolor[HTML]{69BD87}0.055          & \cellcolor[HTML]{99C57C}0.198 & \cellcolor[HTML]{57BB8A}\textbf{0.003} & \cellcolor[HTML]{71BF85}0.079          & \cellcolor[HTML]{68BD87}0.051          & \cellcolor[HTML]{ACC878}0.254          & \cellcolor[HTML]{8EC37F}0.166          \\
RMSProp   & \cellcolor[HTML]{57BB8A}\textbf{0.001} & \cellcolor[HTML]{57BB8A}\textbf{0.000} & \cellcolor[HTML]{BBCB75}0.300 & \cellcolor[HTML]{8EC37F}0.166          & \cellcolor[HTML]{7EC182}0.117          & \cellcolor[HTML]{5DBC89}\textbf{0.021} & \cellcolor[HTML]{EB8C70}0.914          & \cellcolor[HTML]{FDCF67}0.541          \\
SGD       & \cellcolor[HTML]{E67C73}1.000          & \cellcolor[HTML]{E67C73}1.000          & \cellcolor[HTML]{E67C73}1.000 & \cellcolor[HTML]{AAC879}0.248          & \cellcolor[HTML]{57BB8A}\textbf{0.000} & \cellcolor[HTML]{57BB8A}\textbf{0.000} & \cellcolor[HTML]{57BB8A}\textbf{0.001} & \cellcolor[HTML]{57BB8A}\textbf{0.003} \\
Mom (0.9) & \cellcolor[HTML]{E67C73}1.000          & \cellcolor[HTML]{E67C73}1.000          & \cellcolor[HTML]{E77D72}1.000 & \cellcolor[HTML]{E77D72}0.999          & \cellcolor[HTML]{57BB8A}\textbf{0.001} & \cellcolor[HTML]{57BB8A}\textbf{0.000} & \cellcolor[HTML]{59BB8A}\textbf{0.007} & \cellcolor[HTML]{59BB8A}\textbf{0.008} \\
\bottomrule
\end{tabular}

      \label{tbl:c100_reg_low_lr}
    \end{center}
    \begin{center}
      \caption{Different Regularization Methods - High learning rate (0.1)}

\begin{tabular}{lrrrrrr}
\toprule
          & \multicolumn{1}{l}{BN}                 & \multicolumn{1}{l}{DA\_BN}             & \multicolumn{1}{l}{L2\_BN}             & \multicolumn{1}{l}{SC\_BN}             & \multicolumn{1}{l}{DA\_SC\_BN}         & \multicolumn{1}{l}{DA\_L2\_SC\_BN}     \\
 \midrule
Adagrad   & \cellcolor[HTML]{57BB8A}\textbf{0.000} & \cellcolor[HTML]{57BB8A}\textbf{0.002} & \cellcolor[HTML]{E67C73}0.963          & \cellcolor[HTML]{57BB8A}\textbf{0.002} & \cellcolor[HTML]{5FBC89}\textbf{0.023} & \cellcolor[HTML]{5BBB89}\textbf{0.014} \\
Adam      & \cellcolor[HTML]{6FBE85}0.070          & \cellcolor[HTML]{57BB8A}\textbf{0.002} & \cellcolor[HTML]{66BD87}\textbf{0.043} & \cellcolor[HTML]{58BB8A}\textbf{0.004} & \cellcolor[HTML]{99C57C}0.191          & \cellcolor[HTML]{DAD06E}0.377          \\
RMSProp   & \cellcolor[HTML]{57BB8A}\textbf{0.002} & \cellcolor[HTML]{60BC89}\textbf{0.027} & \cellcolor[HTML]{59BB8A}\textbf{0.008} & \cellcolor[HTML]{FBC768}0.562          & \cellcolor[HTML]{EA8971}0.894          & \cellcolor[HTML]{57BB8A}\textbf{0.002} \\
SGD       & \cellcolor[HTML]{57BB8A}\textbf{0.001} & \cellcolor[HTML]{57BB8A}\textbf{0.000} & \cellcolor[HTML]{67BD87}\textbf{0.048} & \cellcolor[HTML]{58BB8A}\textbf{0.005} & \cellcolor[HTML]{57BB8A}\textbf{0.001} & \cellcolor[HTML]{5BBB8A}\textbf{0.013} \\
Mom (0.9) & \cellcolor[HTML]{57BB8A}\textbf{0.001} & \cellcolor[HTML]{57BB8A}\textbf{0.002} & \cellcolor[HTML]{FFD566}0.488          & \cellcolor[HTML]{57BB8A}\textbf{0.003} & \cellcolor[HTML]{58BB8A}\textbf{0.005} & \cellcolor[HTML]{A0C67B}0.212  \\
\bottomrule
\end{tabular}

      \label{tbl:c100_reg_high_lr}
    \end{center}
    \begin{center}
      \caption{Different Activation Functions - High learning rate (0.1)}
      {
\begin{tabular}{lrrrrrr}
\toprule
          & \multicolumn{1}{l}{ReLU}               & \multicolumn{1}{l}{Swish}               & \multicolumn{1}{l}{PReLU}               & \multicolumn{1}{l}{SReLU}              & \multicolumn{1}{l}{Sigmoid}             & \multicolumn{1}{l}{ELU}                \\
\midrule
Adagrad   & \cellcolor[HTML]{5EBC89}\textbf{0.023} & \cellcolor[HTML]{58BB8A}\textbf{0.005} & \cellcolor[HTML]{67BD87}\textbf{0.050} & \cellcolor[HTML]{94C47D}0.182          & \cellcolor[HTML]{FCCA67}0.568          & \cellcolor[HTML]{58BB8A}\textbf{0.003} \\
Adam      & \cellcolor[HTML]{97C57D}0.191          & \cellcolor[HTML]{94C47D}0.182          & \cellcolor[HTML]{64BD88}\textbf{0.039} & \cellcolor[HTML]{6BBE86}0.062          & \cellcolor[HTML]{58BB8A}\textbf{0.005} & \cellcolor[HTML]{57BB8A}\textbf{0.000} \\
RMSProp   & \cellcolor[HTML]{EC9070}0.894          & \cellcolor[HTML]{8FC47E}0.167          & \cellcolor[HTML]{57BB8A}\textbf{0.002} & \cellcolor[HTML]{5ABB8A}\textbf{0.012} & \cellcolor[HTML]{E77D72}0.997          & \cellcolor[HTML]{8AC37F}0.153          \\
SGD       & \cellcolor[HTML]{5BBB8A}\textbf{0.013} & \cellcolor[HTML]{5FBC89}\textbf{0.027} & \cellcolor[HTML]{58BB8A}\textbf{0.005} & \cellcolor[HTML]{71BF85}0.078          & \cellcolor[HTML]{61BC88}\textbf{0.030} & \cellcolor[HTML]{69BE86}0.056          \\
Mom (0.9) & \cellcolor[HTML]{9EC67B}0.212          & \cellcolor[HTML]{5BBB8A}\textbf{0.013} & \cellcolor[HTML]{57BB8A}\textbf{0.001} & \cellcolor[HTML]{71BF85}0.078          & \cellcolor[HTML]{57BB8A}\textbf{0.001} & \cellcolor[HTML]{E88172}0.973    \\
\bottomrule
\end{tabular}

}
      \label{tbl:activation_c100}
    \end{center}
  \end{subtable}
  \vspace{1ex}
  {
    \normalsize
    \newline \newline \centering
    Colour Scale based on p-values :
    \resizebox{0.5\textwidth}{!}{
      \newcolumntype{C}{>{\centering\arraybackslash}p{0.75em}}
\begin{tabular}{*{11}{C}}
\multicolumn{1}{r}{\cellcolor[HTML]{57BB8A}} & \multicolumn{1}{r}{\cellcolor[HTML]{78C083}} & \multicolumn{1}{r}{\cellcolor[HTML]{9AC57C}} & \multicolumn{1}{r}{\cellcolor[HTML]{BBCB75}} & \multicolumn{1}{r}{\cellcolor[HTML]{DDD06E}} & \multicolumn{1}{r}{\cellcolor[HTML]{FFD666}} & \multicolumn{1}{r}{\cellcolor[HTML]{FBC568}} & \multicolumn{1}{r}{\cellcolor[HTML]{F6B36B}} & \multicolumn{1}{r}{\cellcolor[HTML]{F0A06D}} & \multicolumn{1}{r}{\cellcolor[HTML]{EB8E70}} & \multicolumn{1}{r}{\cellcolor[HTML]{E67C73}} \\
0  &  &  &  &   & .5 &  &  &  &   & 1  \\
\multicolumn{2}{c}{S  $>$ D} & & & & & & & & \multicolumn{2}{c}{S $\leq$ D} \\ 
\end{tabular}

    }
    \newline \newline
  }
  \vspace{1ex}
  {
    \normalsize
    \textit{NR - No Regularization, BN - Batchnorm, SC - Skip Connections, DA - Data Augmentation, L2- weight decay, D - Dense Networks and S - Sparse Networks.}}
  \vspace{2ex}
  \caption{\textbf{Wilcoxon Signed Rank Test Results for MLPs with Four Hidden Layers, Trained on CIFAR-100.} We show the results using different optimization and regularization methods, across various sparsity levels as mentioned in Section \ref{subsec:dense_width}. We use a $p$-value of 0.05, with the bold values indicating where we can be statistically confident that sparse networks perform better than dense (reject $H_0$ from \ref{txt:wil_test}). We also use a continuous colour scale to make the results more interpretable. This scale ranges from green (0 - likely that sparse networks perform better than dense) to yellow (0.5 - 50\% chance that sparse networks perform better than dense) to red (1 - highly likely that sparse networks do not outperform dense - cannot reject $H_0$ from \ref{txt:wil_test}). The performance results for all these networks are present in Appendix \ref{appen:detailed_sdsdc}.}\label{tbl:wilcoxon_reg}

\end{table}

\begin{figure}[ht]
  \centering
  \begin{subfigure}{0.475\textwidth} \centering
    \caption{Test Accuracy}\label{fig:c100_diff_act_acc}
    \includegraphics[width=0.9\textwidth]{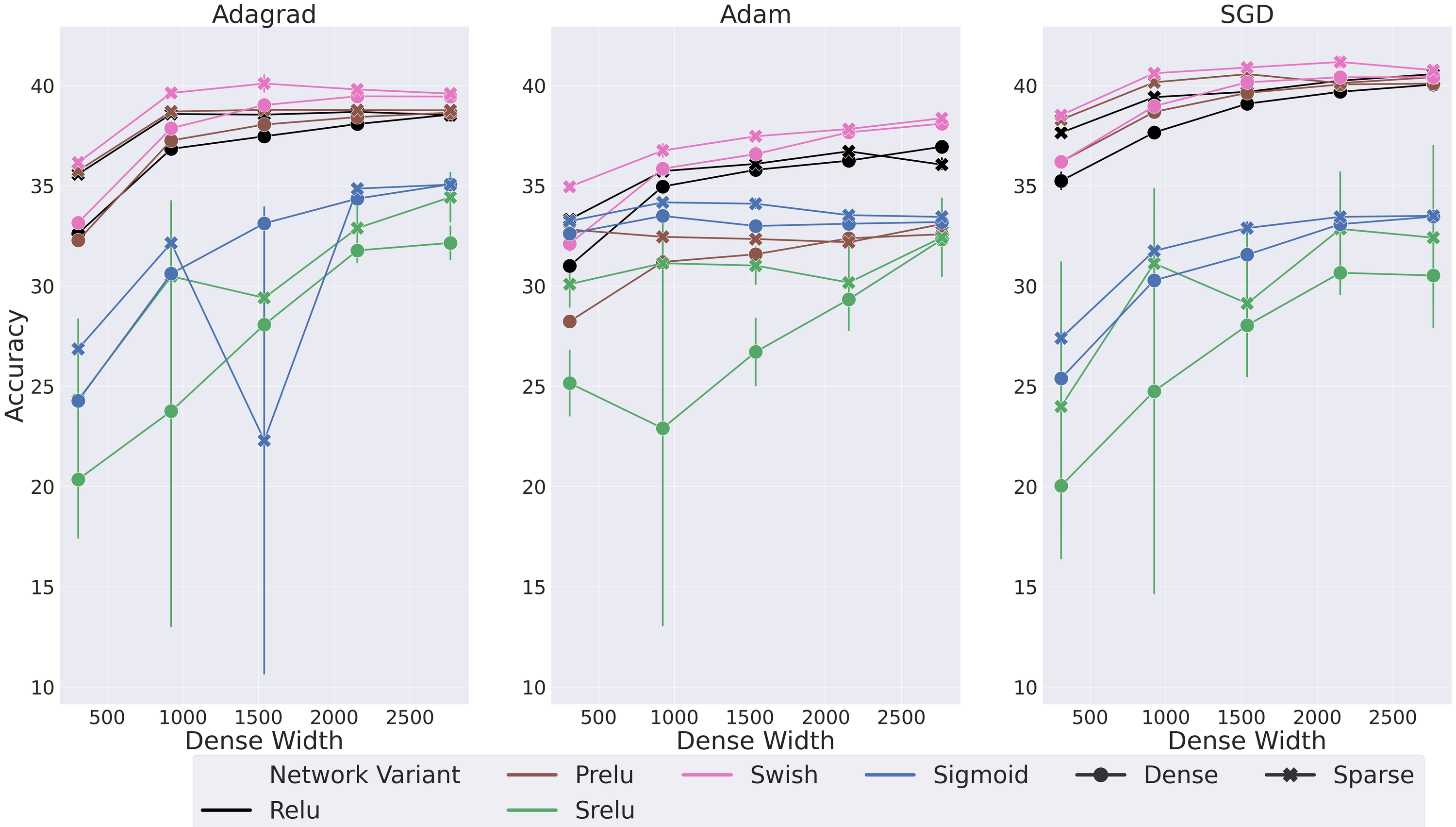}
  \end{subfigure}
  \begin{subfigure}{0.475\textwidth} \centering
    \caption{Gradient Flow}\label{fig:c100_diff_act_gf}
    \includegraphics[width=0.9\textwidth]{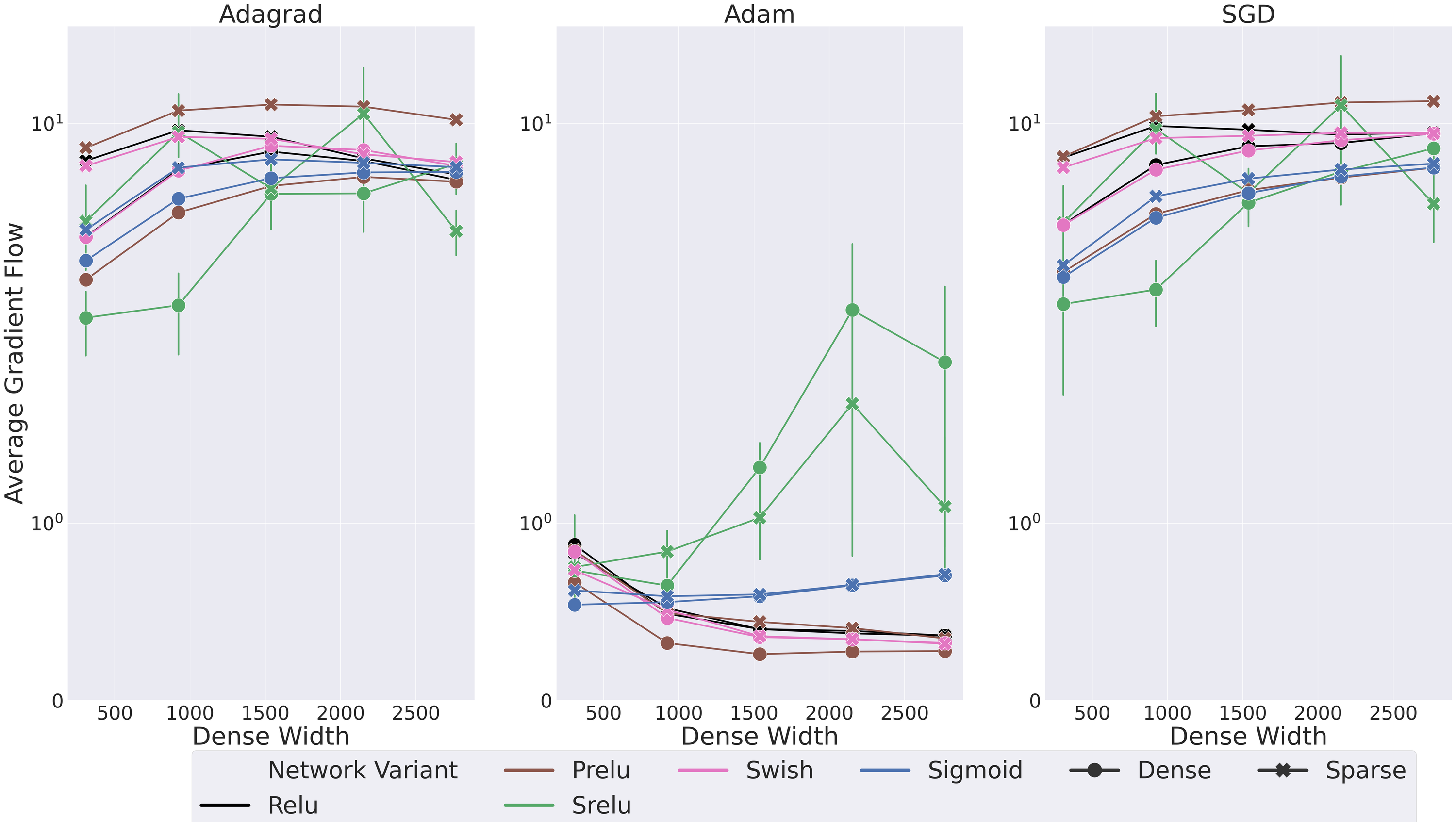}
  \end{subfigure}
  \caption{\textbf{Effect of Activation Functions on Accuracy and Gradient Flow on CIFAR-100, With a Low Learning Rate (0.001).} We see that Swish is the most promising activation function across most optimizers. The results across all optimizers and learning rates are shown in Figure \ref{fig:c100_diff_reg_all_optims_low_lr_acts} and \ref{fig:c100_diff_reg_all_optims_high_lr_acts}.}
  \label{fig:c100_diff_act}
\end{figure}

\subsubsection{EWMA optimizers are sensitive to high gradient flow}

\textbf{In EWMA Optimizers, $L2$ Regularization can hurt Network Performance} For networks without BatchNorm and with $L2$ regularization (configuration \textit{$L2$}), we can see from Figure \ref{fig:c100_low_lr_acc} that when using adaptive methods (Adagrad, Adam and RMSProp), $L2$ regularization mainly hurts sparse network performance. This occurs particularly at high sparsity levels (a dense width of less than 2153). Conversely, equivalent capacity and configured dense networks achieve a performance improvement with $L2$ regularization, compared to their base configuration (comparing \textit{NR} to \textit{L2}).

For networks with BatchNorm, trained with EWMA optimizers, $L2$ regularization adversely affects both sparse and dense network performance. From Figures \ref{fig:c100_high_lr_acc} and \ref{fig:c100_high_lr_gf}, we see that the addition of $L2$ (configurations \textit{L2\_BN} and \textit{DA\_L2\_SC\_BN}) drastically decreases the accuracy of these networks. When we analyse their \texttt{EGF}, from Figure \ref{fig:c100_high_lr_gf}, we see that the addition of $L2$ consistently across all optimizers results in distinctively larger \texttt{EGF} values. This hints at EWMA optimizers being more sensitive to larger gradient norms than other optimizers. 

The poor performance of adaptive methods with $L2$ regularization (specifically EWMA optimizers) agrees with \cite{loshchilov2017decoupled}. They proposed a different formulation of weight decay for Adam, named AdamW, since the current $L2$ regularization formulation for adaptive methods could lead to weights with large gradients being regularized less. We experimentally verified this in Figure \ref{fig:c100_adamw}, showing that the AdamW's weight decay formulation has a lower \texttt{EGF} than the standard $L2$ formulation used in Adam, and this correlates to better network performance in AdamW.

\textbf{Data Augmentation Favours Non-EWMA Optimizers}
When networks are trained with data augmentation and without BatchNorm (configuration \textit{DA}), we see poor test accuracy for EWMA optimizers, across sparse and dense networks (Figure \ref{fig:c100_low_lr_acc}). With the addition of BatchNorm (configuration \textit{BN\_DA}), when using a low learning rate, we see that data augmentation benefits all optimizers (Figure \ref{fig:c100_low_lr_acc}). This behaviour differs when using a high learning rate (Figure \ref{fig:c100_high_lr_acc}), where data augmentation results in a higher gradient flow in all optimizers (comparing \textit{BN} to \textit{DA\_BN}). This then results in lower performance in EWMA optimizers, further emphasizing the sensitivity of these optimizers to higher gradient flow. 

For non-EWMA optimizers, data augmentation consistently increases performance, even with a higher learning rate and higher \textit{EGF}, which suggests data augmentation is better suited for these methods.  

\textbf{EWMA Optimizers Struggle with High Gradient Flow} If we take a closer look at the gradient flow, through the average \texttt{EGF} (Figures \ref{fig:c100_low_lr_gf} and \ref{fig:c100_high_lr_gf}), we see that for EWMA optimizers the worst performing variants (\textit{L2}, \textit{NR}, \textit{DA}, \textit{L2\_BN} and \textit{DA\_L2\_SC\_BN}) consistently have a relatively high \texttt{EGF}, while the best performing interventions (\textit{BN}, \textit{BN\_SC} and \textit{DA\_BN\_SC}) have a lower \texttt{EGF}. This is also true of different activation functions, where the worst performing activation function when using Adam, SReLU, also has the highest \texttt{EGF} (Figure \ref{fig:c100_diff_act}).

In non-EWMA optimizers, a higher gradient flow does not consistently lead to poor performance. This hints that the higher effective learning rates of EWMA optimizers can be problematic during training, primarily when used in conjunction with methods that result in high \texttt{EGF}. The sensitivity of networks trained with EWMA optimizers, particularly sparse networks, to $L2$ and data augmentation suggests that, in their current formulation, these methods are not adequate for sparse networks.

\subsubsection{The potential of non-sparse activation functions - Swish and PReLU}

We also explore the impact of different activation functions - specifically PReLU \citep{he2015delving}, ELU \citep{clevert2015fast}, Swish \citep{ramachandran2017swish}, SReLU \citep{jin2015deep} and Sigmoid \citep{neal1992connectionist} -  on network performance. The best regularization configuration for each optimizer was chosen.

From Table \ref{tbl:activation_c100} and \ref{tbl:activation_c100_low_lr}, we see that Swish and PReLU consistently favour sparse networks across most optimizers in a statistically significant manner. From the performance results shown in Figures \ref{fig:c100_diff_act_acc}, \ref{fig:c100_diff_reg_all_optims_low_lr_acts} and \ref{fig:c100_diff_reg_all_optims_high_lr_acts}, we see that Swish and PReLU are the most promising activation functions, with Swish being the best performing activation function in adaptive methods, and PReLU achieving promising results for networks trained with SGD.

In Appendix \ref{app_sec:activations}, we plot the different activation functions (Figure \ref{fig:act}) and their gradients (Figure \ref{fig:act_grad}). We see that although most activation functions, apart from ReLU, are non-sparse (do not have zero gradients), Swish is the only activation function that allows for the flow of negative gradients due to its non-monotonicity. This leads to Swish having a lower, more stable gradient flow (Figures \ref{fig:c100_diff_act_gf}, \ref{fig:c100_diff_reg_all_optims_low_lr_acts} and \ref{fig:c100_diff_reg_all_optims_high_lr_acts}), which could explain some of its success, particularly for EWMA optimizers. We continue to see a consistent trend in EWMA methods that higher \texttt{EGF} values, for example in SReLU, correspond to poor performance, while promising methods, such as Swish, result in a lower \texttt{EGF}.

We also see that the behaviours mentioned in this section are also present in CIFAR-10 and FMNIST (Table \ref{tbl:fmnist_reg_wilcoxon} and \ref{tbl:c10_reg_wilcoxon}, Figures \ref{fig:fmnist_diff_reg_all_optims_low_lr} and \ref{fig:c100_diff_reg_all_optims_low_lr_all}).

\subsection{Generalization of results across architecture types - Wide ResNet-50} \label{subsec:wres}

\begin{figure}[ht]
  \begin{center} \label{img:acc_c100}
    \includegraphics[width=0.90\linewidth]{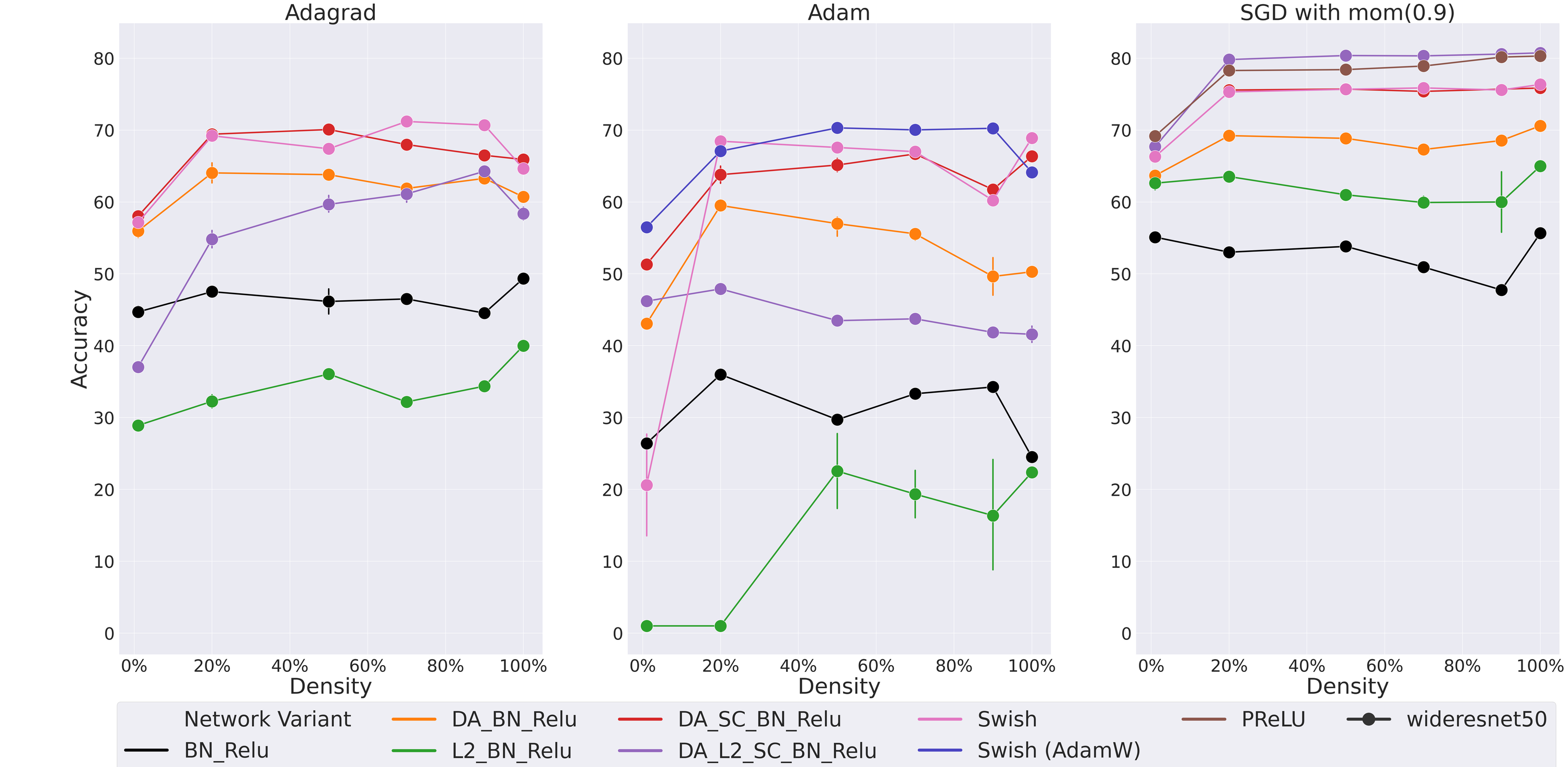}
  \end{center}
  \caption{\textbf{Wide ResNet-50 Test Accuracy on CIFAR-100.} We see that the results achieved on MLPs, using \texttt{SC-SDC}, are also consistent in CNNs. The densities range from 1\% to 100\% (fully dense) and the gradient flow results can be found in Figure \ref{fig:wres_grad_flow}.} \label{fig:wres_acc}
\end{figure}

We move on from \texttt{SC-SDC} and extend our results to more complicated, convolutional architectures. We train Wide ResNet-50 (the WRN-28-10 variant) \citep{Zagoruyko2016} on CIFAR-100. We note from Figure \ref{fig:wres_acc} that most of our results from \texttt{SC-SDC} also hold on Wide ResNet-50, specifically that $L2$ regularization (even with BatchNorm) hurts performance for adaptive methods (Adagrad and Adam) and also results in higher \texttt{EGF} values (Figure \ref{fig:wres_grad_flow}). Furthermore, we see that Swish is a promising activation function for adaptive methods and leads to lower \texttt{EGF} (Figure \ref{fig:wres_grad_flow}). Finally, we see that the combination of Swish and AdamW (configuration \textit{Swish (AdamW)}) achieve good performance for Adam, showing that the AdamW (weight decay) results from \texttt{SC-SDC} are also consistent in Wide ResNet-50. These results show that the \texttt{SC-SDC} results are not constrained to small scale experiments and that they can be used to learn about the dynamics of larger, more complicated networks.

\subsection{Generalization of results from Random Pruning to Magnitude Pruning}

\begin{figure}[ht]
     \centering
    \begin{subfigure}{0.4\textwidth} \centering
        \caption{Test Accuracy - Magnitude Pruning}\label{fig:c100_mag_prune_acc}
        \includegraphics[width=0.9\textwidth]{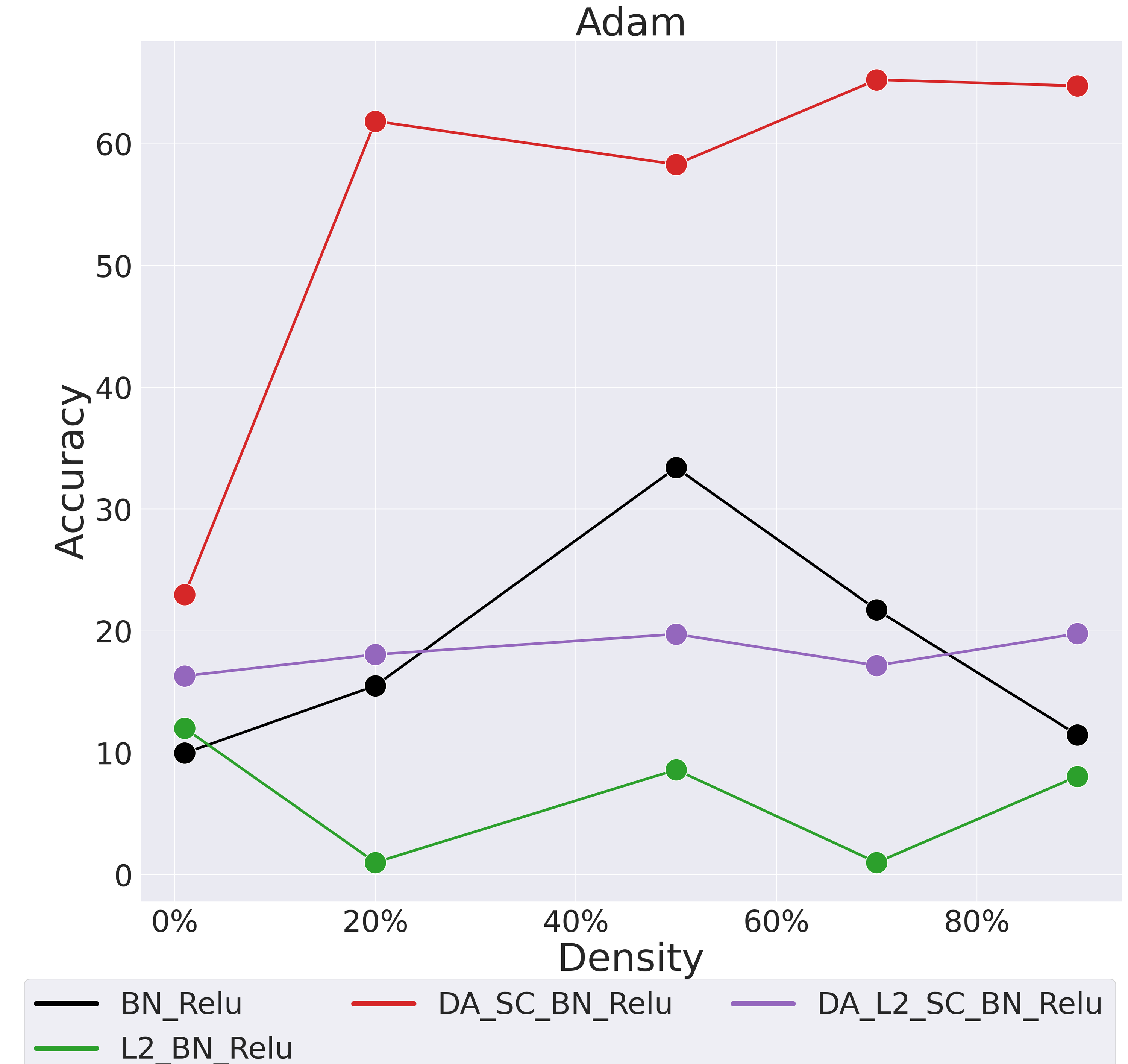}
    \end{subfigure}
     \begin{subfigure}{0.4\textwidth} \centering
        \caption{Gradient Flow - Magnitude Pruning}\label{fig:c100_mag_prune_gf}
        \includegraphics[width=0.9\textwidth]{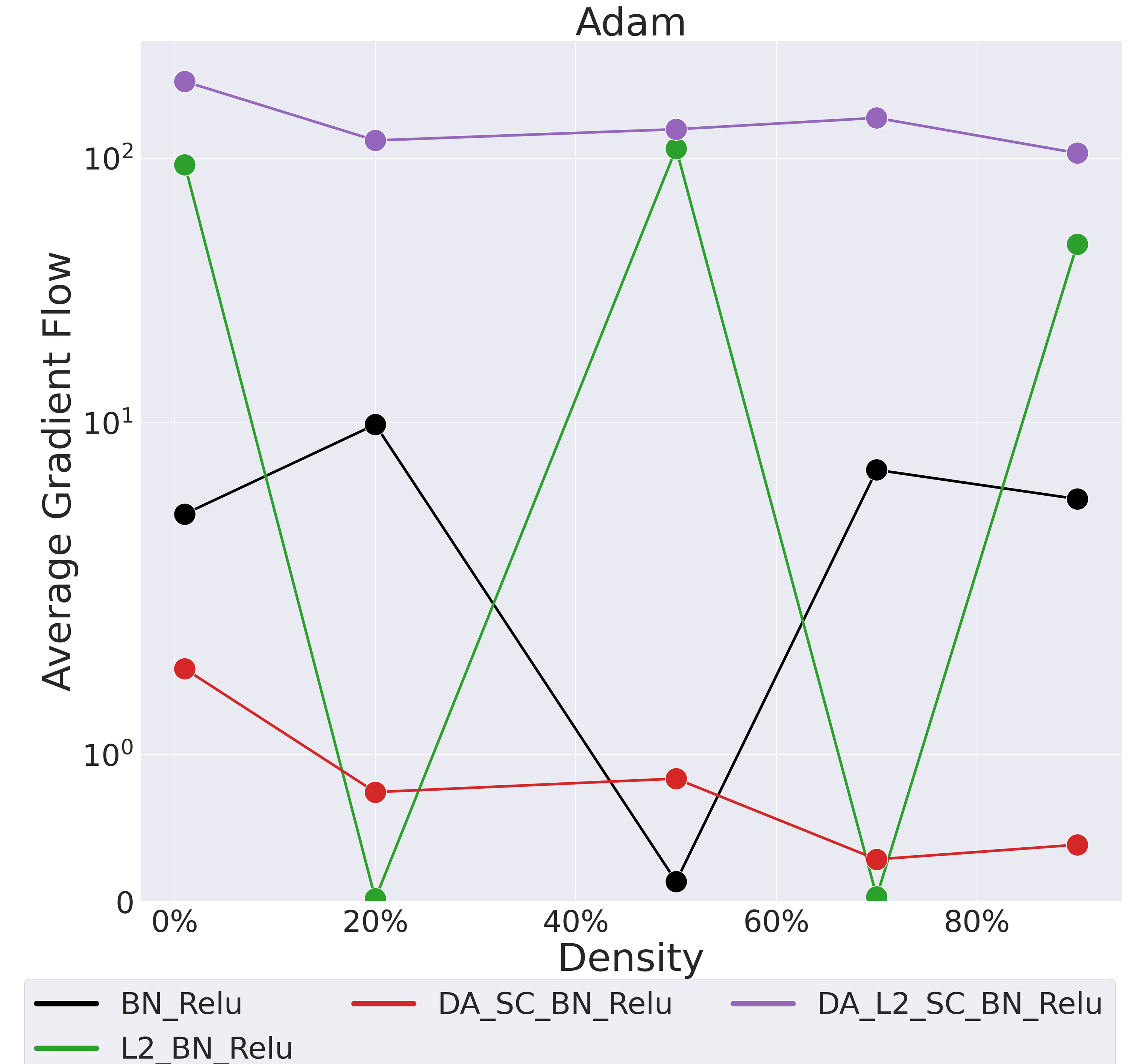}
    \end{subfigure}
    \caption{\textbf{Accuracy and Gradient Flow for Magnitude Pruning.} We see that similarly to randomly pruned networks, magnitude pruned networks trained with Adam and $L2$ lead to high \texttt{EGF} and poor performance. }
    \label{fig:c100_mag_prune}
\end{figure}

We briefly validate if some of our results achieved through random pruning, extend to magnitude pruning \citep{zhu2017prune,Han2015}. For a comparable experimental setting to section \ref{subsec:wres}, we train dense Wide ResNet-50 for 100 epochs (50\% of the training time) and use magnitude pruning to achieve the desired sparsity, and then fine-tune for the remaining 100 epochs. From Figures \ref{fig:c100_mag_prune_acc} and \ref{fig:c100_mag_prune_gf}, we see that although it appears magnitude pruned networks are more susceptible to vanishing gradients, these networks behave similarly to randomly pruned networks. In these networks, $L2$ regularization leads to high \texttt{EGF} values (\textit{L2\_BN} and \textit{DA\_L2\_SC\_BN}), which correlates to poor performance for EWMA optimizers (Adam). This provides evidence that some of the results achieved on random, sparse networks also extend to magnitude pruned networks.

\section{Related work}

\textbf{Pruning at Initialization} Methods that prune at initialization aim to start sparse, instead of first pre-training an overparameterized network and then pruning. These methods use certain criteria to estimate which weights should remain active at initialization. This criteria includes using the connection sensitivity \citep{lee2018snip}, gradient flow (via the Hessian vector product) \citep{wang2020picking} and conservation of synaptic saliency \citep{tanaka2020pruning}.

\textbf{Pruning during Training} Another branch of pruning is Dynamic Sparse Training, which uses information gathered during the training process, to dynamically update the sparsity pattern of these sparse networks  \citep{mostafa2019parameter,bellec2017deep,mocanu2018scalable,dettmers2019sparse,evci2019riggingarXiv191111134E}. While our work is motivated by the same goal of improving the performance of sparse networks and allowing them to converge to the same performance as dense networks, we instead focus on the impact of optimization and regularization choices on sparse networks.

\textbf{Sparse Network Optimization as Pruning Criteria} Optimization in sparse networks has often been neglected in favour of studying network initialization. However, there has been work that has looked at sparse network optimization from different perspectives, mainly as a guide for pruning criteria. This includes using gradient information \citep{NIPS1988_119,optimal-brain-damage,Hassibi93secondorder,karnin1990simple}, approximates of gradient flow \citep{wang2020picking,dettmers2019sparse,evci2020gradient}, and Neural Tangent Kernel (NTK) \citep{liu2020finding} to guide the introduction of sparsity.

\textbf{Sparse Network Optimization to study Network Dynamics} Apart from being used as pruning criteria, optimization information has been used to investigate aspects of sparse networks, such as their loss landscape \citep{evci2019difficulty}, how they are impacted by SGD noise \citep{frankle2019linear}, the effect of different activation functions \citep{dubowski2020activation} and their weight initialization \citep{lee2019signal}. Our work differs from these approaches as we consider more aspects of the optimization and regularization process in a controlled experimental setting (\texttt{SC-SDC}), while using \texttt{EGF} to reason about some of the results.

\section{Conclusions and Future Work}
In this work, we take a more comprehensive view of sparse optimization strategies and introduce appropriate tooling to measure the impact of architecture and optimization choices on sparse networks (\texttt{EGF}, \texttt{SC-SDC}).

Our results show that BatchNorm is critical to training sparse networks, more so than for dense networks, as it helps stabilize gradient flow.
Furthermore, we show that EWMA optimizers (Adam \citep{kingma2014adam} and RMSProp \citep{hinton2012neural}) are sensitive to high gradient flow (\texttt{EGF}). This results in these optimizers, at times, performing poorly when used with $L2$ regularization or data augmentation. We also show the potential of non-sparse activation functions for sparse networks such as Swish \citep{ramachandran2017swish} and PReLU \citep{he2015delving}, with Swish's non-monotonic formulation allowing for better gradient flow. Finally, we show that our results extend to more complicated models, like Wide ResNet-50 \citep{Zagoruyko2016} and popular pruning methods, such as magnitude pruning \citep{zhu2017prune,Han2015}.

We trust that this work emphasizes that initialization is simply one piece of the sparse network puzzle and that a broader view of sparse network training is necessary for the benefits of sparsity to be fully realized. Needless to say, this research is simply the start at investigating sparse network optimization and does not cover all questions related to it. Future directions could include using \texttt{EGF} and \texttt{SC-SDC} to study other regularization methods, such as Dropout \citep{srivastava2014dropout} or $L1$ regularization \citep{tibshirani1996regression}, or other normalization methods, such as Layer Normalization \citep{ba2016layer}. Furthermore, these tools could inspire novel pruning, regularization, or optimization techniques, that are grounded in stabilizing the gradient flow in sparse networks.

\section*{Acknowledgements}
We thank reviewers and our peers for their valuable inputs and discussions. Notably, we thank Arthur Gretton and Abebe Tessera. Furthermore, we thank Shunmuga Pillay for his assistance with running the large scale experiments on a GPU cluster.

\bibliography{arxiv}
\bibliographystyle{iclr2021_conference}
\newpage
\appendix

\section{SC-SDC}
In this section, we provide more information about \texttt{SC-SDC} and its benefits.

\subsection{SC-SDC implementation details} \label{sdsdc:implementation}
\textbf{Wilcoxon Signed Rank Test} \texttt{SC-SDC} uses the Wilcoxon Signed Rank Test as the statistical test to compare sparse and dense networks. This test is a non-parametric statistical test that compares dependent or paired samples, without assuming the differences between the paired experiments are normally distributed \citep{mcdonald2009handbook,demvsar2006statistical}.

\textbf{Random Sparsity} Our work focuses on the training dynamics of random, sparse networks. We achieve random sparsity, by generating a random mask for each layer and then multiply the weights by this mask during each forward pass. The sparsity is distributed evenly across the network. For example, a 20\% sparse MLP has 20\% of the weights remaining in each layer.

\begin{figure}[ht]
  \centering
  \includegraphics[width=0.65\linewidth]{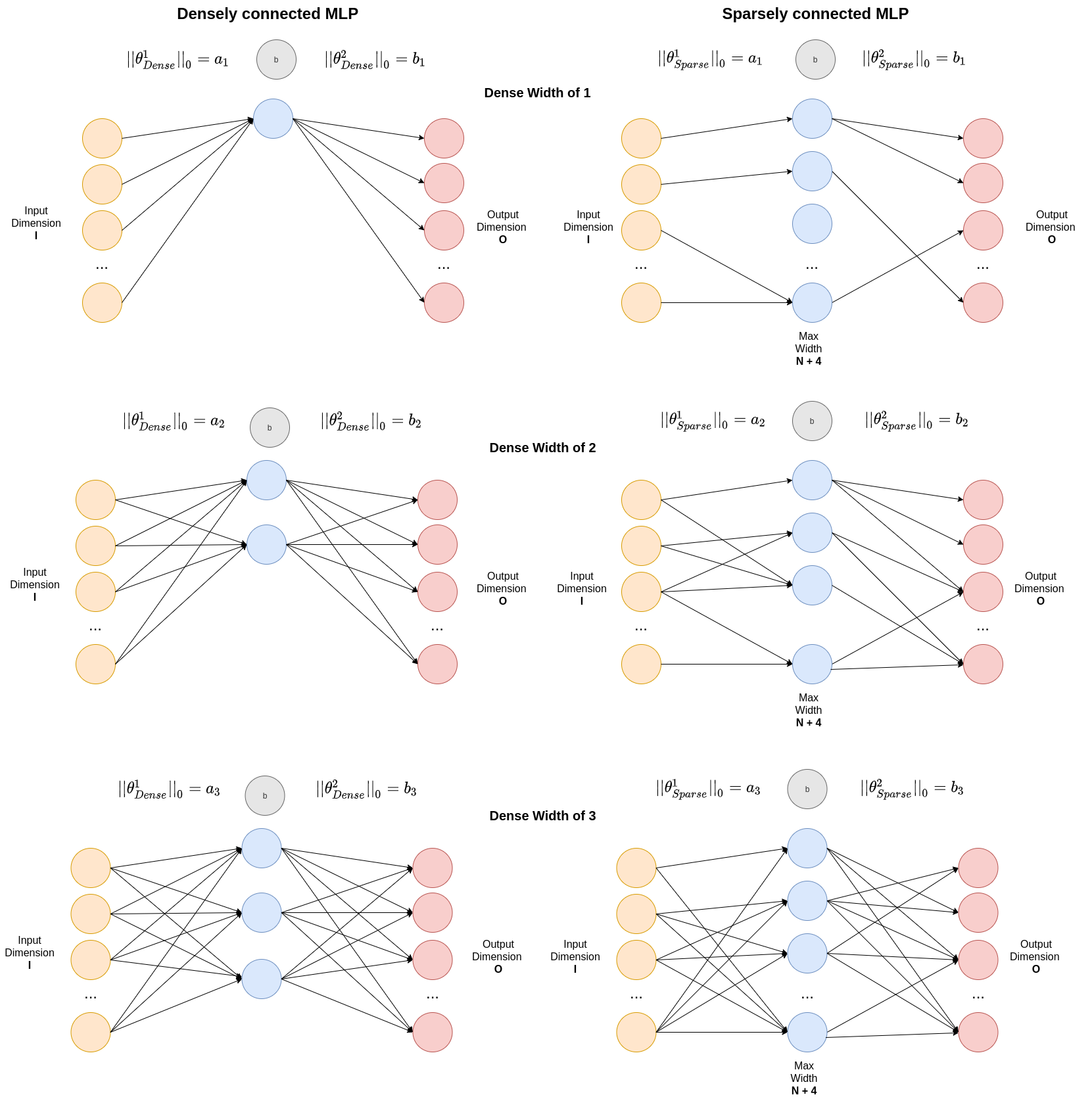}
  \caption{\textbf{Fairly Comparing Sparse and Dense Networks at Different Widths.} We show how we ensure that sparse and dense networks have the same parameter count, at different widths.}
  \label{fig:dense_width}
\end{figure}

\textbf{Dense Width} A critical component to how we specify our experiments is a term we define as \textbf{dense width}. In order to fairly compare sparse and dense networks, we need them to have the same number of active connections at each depth. In the case of sparse networks, this means ensuring they have the same number of active connections as the dense networks, while remaining sparse. \textbf{Dense width} refers to the width of a network if that network was dense. This process of comparing sparse and dense networks at different \textbf{dense widths} is illustrated in Figure \ref{fig:dense_width}.

\textbf{Fair Comparison of Sparse and Dense Networks} As can be seen from Figure \ref{fig:dense_width}, \texttt{SC-SDC} ensures the exact same active parameter count, but the sparse networks will be connected to more neurons.  It is possible that the increased number of activations being used can lead to sparse networks having higher representational power. However, most work on expressivity of neural networks looks at this from a depth perspective and proves certain depths of networks are universal approximators \citep{eldan2016power,hornik1989multilayer,funahashi1989approximate}. To this end, we ensure these networks have the same depth, but we believe going forward an interesting direction would be ensuring they have a similar amount of active neurons.

\textbf{SC-SDC Comparison Details}
For completeness, we provide more details of how we ensure sparse and dense networks are of the same capacity.

Following from Equation \ref{sdsdc:initialize}, to ensure the same number of weights in sparse and dense networks, we can ensure they have the same number of active weights at each layer as follows:
\begin{equation}
  \displaystyle ||\va_{S}^{l}||_0  = ||\va_{D}^{l}||_0 , \quad \text { for } \quad l=1, \ldots, L \qquad ,
\end{equation}
where $\displaystyle \va_{S}^{l}$ is the nonzero weights in layer $l$ of sparse network $S$ and $\displaystyle \va_{D}^{l}$ is the nonzero weights in layer $l$ of dense network $D$ (all the weights since no masking occurs).

This is achieved by masking each of the weight layers of sparse network $S$:
\begin{equation}
  \displaystyle \va_{S}^{l} = \vtheta_{S}^{l} \odot m^{l}  \quad \text { for } \quad l=1, \ldots, L \qquad ,
\end{equation}
where $m^{l}$ is a random binary matrix (mask) for layer $l$, s.t. $|| m^{l}||_0 = ||\va_{D}^{l}||_0$, where $||\va_{D}^{l}||_0$ is the number of active (nonzero) weights in the dense network and is determined by the chosen comparison width.

For \texttt{SC-SDC}, we need a maximum network width $N_{MW}$ and comparison width $N_{W}$.  We choose a max network width $N_{MW}$ of $n + 4$, where $n$ is the input dimension of the network. In the case of CIFAR, $n=3072$ and so our maximum width $N_{MW}=3076$. The choice of $n+4$ follows from \cite{lu2017expressive}, where the authors prove a universal approximation theorem for width-bounded ReLU networks, with width bounded to $n + 4$. Our comparison width, $N_{W}$, is equivalent to \textbf{dense widths} we vary in our experiments - 308, 923, 1538, 2153, 2768 (10\%, 30\%, 50\%, 70\% and 90\%  of our maximum width $N_{MW}$(3076) when using CIFAR datasets).

The dimensions of each of layers of the different networks, $S$ (sparse) and $D$ (dense), are as follows:
\begin{enumerate}
  \item First Layer:
        \begin{equation}
          \displaystyle \vtheta_{D}^{1} \in \mathbb{R}^{I \times N_{W}} \quad , \quad  \displaystyle \vtheta_{S}^{1} \in \mathbb{R}^{I \times N_{MW}} \quad, \quad m^{1} \in \{0,1\}^{I \times N_{MW}}   \quad \quad  \\
        \end{equation}
  \item Intermediate Layers:
        \begin{equation}
          \displaystyle \vtheta_{S}^{1} \in \mathbb{R}^{N_{W} \times N_{W}} \quad, \quad \displaystyle \vtheta_{S}^{1} \in \mathbb{R}^{ N_{MW} \times  N_{MW}} \quad, \quad m^{\{2,\dots,L-1\}} \in \{0,1\}^{N_{MW}\times  N_{MW}}
        \end{equation}
  \item Final Layer:
        \begin{equation}
          \displaystyle \vtheta_{S}^{L} \in \mathbb{R}^{N_{W} \times O} \quad, \quad \displaystyle \vtheta_{S}^{L} \in \mathbb{R}^{ N_{MW} \times O} \quad, \quad m^{L} \in \{0,1\}^{N_{MW}\times O}
        \end{equation}
\end{enumerate}

, where $N_{MW}$ is maximum width of the sparse layer, $N_{W}$ is the comparison width,  $I$ is the input dimension, $O$ is output dimension,  $L$ is the number of layers in the network, $\displaystyle \vtheta_{S}^{l}$ is the weights in layer $l$ of sparse network $S$ and $\displaystyle \vtheta_{D}^{l}$ is the weights in layer $l$ of dense  network $D$.

\begin{figure}[ht]
  \centering
  \includegraphics[width=0.8\linewidth]{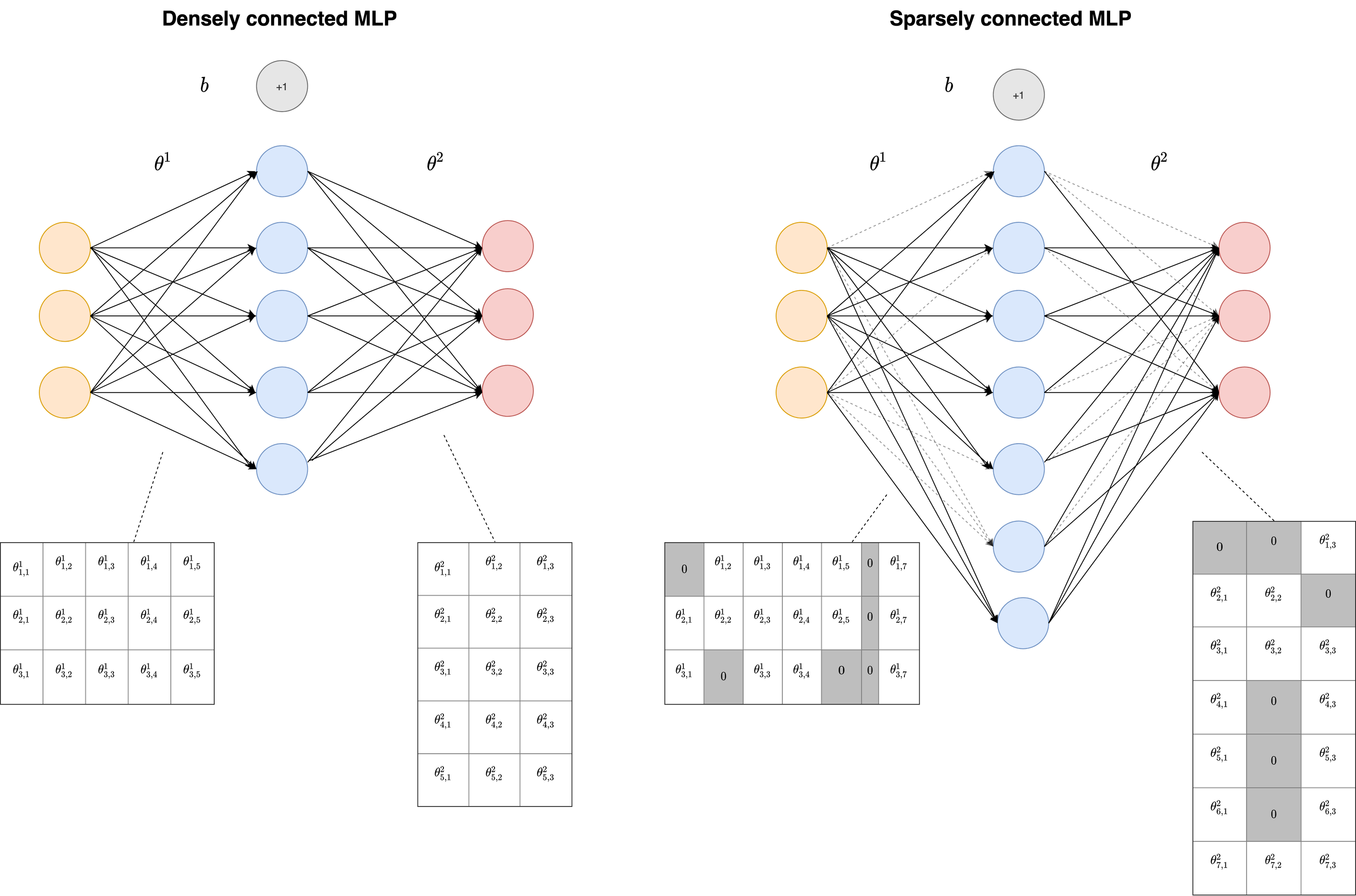}
  \caption{\textbf{Fair Comparison of Sparse and Dense Networks.} We fairly compare sparse and dense networks, by ensuring they have the same number of connections per layer and turning off the bias unit.}
  \label{fig:same_num_connections_sparse_vs_dense}
\end{figure}

This process would be the same for convolutional layers, but there would be a third dimension to handle the different channels. In Figure \ref{fig:same_num_connections_sparse_vs_dense}, we provide an illustrative example showing how to ensure sparse and dense networks are compared fairly.

\subsection{Benefits of SC-SDC}
The benefits of \texttt{SC-SDC} can be summarized as follows:
\begin{itemize}
  \item \textbf{We can better understand sparse network optimization.}
        \texttt{SC-SDC} allows us to identify which optimization or regularization methods are poorly suited to sparse networks in a controlled setting, ensuring the results are a direct result of the sparse connections themselves.
  \item \textbf{Learn at what parameter and size budget, sparse networks are better than dense.}
        Comparing sparse and dense networks of the same capacity allows us to see which architecture is better at different configurations. In configurations where sparse architectures perform better, we could exploit advances in sparse matrix computation and storage \citep{zhao2018bridging,merrill2016merge,ma2019optimizing,zhang2019snap,zhang2020sparch} to simply default to sparse architectures.
\end{itemize}

\section{Gradient Flow} \label{sec:grad_flow_Appendix}

\begin{figure}[ht]
  \caption{Gradient Flow in CIFAR-100 using $EGF_1$, with low learning rate (0.001)}
  \label{fig:EGF1_0.001}
  \begin{center}
    \includegraphics[width=1\textwidth]{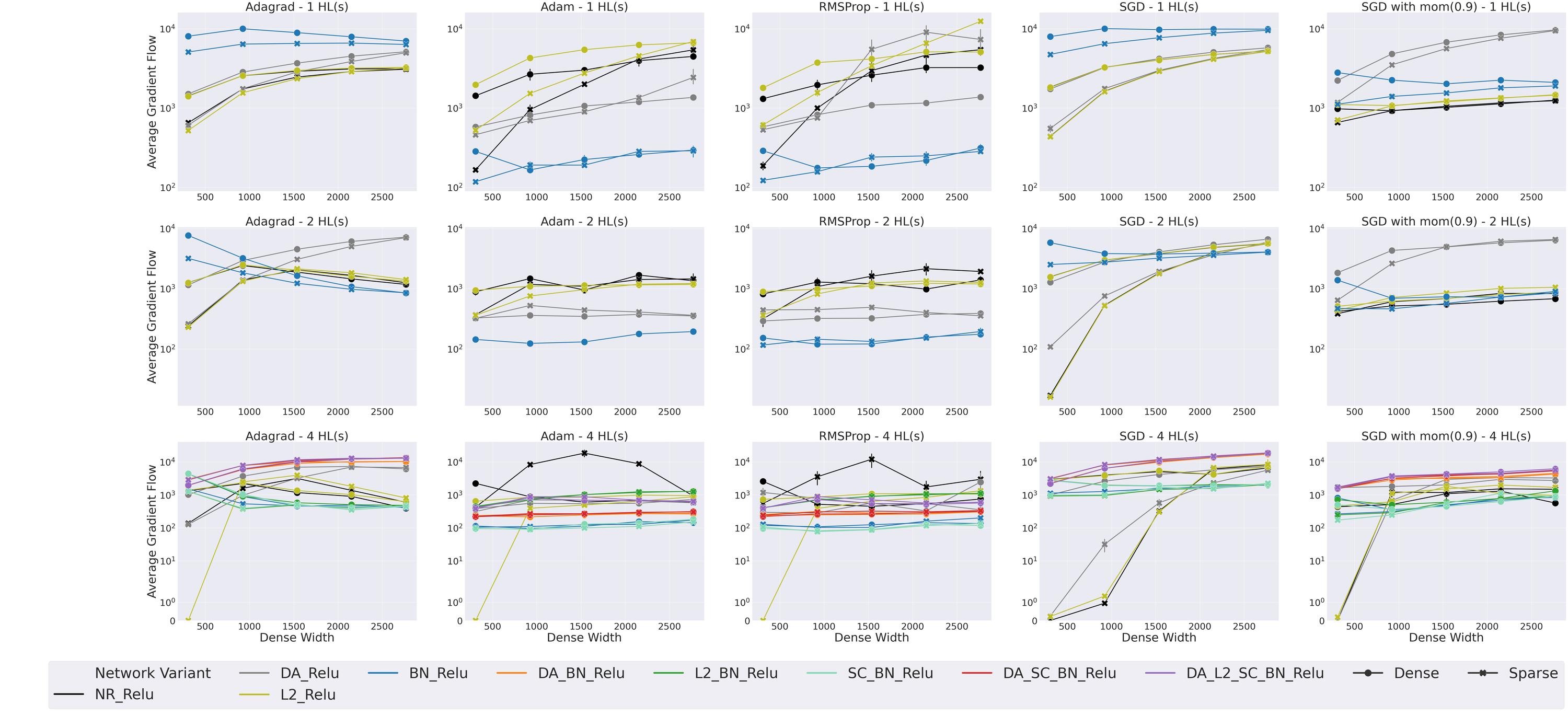}
  \end{center}
\end{figure}

\begin{figure}[ht]
  \caption{Gradient Flow in CIFAR-100 using $|| \vg ||_1$, with low learning rate (0.001)}
  \label{fig:g1_0.001}
  \begin{center}
    \includegraphics[width=1\textwidth]{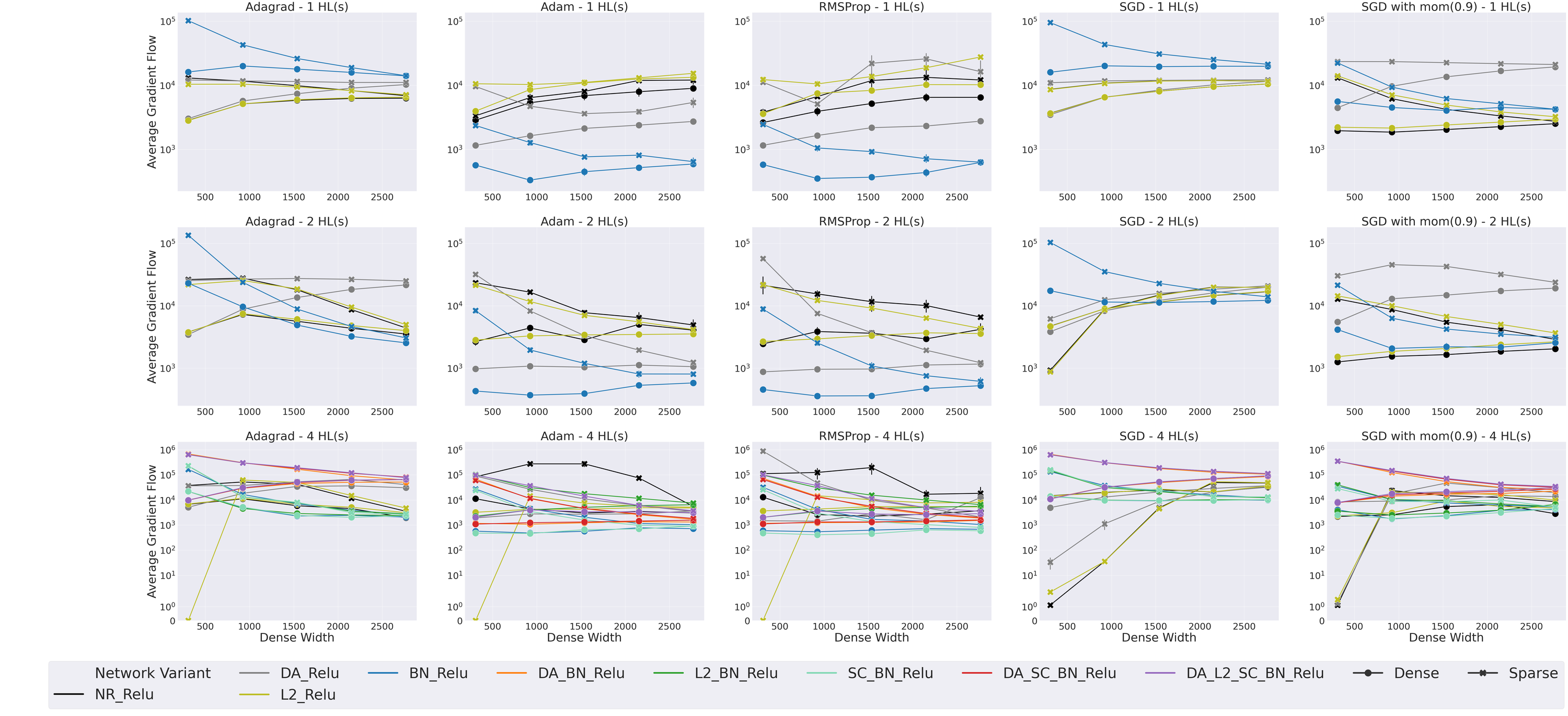}
  \end{center}
\end{figure}

\begin{figure}[ht]
  \caption{Gradient Flow in CIFAR-100 using $|| \vg ||_2$ , with low learning rate (0.001)}
  \label{fig:g2_0.001}
  \begin{center}
    \includegraphics[width=1\textwidth]{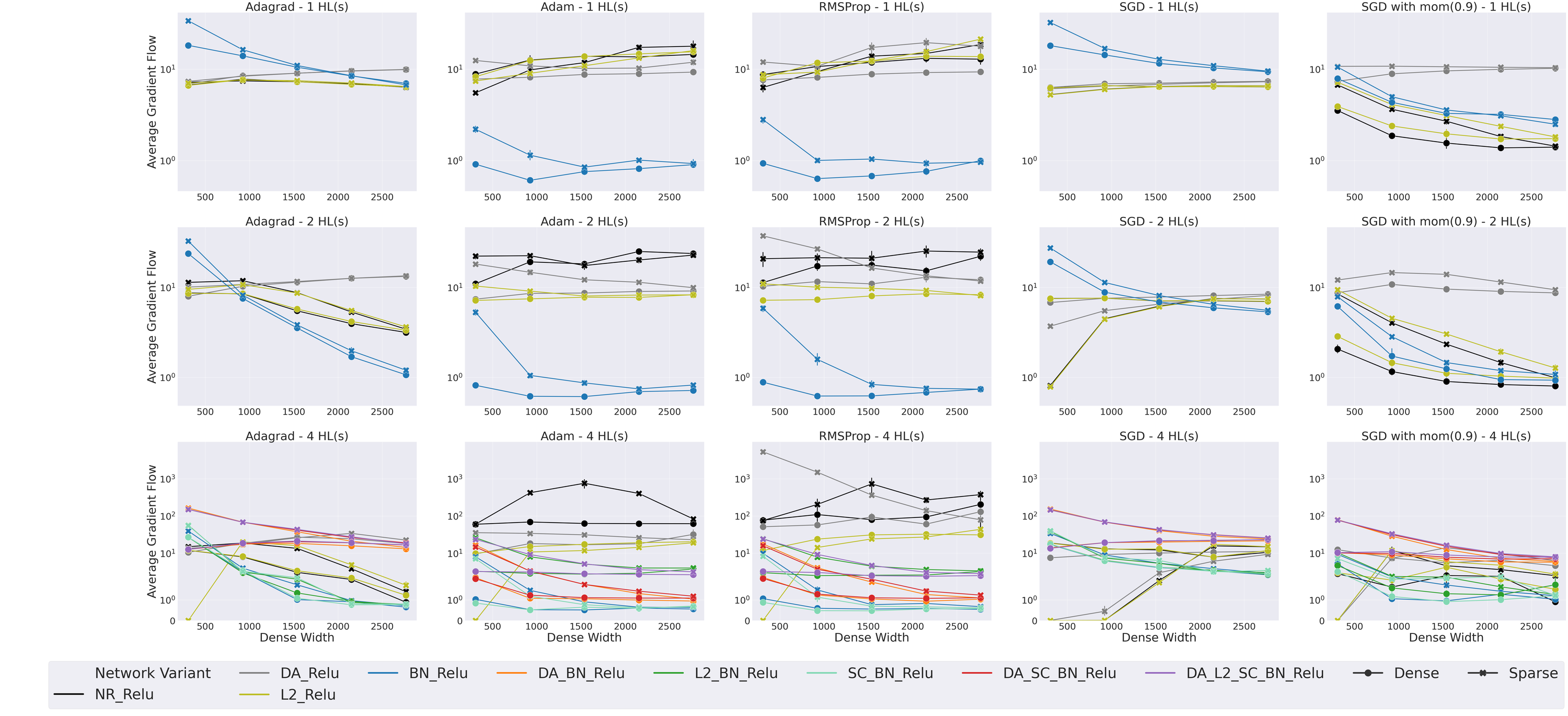}
  \end{center}
\end{figure}

\begin{figure}[ht]
  \caption{Gradient Flow in CIFAR-100 using $EGF_1$, with high learning rate (0.1)}
  \label{fig:EGF1_0.1}
  \begin{center}
    \includegraphics[width=0.8\textwidth]{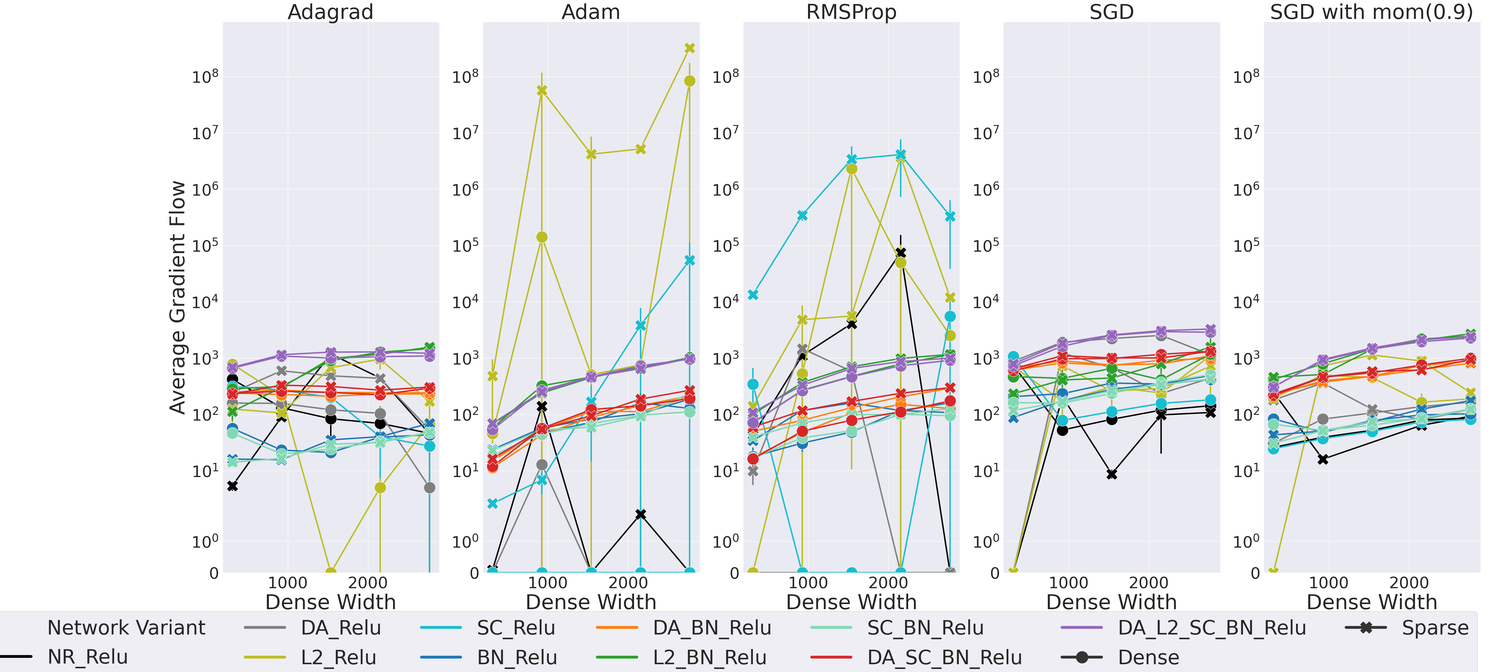}
  \end{center}
\end{figure}

\begin{figure}[ht]
  \caption{Gradient Flow in CIFAR-100 using $|| \vg ||_1$, with high learning rate (0.1)}
  \label{fig:g1_0.1}
  \begin{center}
    \includegraphics[width=0.8\textwidth]{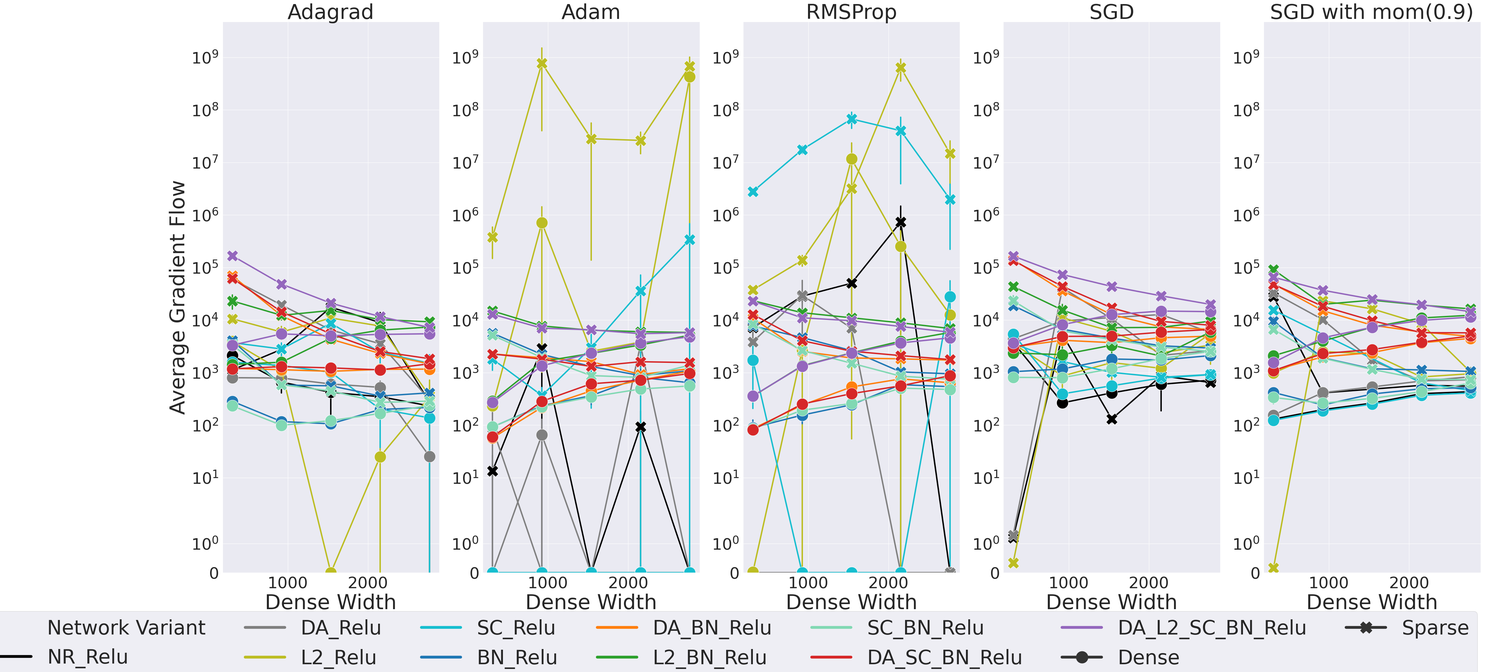}
  \end{center}
\end{figure}

\begin{figure}[ht]
  \caption{Gradient Flow in CIFAR-100 using $|| \vg ||_2$ , with high learning rate (0.1)}
  \label{fig:g2_0.1}
  \begin{center}
    \includegraphics[width=0.8\textwidth]{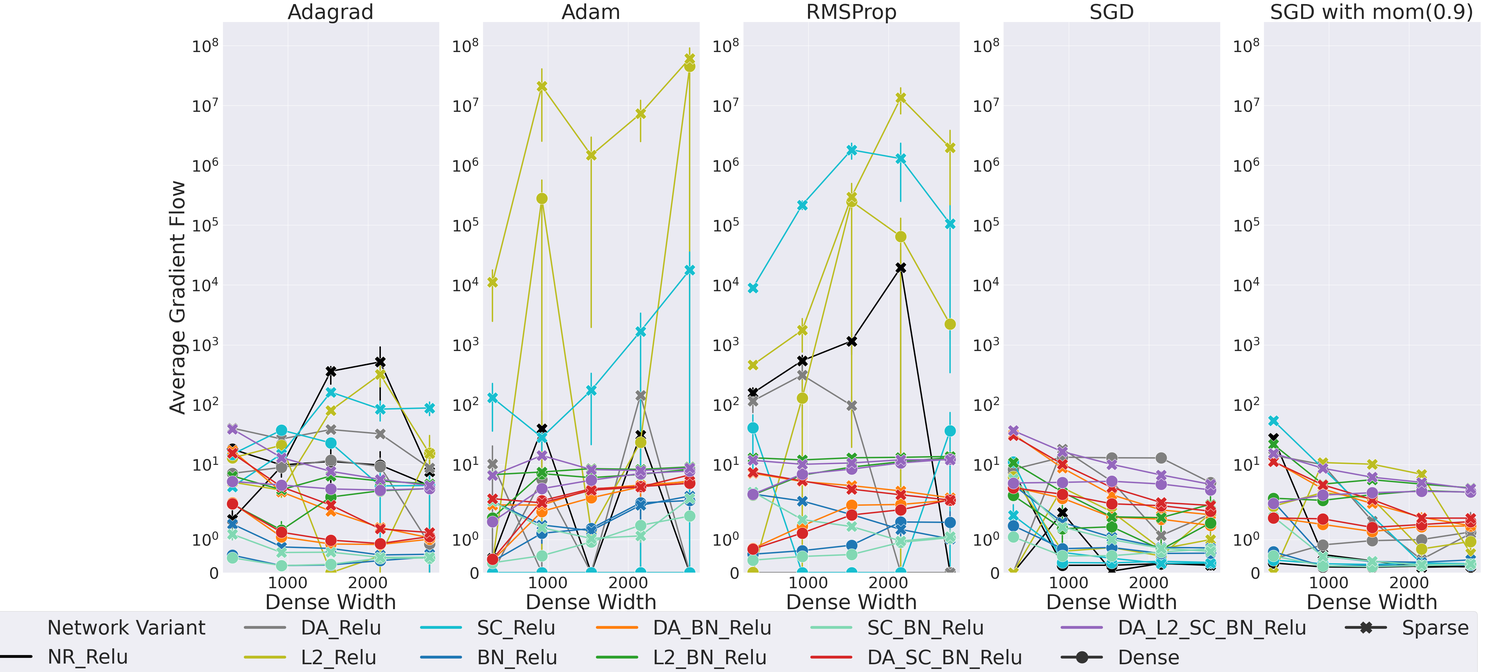}
  \end{center}
\end{figure}

In the main body of the paper, we use $EGF_2$ (Equation \ref{eq:EGF}) as our notion of gradient flow for its comparative benefits with other measures of gradient flow. In this section, for completeness, we present the full set of results using the different formulations of gradient flow on CIFAR-100. Namely, we show $|| \vg ||_1$ (Equation \ref{eq:g}) (Figure \ref{fig:g1_0.001} and  \ref{fig:g1_0.1}) ,$|| \vg ||_2$ (Equation \ref{eq:g})  (Figure \ref{fig:g2_0.001} and \ref{fig:g2_0.1}) and $EGF_1$ (Equation \ref{eq:EGF})  (Figure \ref{fig:EGF1_0.001} and \ref{fig:EGF1_0.1}).

\section{Adaptive Methods} \label{app_sec:adap}
Not all adaptive methods are created equally. We briefly show the different update rules of the adaptive learning rate methods we use in our experiments. In Equations \ref{eq:adap}, we highlight that Adam and RMSProp have the same second moment estimate \highlight{\mathbf{v}_{t}}, which uses
an exponential weighted moving average (EWMA) of past squared gradients to obtain an estimate of the variance of the gradient, which differs from Adagrad's \highlighttwo{\mathbf{v}_{t}}.

\begin{equation}\label{eq:adap}
  \begin{array}{lll}
    \text{Adagrad}                                                                                     & \qquad \text{RMSProp} & \qquad \text{Adam}                                                                                      \\
    \highlighttwo{\mathbf{v}_{t} =\mathbf{v}_{t-1}+\mathbf{g}_{t}^{2}}                                 & \qquad
    \highlight{\mathbf{v}_{t} =\gamma \mathbf{v}_{t-1}+(1-\gamma) \mathbf{g}_{t}^{2}}                  & \qquad
    \mathbf{m}_{t} =\beta_{1} \mathbf{m}_{t-1}+\left(1-\beta_{1}\right) \mathbf{g}_{t}                                                                                                                                                   \\

    \mathbf{w}_{t} =\mathbf{w}_{t-1}-\dfrac{\eta}{\sqrt{\mathbf{v}_{t}+\epsilon}} \odot \mathbf{g}_{t} & \qquad
    \mathbf{w}_{t} =\mathbf{w}_{t-1}-\dfrac{\eta}{\sqrt{\mathbf{v}_{t}+\epsilon}} \odot \mathbf{g}_{t} & \qquad
    \highlight{\mathbf{v}_{t} =\beta_{2} \mathbf{v}_{t-1}+\left(1-\beta_{2}\right) \mathbf{g}_{t}^{2}}                                                                                                                                   \\

                                                                                                       & \qquad                & \qquad \mathbf{m}_{t} = \dfrac{\mathbf{m}_{t}}{1-\beta_{1}}                                             \\
                                                                                                       & \qquad                & \qquad   \mathbf{v}_{t} = \dfrac{\mathbf{v}_{t}}{1-\beta_{2}}                                           \\
                                                                                                       & \qquad                & \qquad  \mathbf{g}_{t}^{\prime}=\dfrac{\eta \hat{\mathbf{m}}_{t}}{\sqrt{\hat{\mathbf{v}}_{t}}+\epsilon} \\
                                                                                                       & \qquad                & \qquad \mathbf{w}_{t} =\mathbf{w}_{t-1}-\mathbf{g}_{t}^{\prime}                                         \\
  \end{array}
\end{equation}
where $\displaystyle \mathbf{g}_{t}$ are the gradients of a network,  $\displaystyle \mathbf{m}_{t}$ is the estimate of the first moment of the gradient, $\displaystyle \mathbf{v}_{t}$ is the estimate of the second moment of the gradient, $\displaystyle \eta$ is the learning rate, $\displaystyle \beta_{1}$,$\displaystyle \beta_{2}$,$\displaystyle \gamma$ are weighting parameters.

This difference between EWMA optimizers (Adam and RMSProp) and Adagrad is two-fold:
\begin{enumerate}
  \item \label{diff_weighting} Adagrad's \highlighttwo{\mathbf{v}_{t}} is evenly weighted between the current squared gradient $\mathbf{{g}_{t}^{2}}$ and the previous value of $\mathbf{v}_{t}$. Adam and RMSProp's \highlight{\mathbf{v}_{t}} use $0.9$ \citep{hinton2012neural} or $0.999$ \citep{kingma2014adam} as a weighting parameter for the past $\mathbf{v}_{t}$, which weights past $\mathbf{v}_{t-1}$ more than $\mathbf{g}_{t}$.
  \item \label{higher_lr} Adagrad's \highlighttwo{\mathbf{v}_{t}} grows almost linearly over time \citep{zhang2019dive}, which results in fast decline in learning rate (a higher $\mathbf{v}_{t}$ results in a smaller effective learning rate since $\eta$ is divided by $\mathbf{v}_{t}$). Since Adam and RMSProp's \highlight{\mathbf{v}_{t}} is multiplied by a weighting factor, this value decreases over time. This occurs even if we choose a weighting factor of $0.5$, since both $\mathbf{v}_{t}$ and $\mathbf{g}_{t}$ will be multiplied by $0.5$. A smaller \highlight{\mathbf{v}_{t}}, results in a higher effective learning rate, meaning learning rates stay higher for longer.
\end{enumerate}

We believe the higher effective learning rate (point \ref{higher_lr}) is why Adam and RMSProp behave similarly in some contexts. Furthermore, it has been noted that at times, when the \highlight{\mathbf{v}_{t}} estimate blows up, it can cause optimizers to fail to converge \citep{zaheer2018adaptive,zhang2019dive}.
\highlight{\mathbf{v}_{t}} is also present in the formulation of Adadelta \citep{zeiler2012adadelta}.

\section{Activation Functions}  \label{app_sec:activations}
\begin{figure}[hb]
  \centering
  \begin{subfigure}{0.475\textwidth} \centering
    \caption{Activation Function with inputs [-5,5]}\label{fig:act}
    \includegraphics[width=\textwidth]{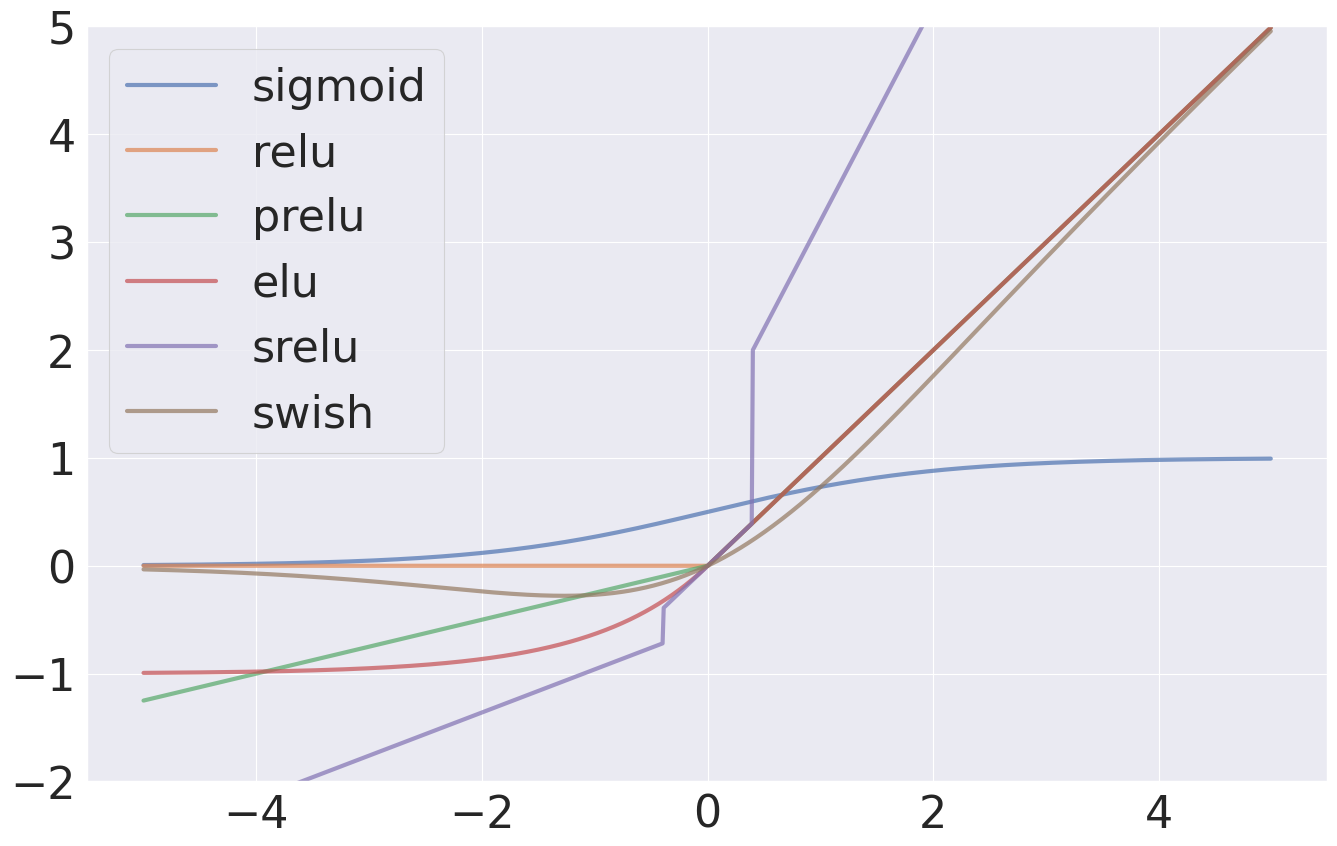}
  \end{subfigure}
  \hfill
  \begin{subfigure}{0.475\textwidth} \centering
    \caption{Derivative of Activation Function with inputs [-5,5]}\label{fig:act_grad}
    \includegraphics[width=\textwidth]{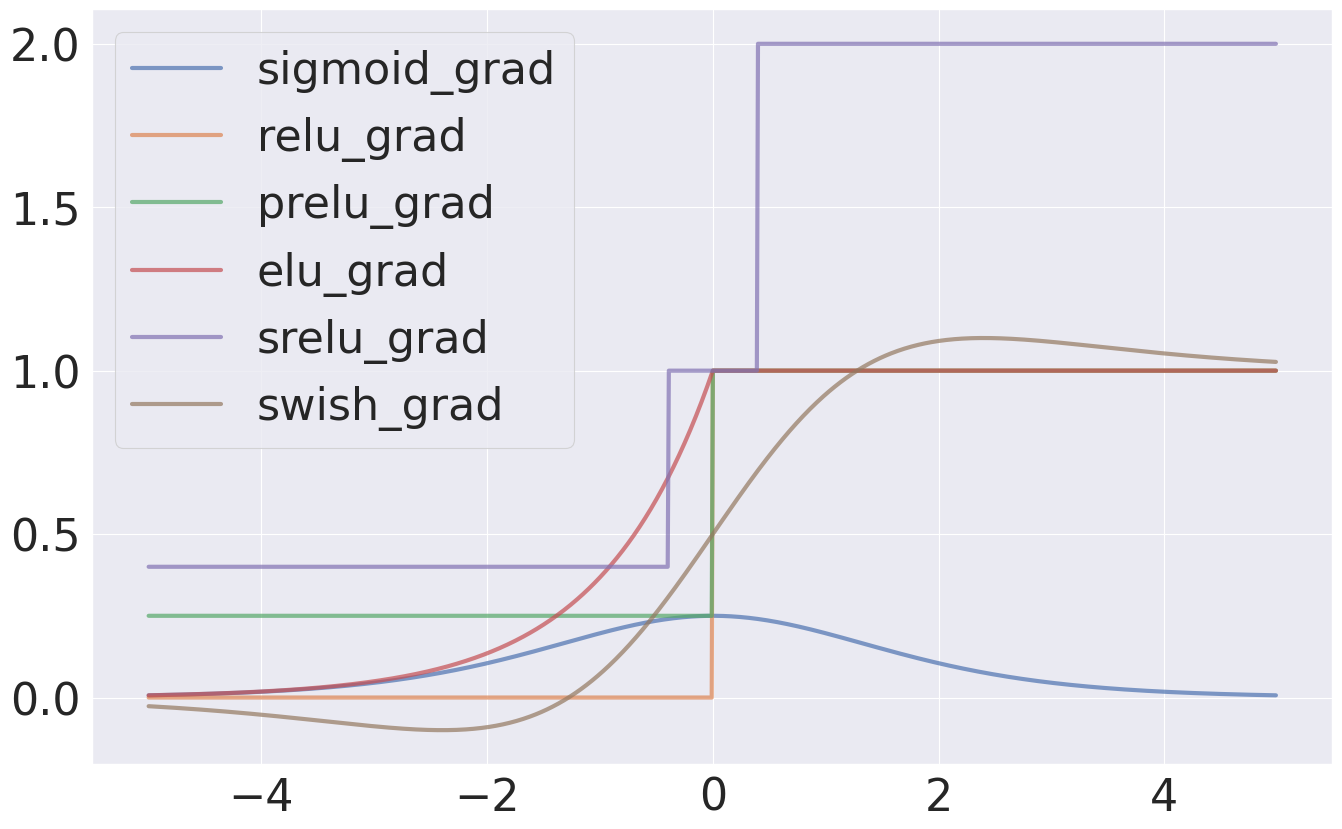}
  \end{subfigure}
  \vskip\baselineskip
  \caption{\textbf{Plot of Activation Functions.} We plot the activation functions we used and their derivatives. We see that Swish is the only activation function that allows for the flow of negative gradients, due its non-monotonicity.}
\end{figure}

We plot the activation functions we used (Figure \ref{fig:act}) and their gradients (Figure \ref{fig:act_grad}).

\section{Detailed Results for \texttt{SC-SDC}}\label{appen:detailed_sdsdc}

In this section, we present the detailed results for our experiments. For the tables, we use the following colour scale, depending on the p-value from the Wilcoxon Signed Rank Test.

Colour Scale based on p-values:
\newcolumntype{C}{>{\centering\arraybackslash}p{0.75em}}
\begin{tabular}{*{11}{C}}
\multicolumn{1}{r}{\cellcolor[HTML]{57BB8A}} & \multicolumn{1}{r}{\cellcolor[HTML]{78C083}} & \multicolumn{1}{r}{\cellcolor[HTML]{9AC57C}} & \multicolumn{1}{r}{\cellcolor[HTML]{BBCB75}} & \multicolumn{1}{r}{\cellcolor[HTML]{DDD06E}} & \multicolumn{1}{r}{\cellcolor[HTML]{FFD666}} & \multicolumn{1}{r}{\cellcolor[HTML]{FBC568}} & \multicolumn{1}{r}{\cellcolor[HTML]{F6B36B}} & \multicolumn{1}{r}{\cellcolor[HTML]{F0A06D}} & \multicolumn{1}{r}{\cellcolor[HTML]{EB8E70}} & \multicolumn{1}{r}{\cellcolor[HTML]{E67C73}} \\
0  &  &  &  &   & .5 &  &  &  &   & 1  \\
\multicolumn{2}{c}{S  $>$ D} & & & & & & & & \multicolumn{2}{c}{S $\leq$ D} \\ 
\end{tabular}

where $S$ refers to sparse networks and $D$ refers to dense networks.

\subsection{Detailed Results}
\begin{figure}[ht]
  \caption{Effect of Regularization on Accuracy and Gradient Flow for Dense and Sparse Networks, with Four Hidden Layers on FMNIST, with low learning rate (0.001)}
  \label{fig:fmnist_diff_reg_all_optims_low_lr}
  \begin{subfigure}{\linewidth}\centering
    \caption{Test Accuracy for Dense and Sparse Networks on FMNIST}\label{fig:fmnist_diff_reg_acc_low_lr_acc}
    \includegraphics[width=1\linewidth]{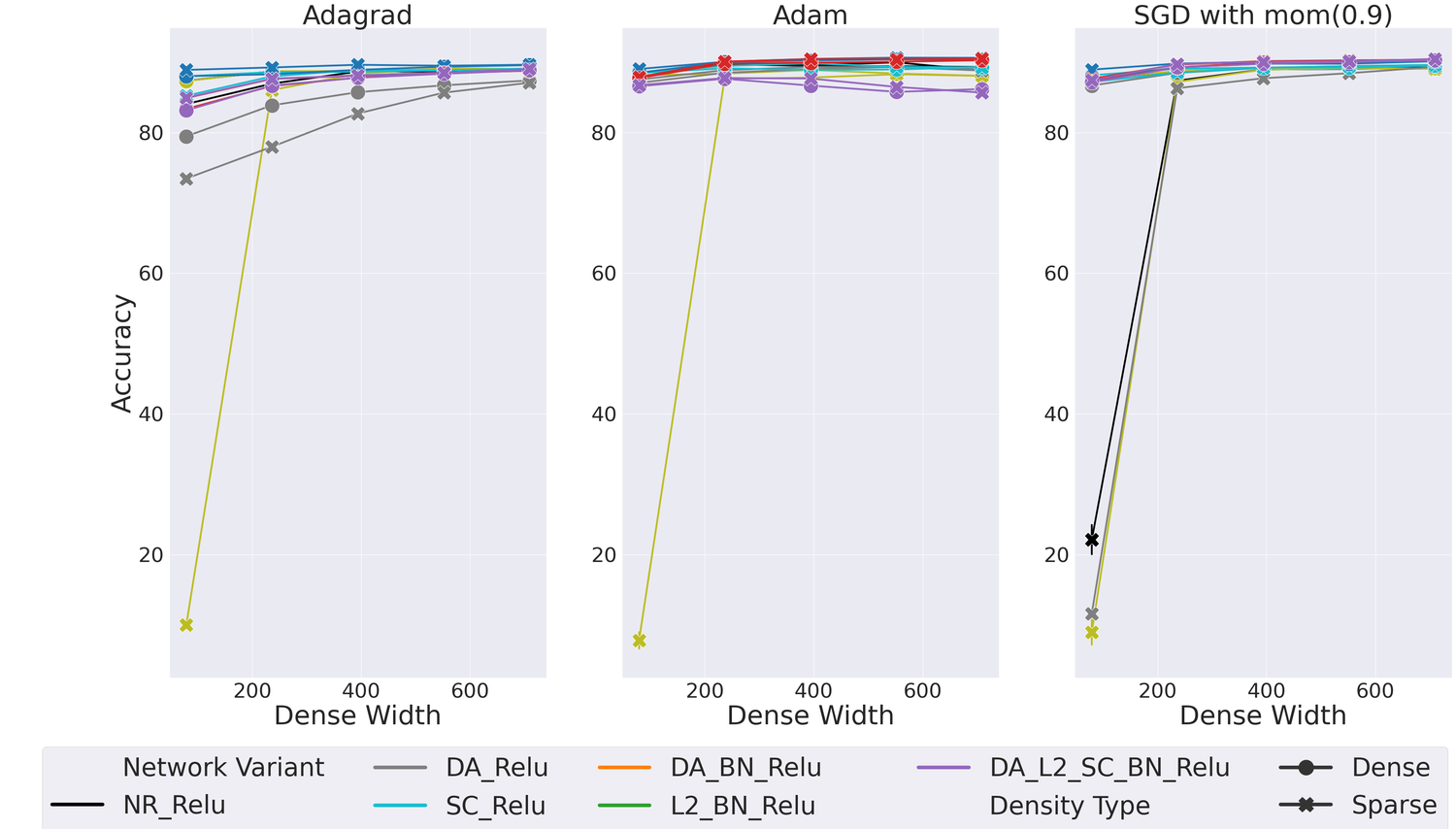}
  \end{subfigure}
  \begin{subfigure}{\linewidth}\centering
    \begin{center}
      \caption{Gradient Flow for Dense and Sparse Networks on FMNIST}\label{fig:fmnist_diff_reg_acc_low_lr_grad_flow}
      \includegraphics[width=1\linewidth]{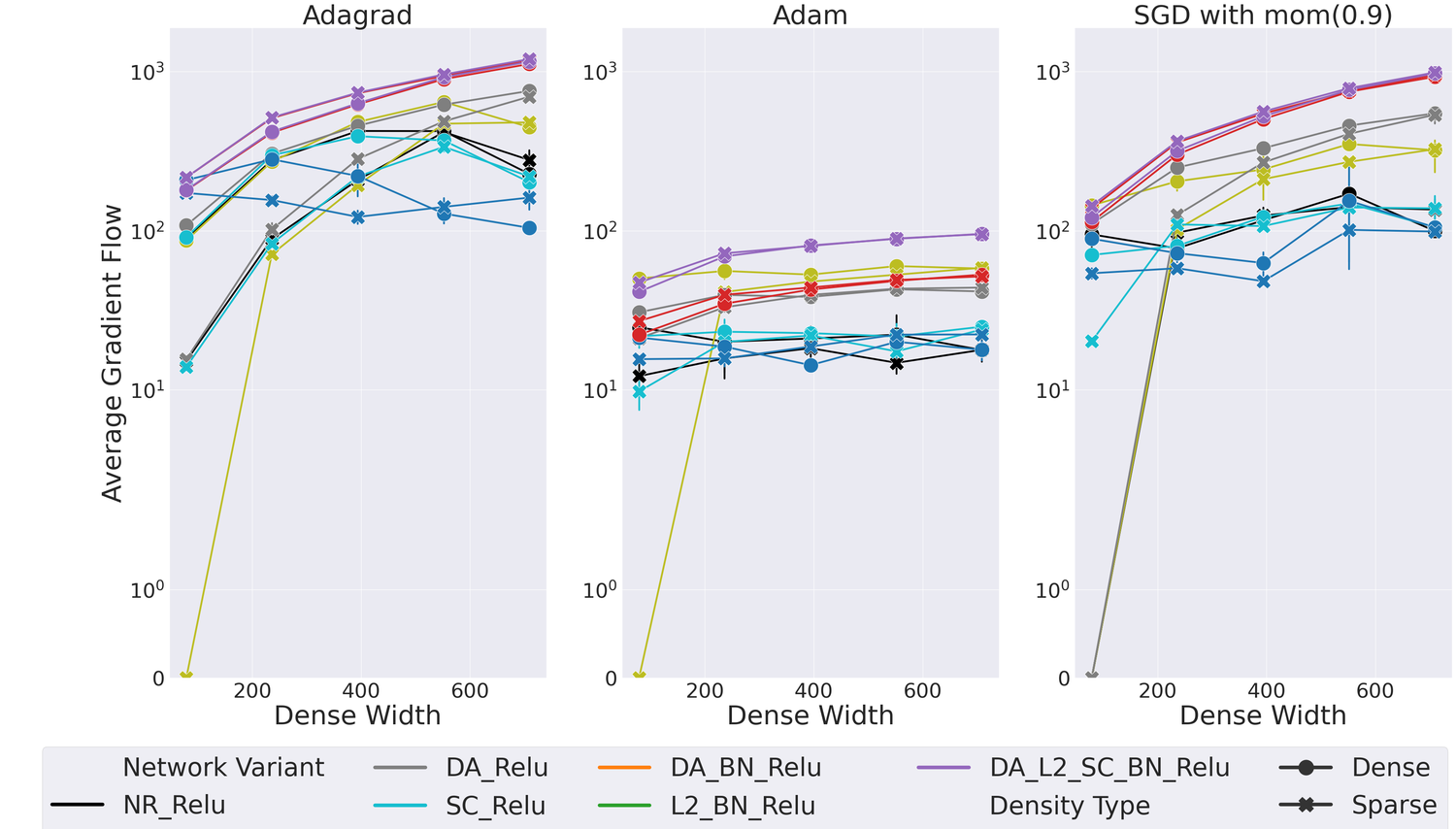}
    \end{center}
  \end{subfigure}
\end{figure}

\begin{figure}[ht]
  \caption{Effect of Regularization on Accuracy and Gradient Flow for Dense and Sparse Networks, on CIFAR-10, with low learning rate (0.001)}
  \label{fig:c10_diff_reg_all_optims_low_lr}
  \begin{subtable}{\textwidth}\centering
    \begin{center}
      \caption{Test Accuracy for Dense and Sparse Networks}\label{fig:c10_diff_reg_acc_low_lr_acc}
      \includegraphics[width=0.8\textwidth]{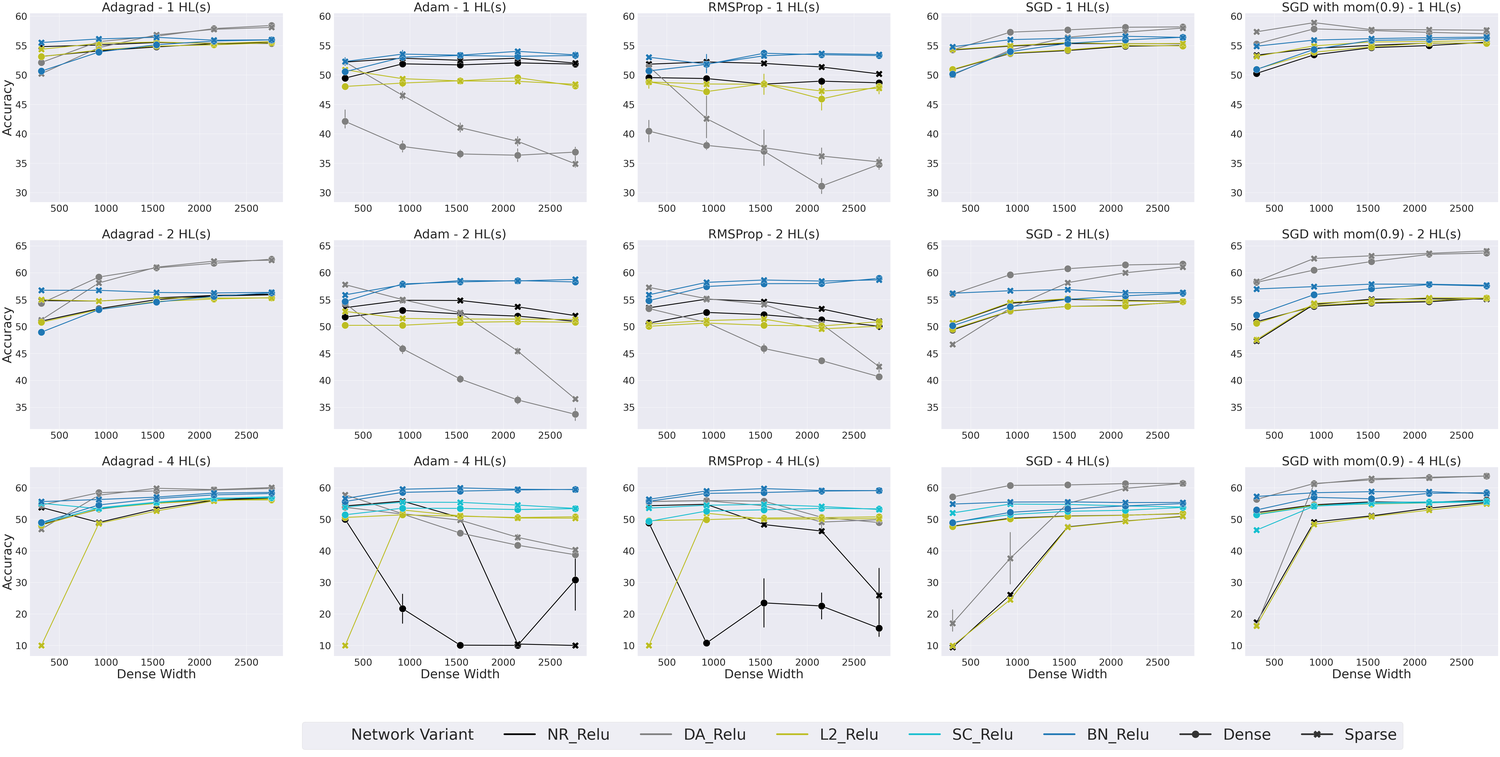}
    \end{center}
    \begin{center}
      \caption{Gradient Flow for Dense and Sparse Networks}\label{fig:c10_diff_reg_acc_low_lr_grad_flow}
      \includegraphics[width=0.8\textwidth]{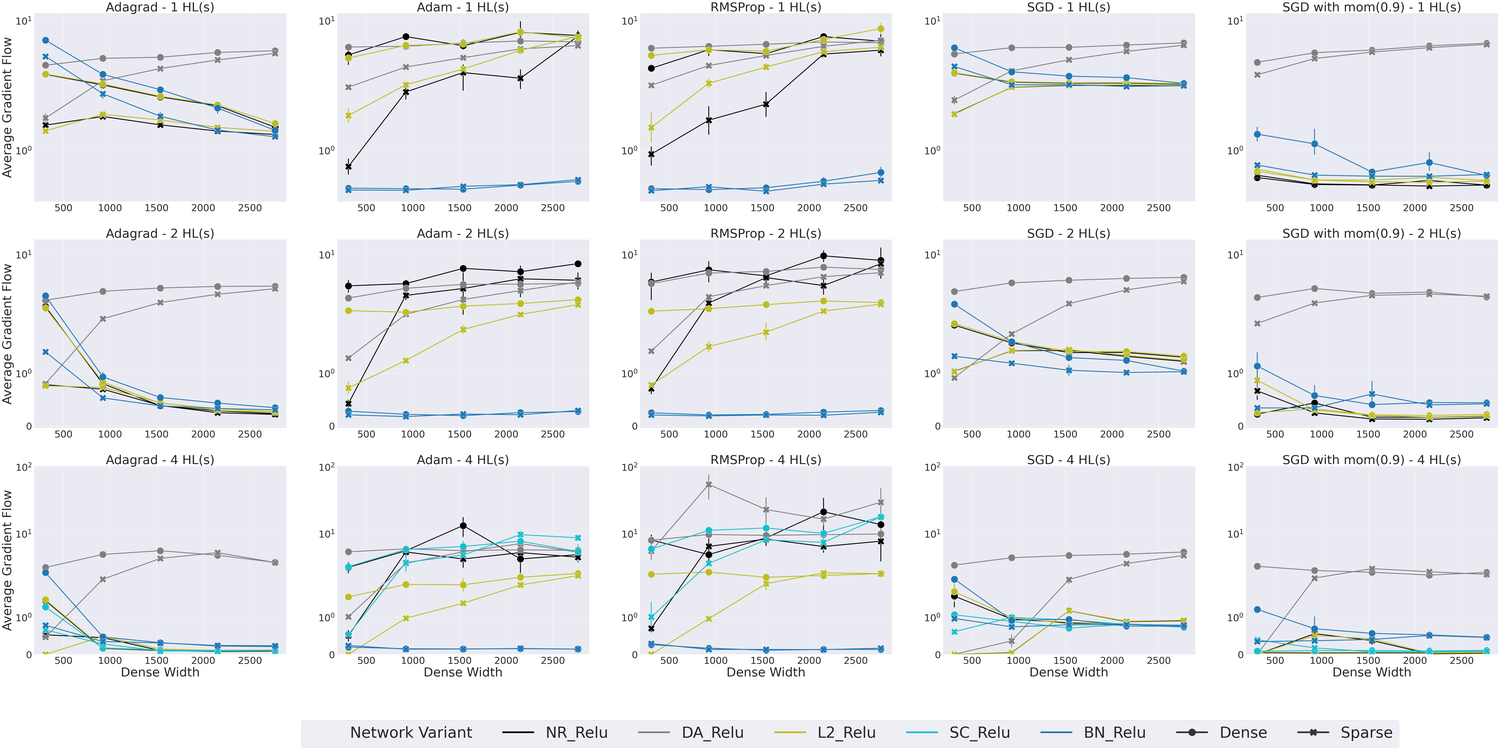}
    \end{center}
  \end{subtable}
\end{figure}

\begin{figure}[ht]
  \caption{Effect of Regularization on Accuracy and Gradient Flow for Dense and Sparse Networks, with Four Hidden Layers on CIFAR-10, with high learning rate (0.1)}
  \label{fig:c10_diff_reg_all_optims_high_lr}
  \begin{subtable}{\textwidth}\centering
    \begin{center}
      \caption{Test Accuracy for Dense and Sparse Networks}\label{fig:c10_diff_reg_acc_high_lr_acc}
      \includegraphics[width=1\textwidth]{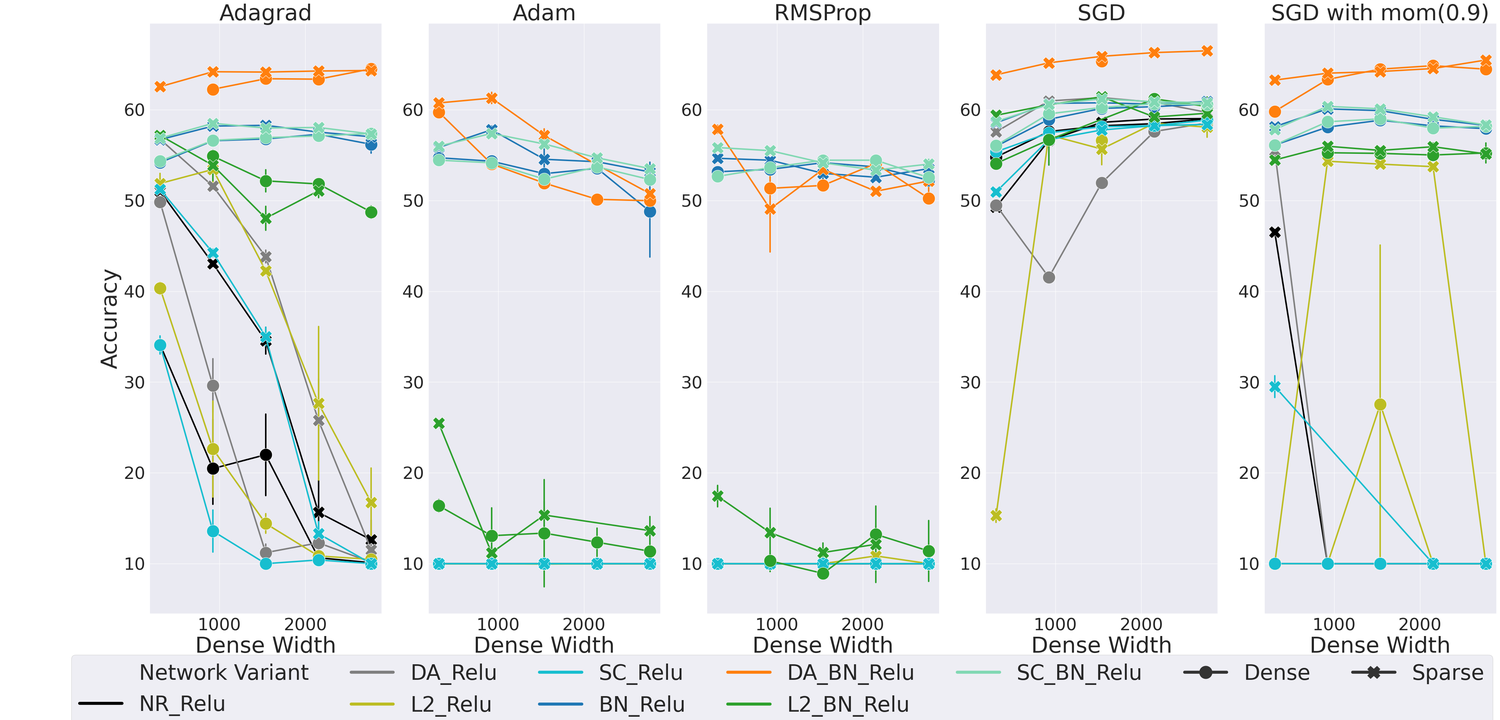}
    \end{center}
    \begin{center}
      \caption{Gradient Flow for Dense and Sparse Networks}\label{fig:c10_diff_reg_acc_high_lr_grad_flow}
      \includegraphics[width=1\textwidth]{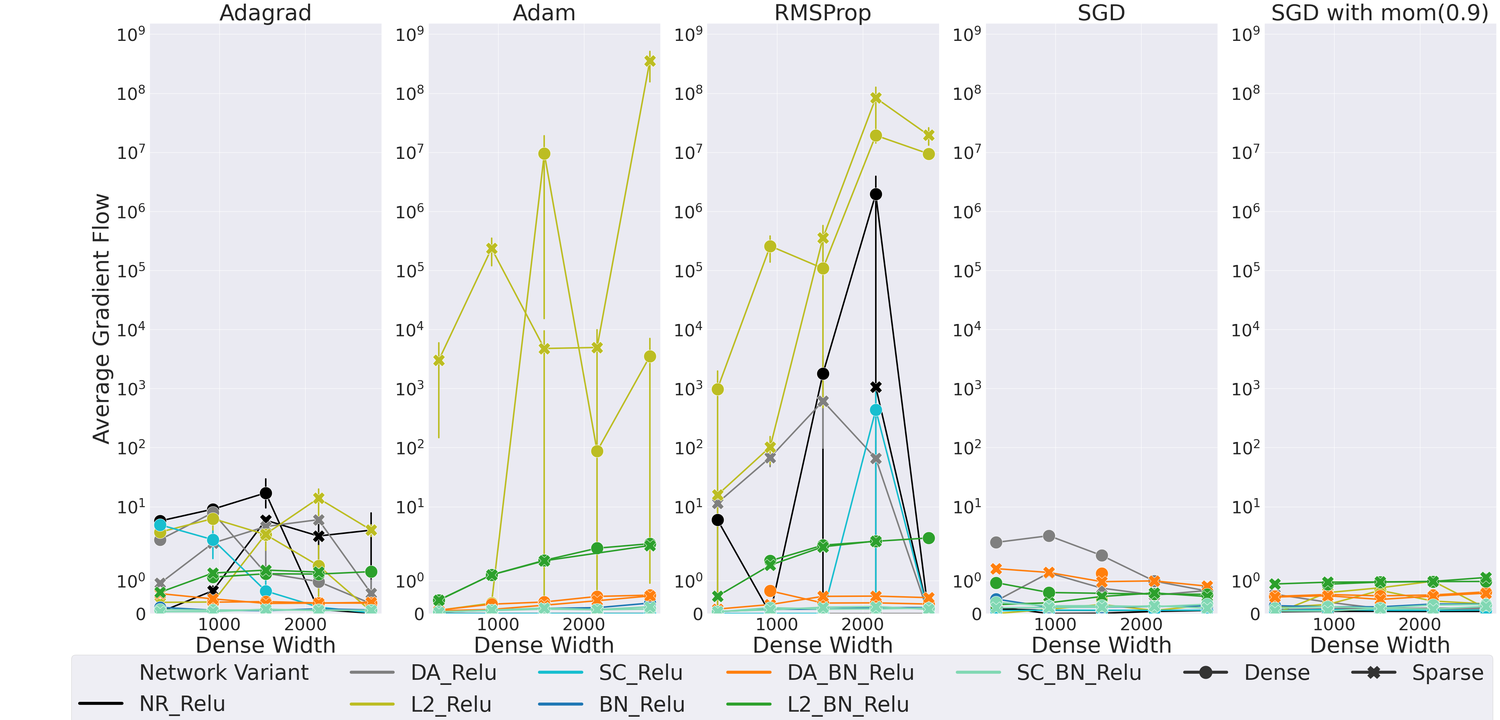}
    \end{center}
  \end{subtable}
\end{figure}

\begin{figure}
  \centering
  \caption{Effect of Activation Functions on Accuracy and Gradient Flow for Dense and Sparse Networks with Four Hidden Layers on CIFAR-10, with low learning rate (0.001)}
  \label{fig:c10_diff_acts_all_optims_low}
  \begin{subfigure}{0.45\textwidth}
    \centering
    \includegraphics[width=.80\textwidth]{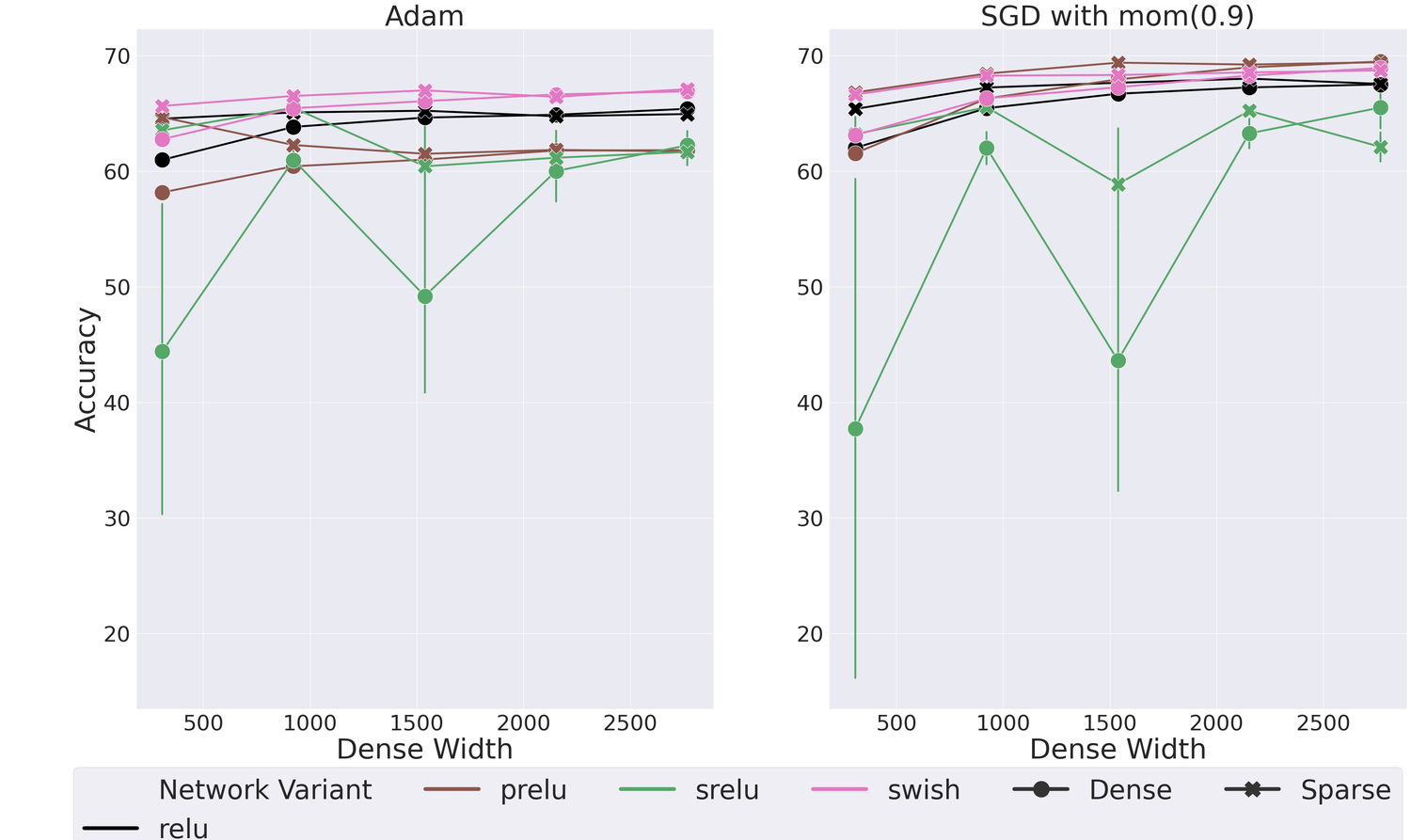}
    \caption{Test Accuracy for Dense and Sparse Networks on CIFAR-10}\label{fig:c10_diff_acts_all_optims_low_acc}
  \end{subfigure}%
  \begin{subfigure}{0.45\textwidth}
    \centering
    \includegraphics[width=.80\textwidth]{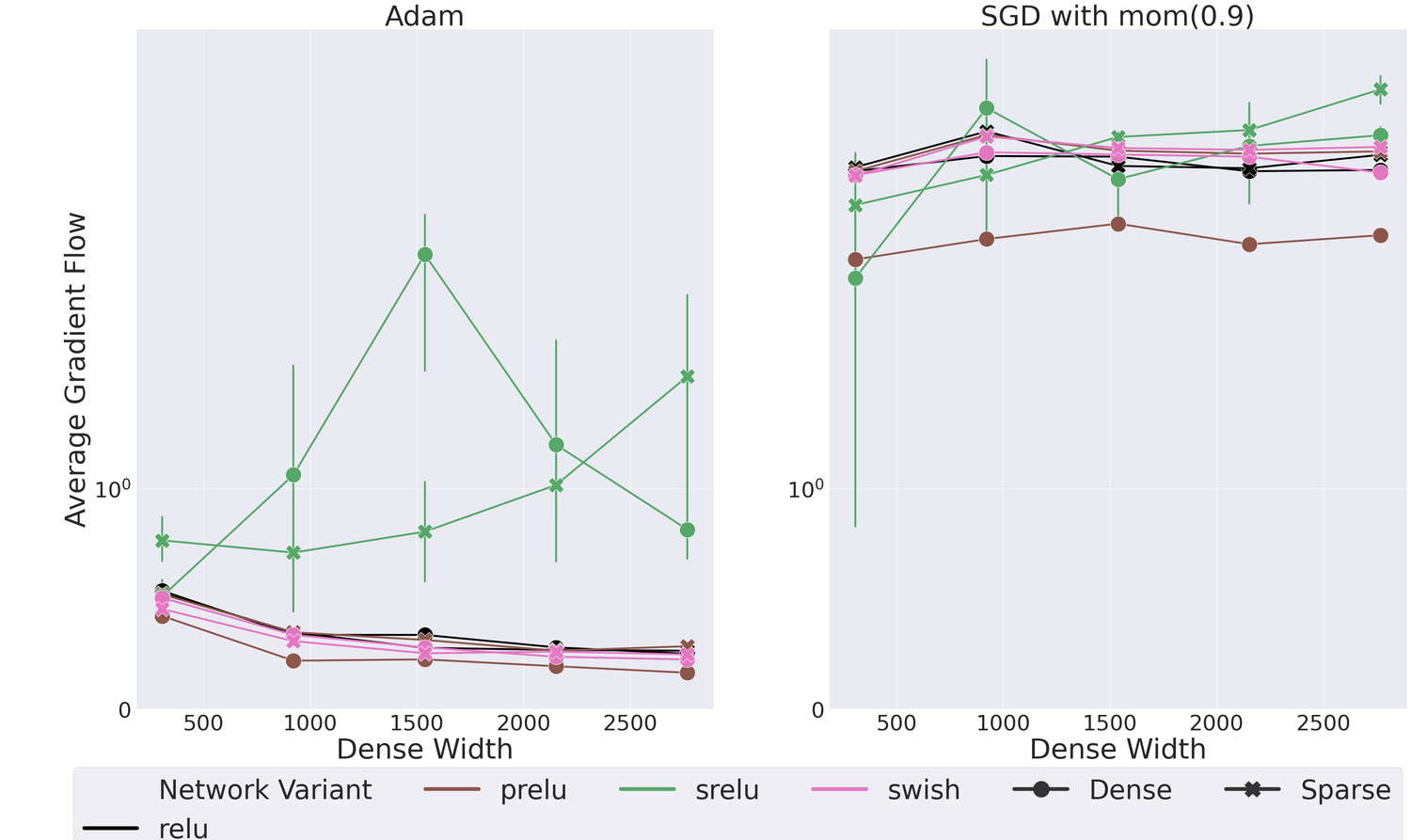}
    \caption{Gradient Flow for Dense and Sparse Networks on CIFAR-10}\label{fig:c10_diff_acts_all_optims_low_grad_flow}
  \end{subfigure}
\end{figure}

\begin{figure}[ht]
  \caption{Effect of Regularization on Accuracy and Gradient Flow for Dense and Sparse Networks with One, Two and Four Hidden Layers on CIFAR-100, with low learning rate (0.001)}
  \label{fig:c100_diff_reg_all_optims_low_lr_all}
  \begin{subtable}{\linewidth}\centering
    \begin{center}
      \caption{Test Accuracy for Dense and Sparse Networks on CIFAR-100}\label{fig:c100_diff_reg_acc_low_lr_all}
      \includegraphics[width=1\linewidth]{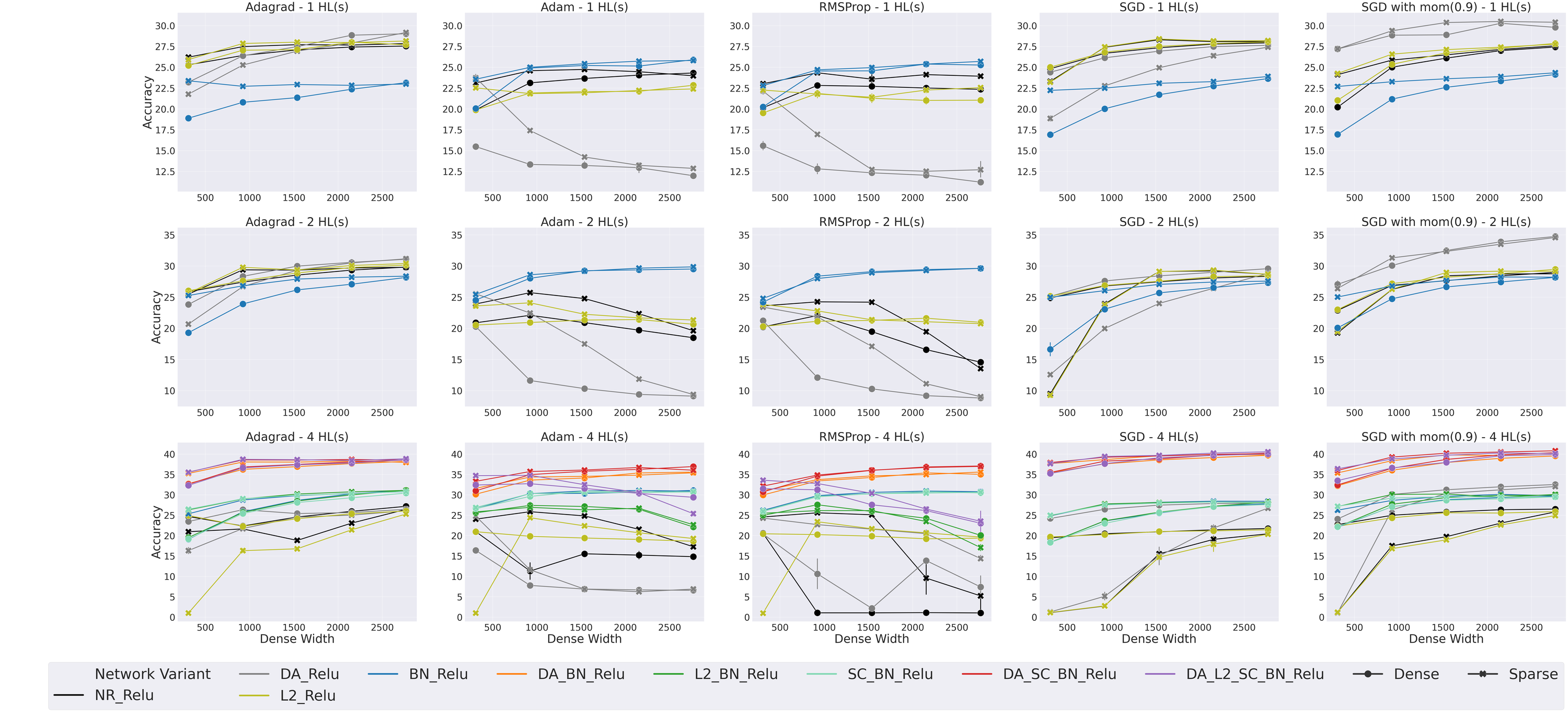}
    \end{center}
    \begin{center}
      \caption{Gradient Flow for Dense and Sparse Networks on CIFAR-100}\label{fig:c100_diff_reg_grad_flow_all}
      \includegraphics[width=1\linewidth]{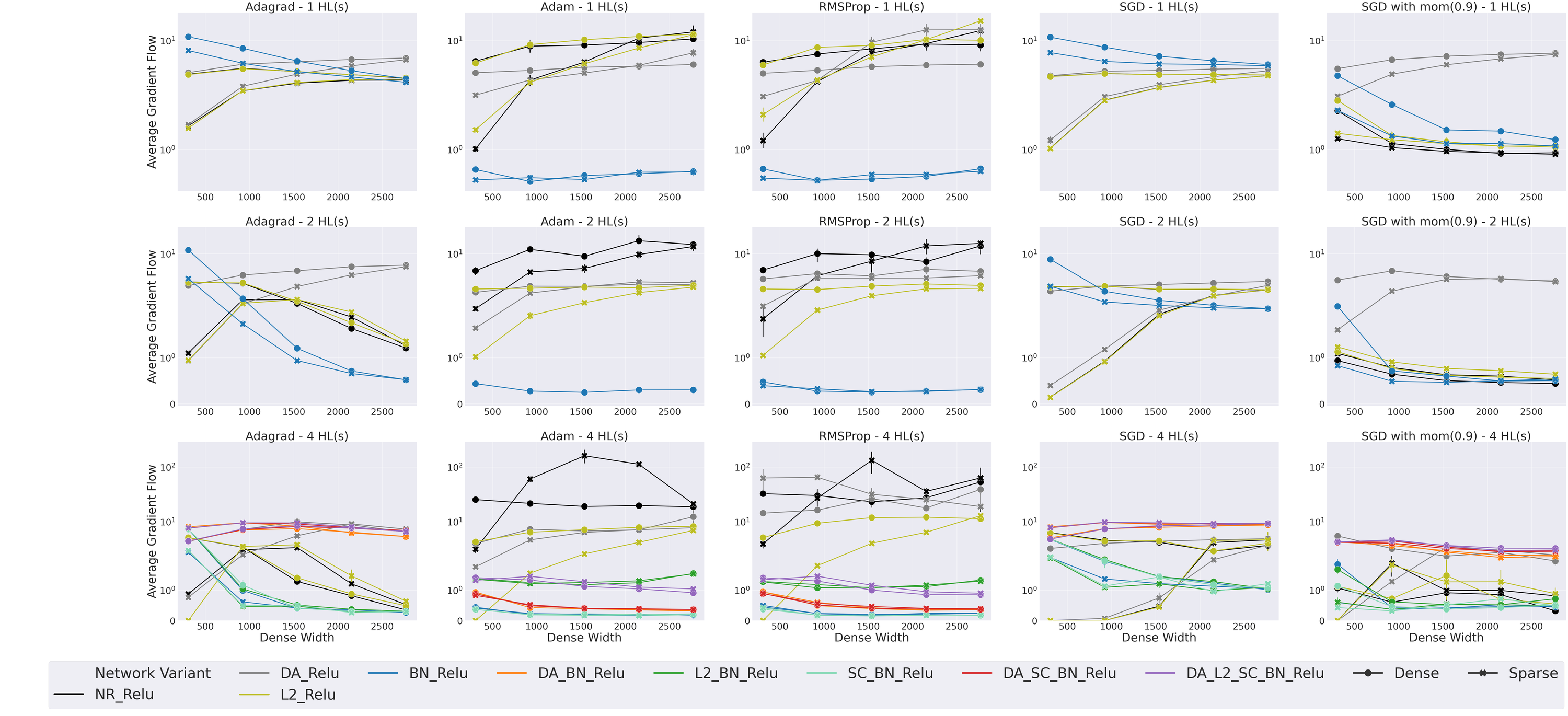}
    \end{center}
  \end{subtable}
\end{figure}

\begin{figure}[ht]
  \caption{Effect of Activation Functions on Accuracy and Gradient Flow for Dense and Sparse Networks with Four Hidden Layers on CIFAR-100, with low learning rate (0.001)}
  \label{fig:c100_diff_reg_all_optims_low_lr_acts}
  \begin{subtable}{\textwidth}\centering
    \begin{center}
      \caption{Test Accuracy for Dense and Sparse Networks on CIFAR-100}\label{fig:c100_diff_reg_acc_low_lr_acts}
      \includegraphics[width=1\textwidth]{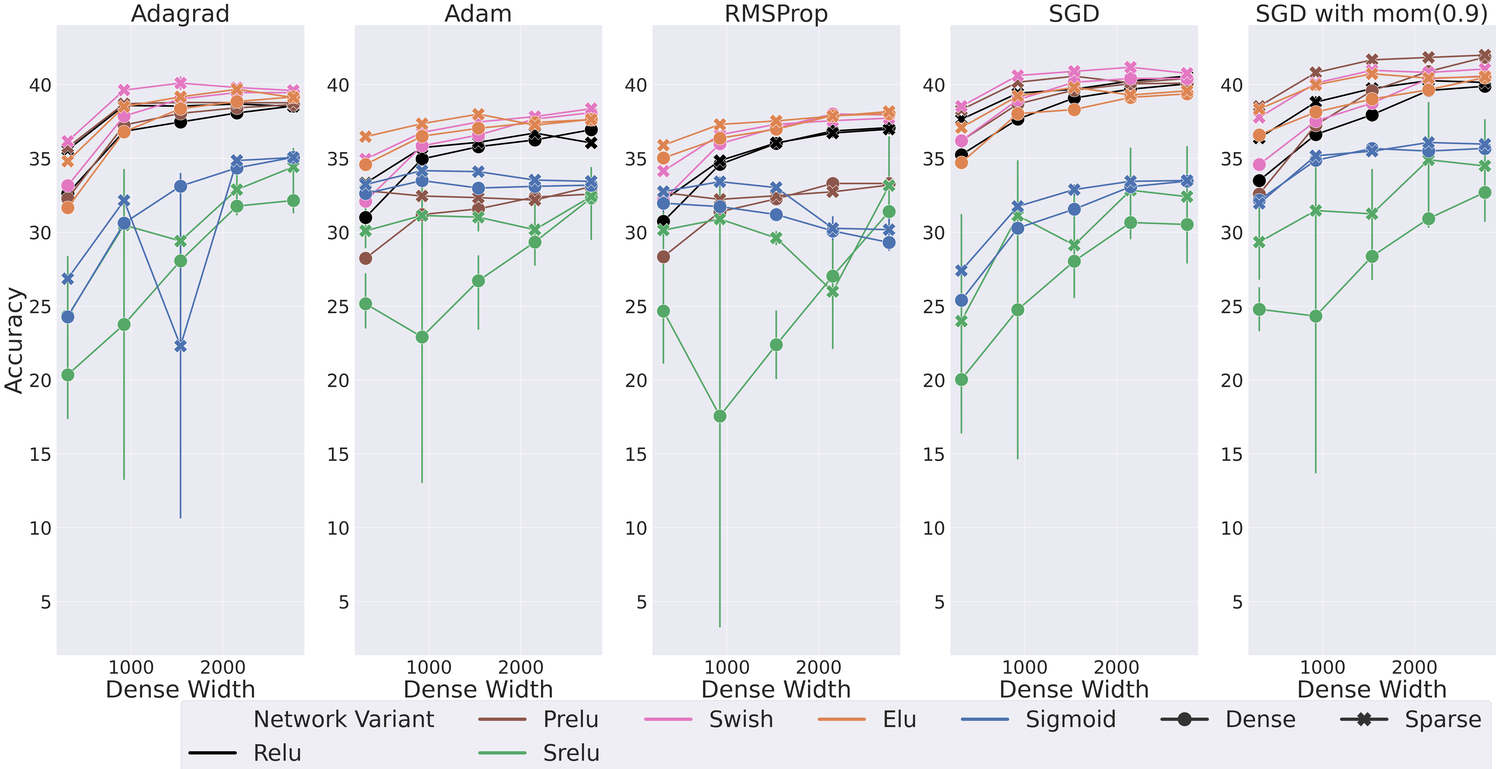}
    \end{center}
    \begin{center}
      \caption{Gradient Flow for Dense and Sparse Networks on CIFAR-100}\label{fig:c100_diff_reg_grad_flow_acts}
      \includegraphics[width=1\textwidth]{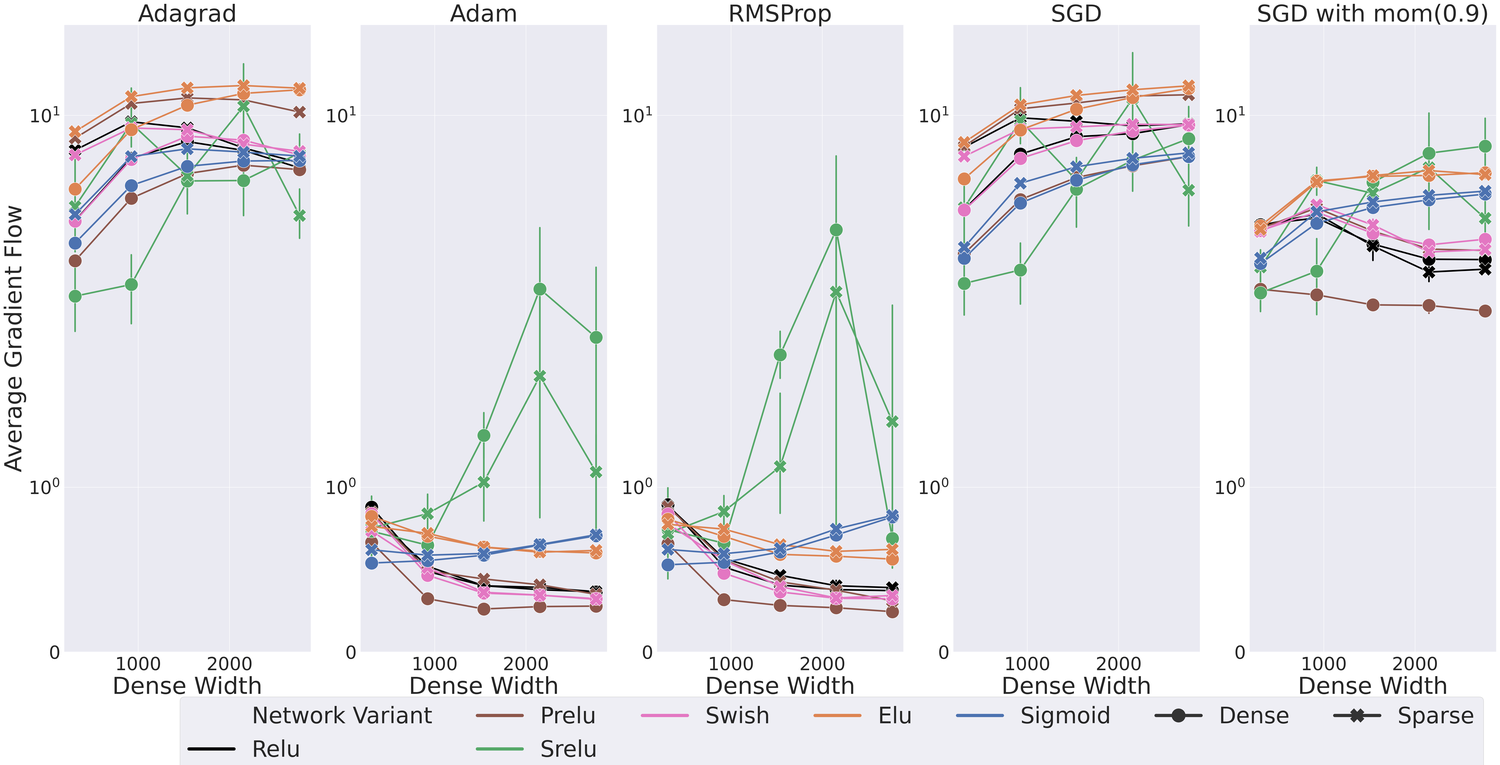}
    \end{center}
  \end{subtable}
\end{figure}

\begin{figure}[ht]
  \caption{Effect of Regularization (\textit{with BatchNorm}) on Accuracy and Gradient Flow for Dense and Sparse Networks, with Four Hidden Layers on CIFAR-100, with high learning rate (0.1)}
  \label{fig:c100_diff_reg_all_optims_high_lr_with_batchnorm}
  \begin{subtable}{\textwidth}\centering
    \begin{center}
      \caption{Test Accuracy for Dense and Sparse Networks on CIFAR-100}\label{fig:c100_diff_reg_acc_high_lr_with_batchnorm}
      \includegraphics[width=1\textwidth]{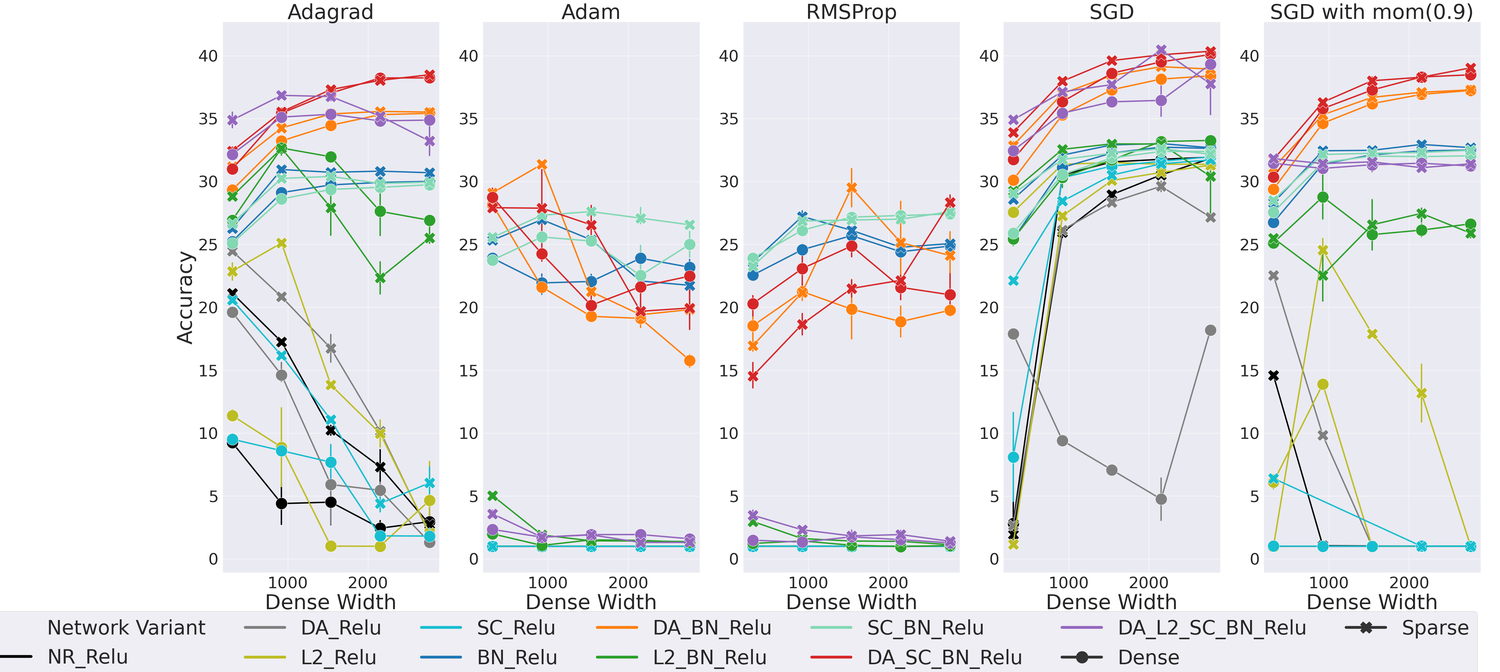}
    \end{center}
    \begin{center}
      \caption{Gradient Flow for Dense and Sparse Networks on CIFAR-100}\label{fig:c100_diff_reg_grad_flow_high_lr_with_batchnorm}
      \includegraphics[width=1\textwidth]{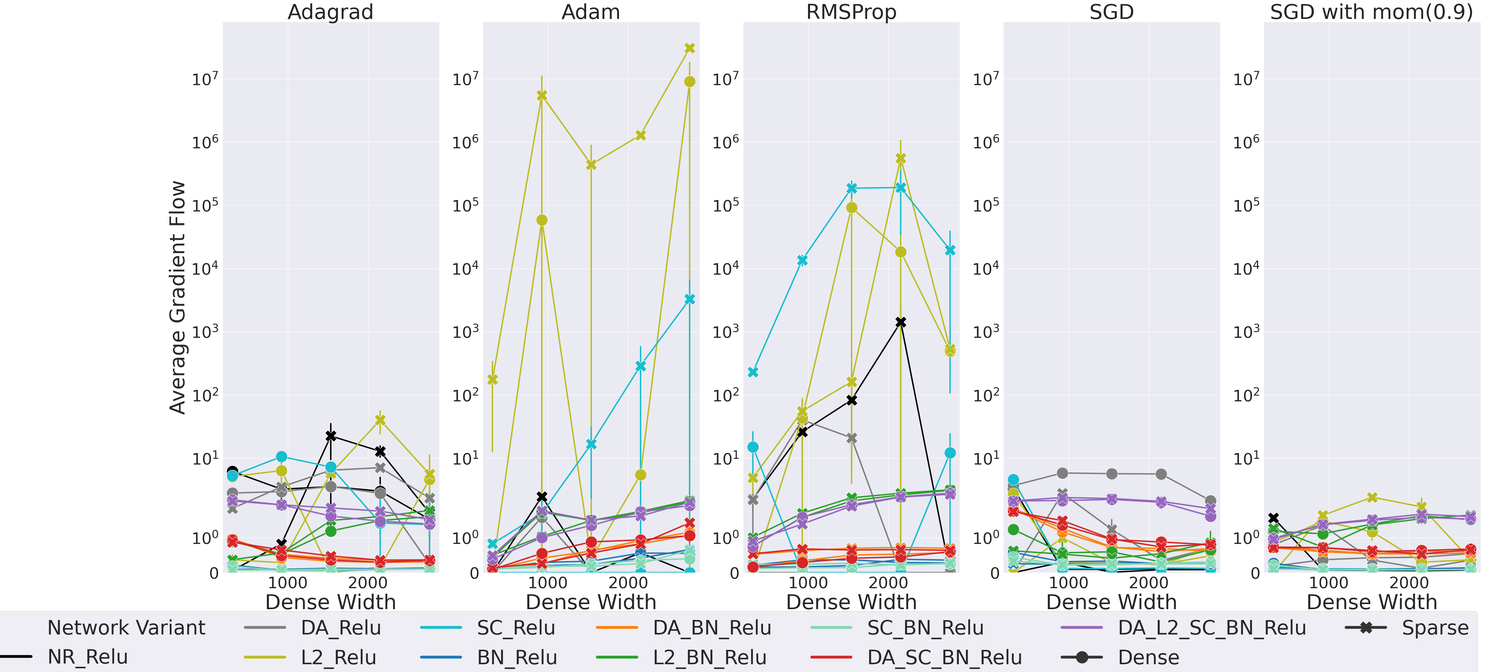}
    \end{center}
  \end{subtable}
\end{figure}

\begin{figure}[ht]
  \caption{Effect of Activation Functions on Accuracy and Gradient Flow for Dense and Sparse Networks with Four Hidden Layers on CIFAR-100, with high learning rate (0.1)}
  \label{fig:c100_diff_reg_all_optims_high_lr_acts}
  \begin{subtable}{\textwidth}\centering
    \begin{center}
      \caption{Test Accuracy for Dense and Sparse Networks on CIFAR-100}\label{fig:c100_diff_reg_acc_high_lr_acts}
      \includegraphics[width=1\textwidth]{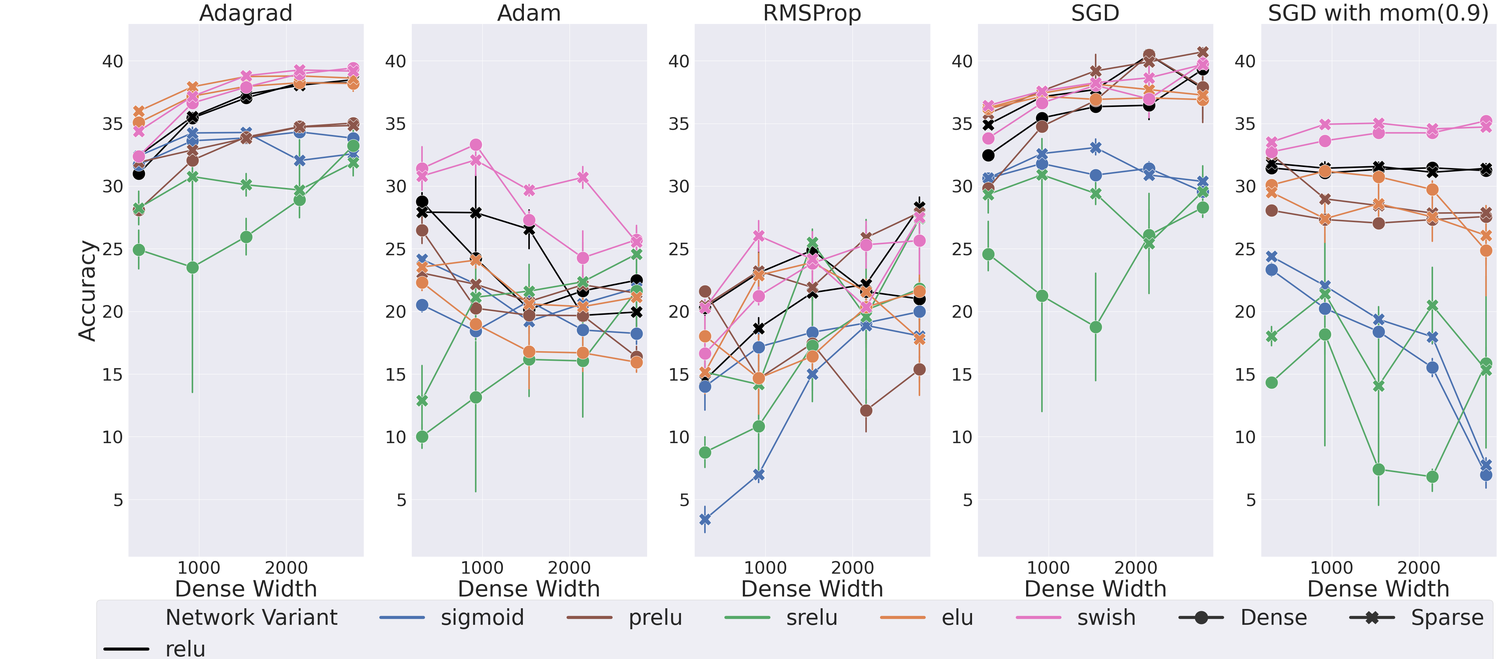}
    \end{center}
    \begin{center}
      \caption{Gradient Flow for Dense and Sparse Networks on CIFAR-100}\label{fig:c100_diff_reg_grad_flow_high_lr_acts}
      \includegraphics[width=1\textwidth]{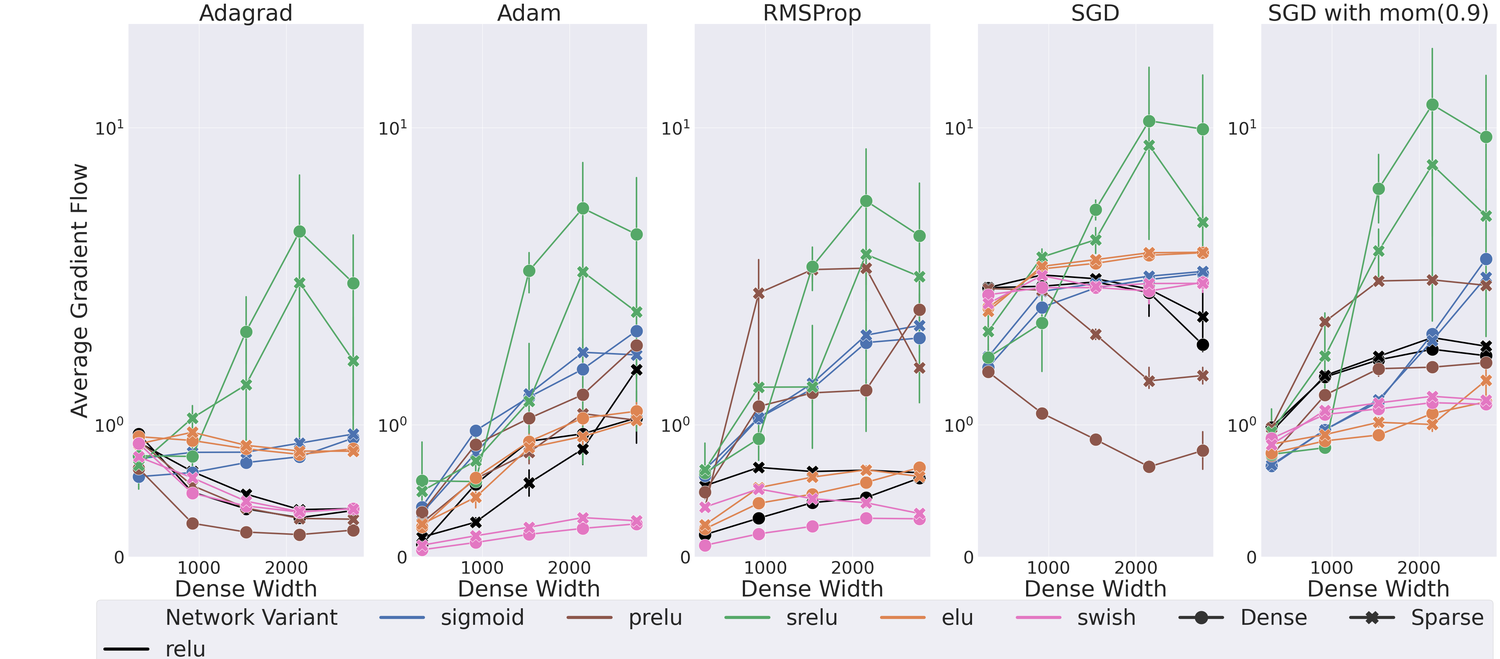}
    \end{center}
  \end{subtable}
\end{figure}

\begin{figure}
  \centering
  \caption{\textbf{Comparison of Adam and AdamW for Dense and Sparse Networks.} We compare the performance of Adam and AdamW for dense and sparse Networks with four hidden layers on CIFAR-100, with a high learning rate (0.1). We see that AdamW's weight decay formulation has a lower \texttt{EGF} than the standard L2 formulation used in Adam and this correlates to better network performance in AdamW}
  \label{fig:c100_adamw}
  \begin{subfigure}{0.45\linewidth}
    \centering
    \includegraphics[width=.8\linewidth]{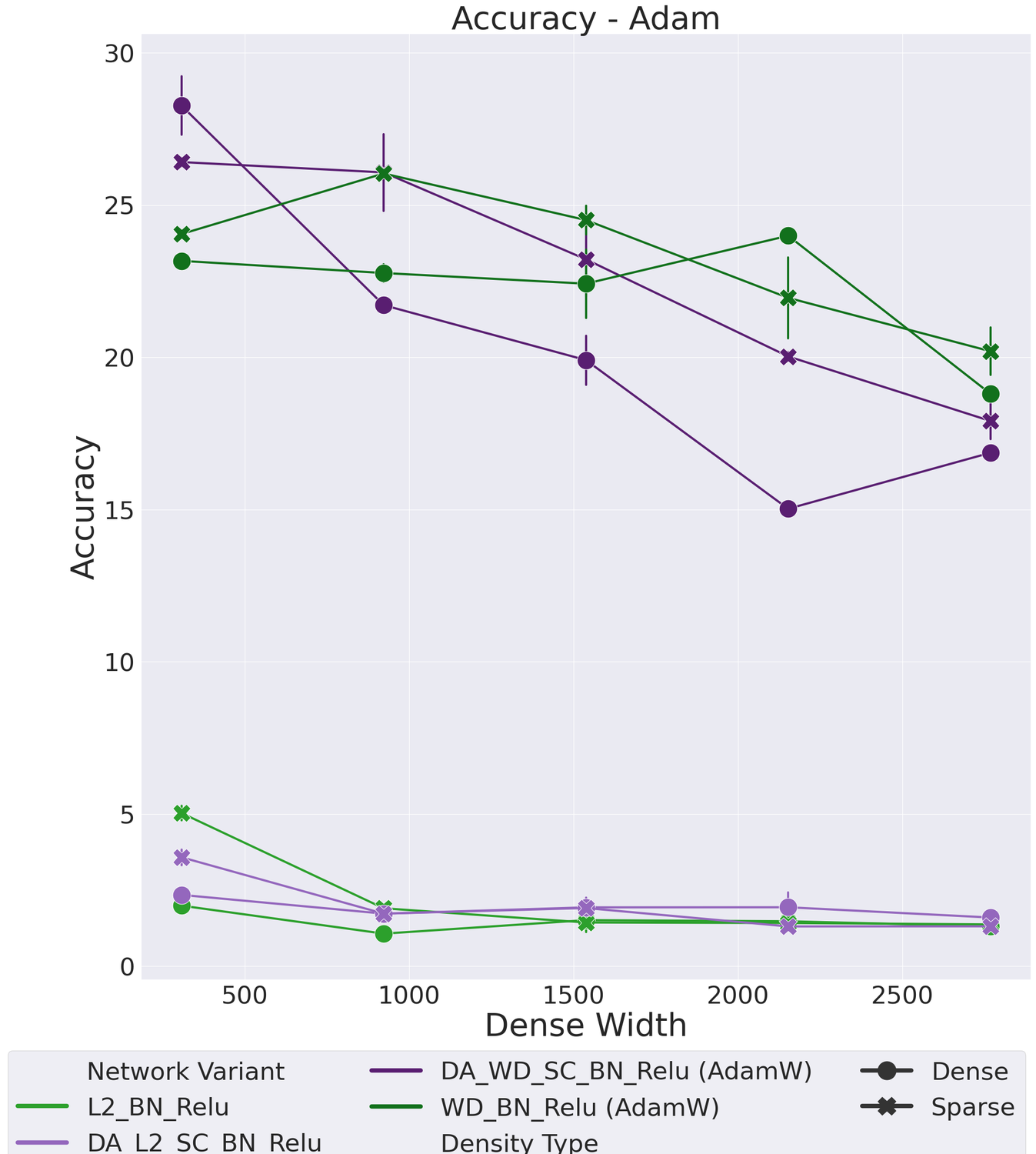}
    \caption{Test Accuracy for Dense and Sparse Networks on CIFAR-100}\label{fig:c100_adamw_acc}
  \end{subfigure}%
  \begin{subfigure}{0.45\linewidth}
    \centering
    \includegraphics[width=.8\linewidth]{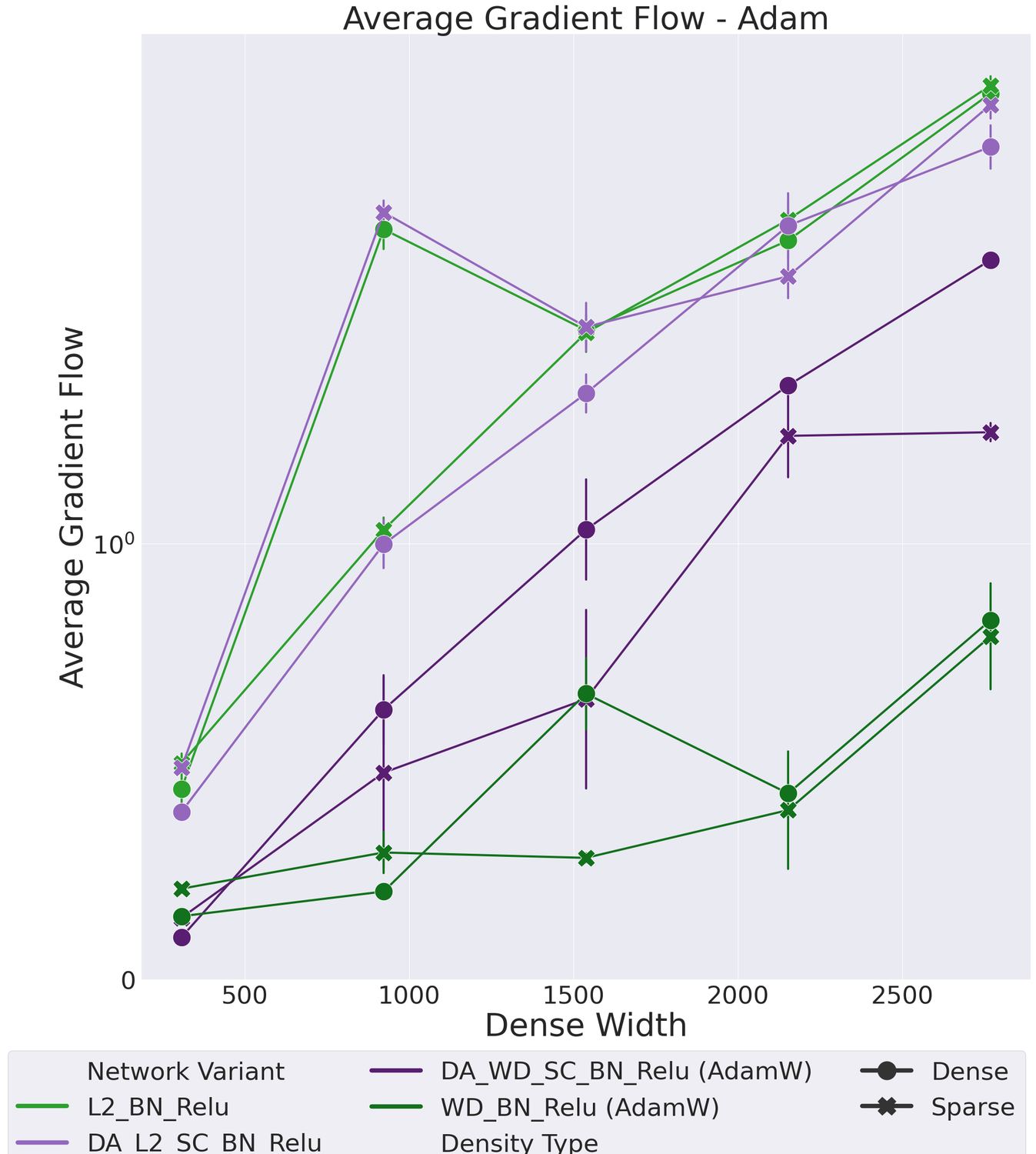}
    \caption{Gradient Flow for Dense and Sparse Networks on CIFAR-100}\label{fig:c100_adamw_acc_grad_flow}
  \end{subfigure}
\end{figure}

\begin{figure}[ht]
  \begin{center}
    \caption{\textbf{Wide ResNet-50 Gradient Flow on CIFAR-100.} We illustrate the gradient flow of Wide ResNet-50 on CIFAR-100, with the density ranging from 1\% to 100\%. The accuracy results can be found in Figure \ref{fig:wres_acc}.} \label{fig:wres_grad_flow}
    \includegraphics[width=0.9\linewidth]{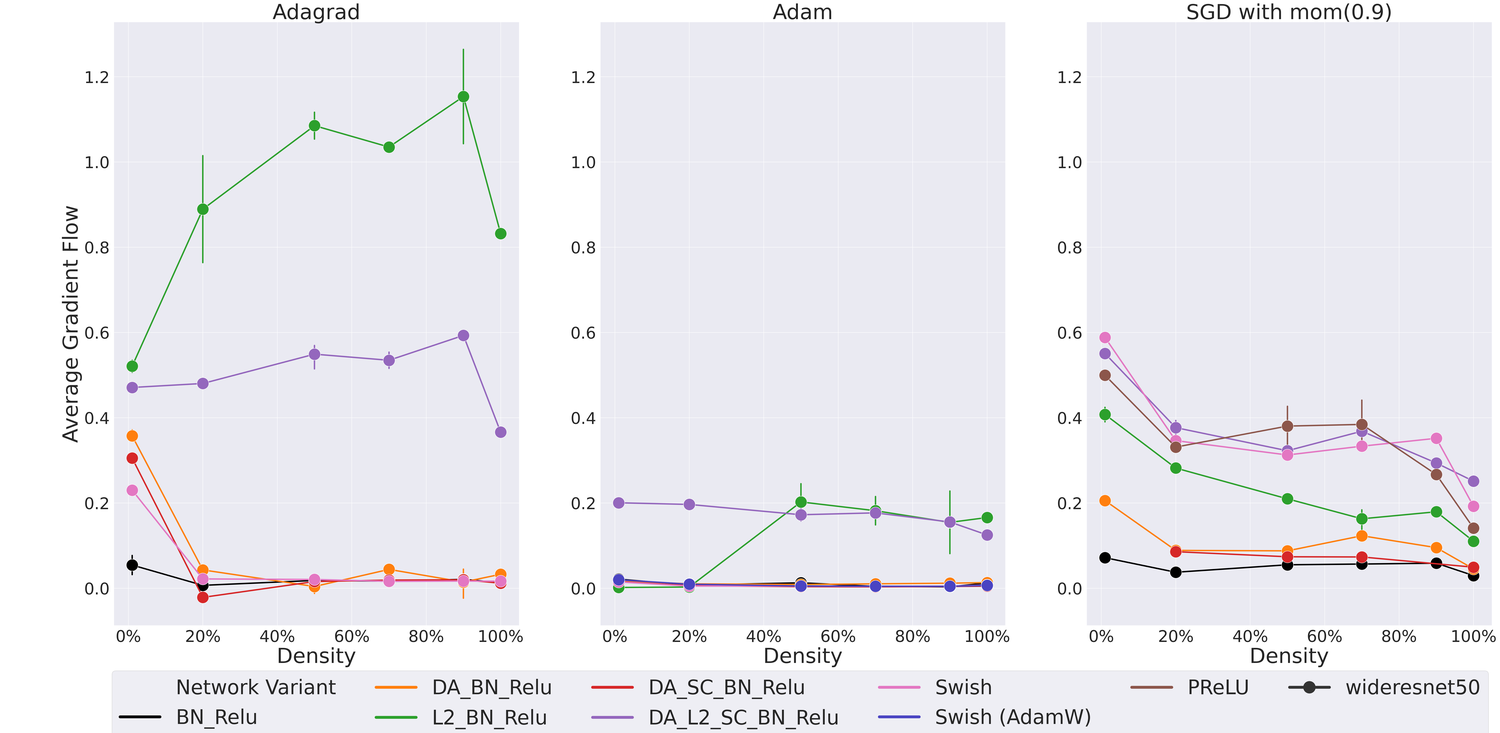}
  \end{center}
\end{figure}

We present the detailed results as follows:
\begin{enumerate}
  \item FMNIST
        \begin{enumerate}
          \item All \texttt{SC-SDC} results for Four Hidden Layers with different regularization methods - Table \ref{tbl:fmnist_reg_wilcoxon}.
          \item Effects of Regularization with a low learning rate (0.001)  - Figure \ref{fig:fmnist_diff_reg_all_optims_low_lr}.
        \end{enumerate}
  \item CIFAR-10
        \begin{enumerate}
          \item All \texttt{SC-SDC} results for Four Hidden Layers with different regularization methods - Table \ref{tbl:c10_reg_wilcoxon}.
          \item Effects of Regularization with a low learning rate (0.001) - Figure \ref{fig:c10_diff_reg_all_optims_low_lr}.
          \item Effects of Regularization with a high learning rate (0.1) - Figure \ref{fig:c10_diff_reg_all_optims_high_lr}.
          \item Effects of Activations with a low learning rate (0.001) - Figure \ref{fig:c10_diff_acts_all_optims_low}.
        \end{enumerate}

  \item CIFAR-100
        \begin{enumerate}
          \item Detailed Results with a low learning rate (0.001)
                \begin{enumerate}
                  \item One, Two and Four hidden layers showing all forms of regularization - accuracy and gradient flow (Figure \ref{fig:c100_diff_reg_all_optims_low_lr_all}).
                  \item Four Hidden Layers with different activation functions - accuracy and gradient flow (Figure \ref{fig:c100_diff_reg_all_optims_low_lr_acts}).
                  \item All \texttt{SC-SDC} results for Four Hidden Layers with different regularization and activation methods- Table \ref{tbl:wilcoxon_low_lr_all}.
                \end{enumerate}
          \item   Detailed Results with a high learning rate (0.1)
                \begin{enumerate}
                  \item Four Hidden Layers with BatchNorm - accuracy and gradient flow (Figure \ref{fig:c100_diff_reg_all_optims_high_lr_with_batchnorm}).

                  \item Four Hidden Layers with different activation functions - accuracy and gradient flow (Figure \ref{fig:c100_diff_reg_all_optims_high_lr_acts}).
                  \item Adam vs AdamW - accuracy and gradient flow (Figure \ref{fig:c100_adamw}).
                  \item Gradient Flow for Wide ResNet-50 - Figure \ref{fig:wres_grad_flow}.
                \end{enumerate}
        \end{enumerate}
\end{enumerate}

\begin{table}[ht]
  \small
  \begin{center}
    \resizebox{0.95\textwidth}{!}{%
      \begin{tabular}{lrrrrrrr}
\toprule
          & \multicolumn{1}{l}{NR\_Relu}  & \multicolumn{1}{l}{L2\_Relu}  & \multicolumn{1}{l}{DA\_Relu}  & \multicolumn{1}{l}{SC\_Relu}           & \multicolumn{1}{l}{BN\_Relu}           & \multicolumn{1}{l}{DA\_SC\_BN\_Relu}   & \multicolumn{1}{l}{DA\_L2\_SC\_BN\_Relu} \\
\midrule
Adagrad   & \cellcolor[HTML]{E77E72}0.993 & \cellcolor[HTML]{E77E72}0.992 & \cellcolor[HTML]{E67C73}1.000 & \cellcolor[HTML]{E98771}0.941          & \cellcolor[HTML]{57BB8A}\textbf{0.001} & \cellcolor[HTML]{57BB8A}\textbf{0.001} & \cellcolor[HTML]{57BB8A}\textbf{0.000} \\
Adam      & \cellcolor[HTML]{90C47E}0.170 & \cellcolor[HTML]{E77D72}0.995 & \cellcolor[HTML]{74BF84}0.088 & \cellcolor[HTML]{5BBB8A}\textbf{0.012} & \cellcolor[HTML]{5DBC89}\textbf{0.020} & \cellcolor[HTML]{57BB8A}\textbf{0.000} & \cellcolor[HTML]{70BF85}0.076          \\
Mom (0.9) & \cellcolor[HTML]{E77D72}1.000 & \cellcolor[HTML]{E77E72}0.991 & \cellcolor[HTML]{E67C73}1.000 & \cellcolor[HTML]{EA8A71}0.924          & \cellcolor[HTML]{57BB8A}\textbf{0.000} & \cellcolor[HTML]{59BB8A}\textbf{0.008} & \cellcolor[HTML]{59BB8A}\textbf{0.008} \\
\bottomrule
\end{tabular}
    }
  \end{center}
  \caption{\textbf{Wilcoxon Signed Rank Test Results for MLPs With Four Hidden Layers, With a Low Learning Rate (0.001), trained on FMNIST.}}
  \label{tbl:fmnist_reg_wilcoxon}
\end{table}

\begin{table}[ht]
  \small
  \begin{subtable}{0.45\textwidth}\centering
    \caption{Different Regularization Methods - Low learning rate (0.001)}
    \begin{tabular}{lrrrrr}
\toprule
          & \multicolumn{1}{l}{NR}                 & \multicolumn{1}{l}{DA}                 & \multicolumn{1}{l}{L2}        & \multicolumn{1}{l}{SC}                 & \multicolumn{1}{l}{BN}                 \\
\midrule
Adagrad   & \cellcolor[HTML]{FDCD67}0.555          & \cellcolor[HTML]{F6B66A}0.681          & \cellcolor[HTML]{E77D72}0.999 & \cellcolor[HTML]{B2C977}0.271          & \cellcolor[HTML]{57BB8A}\textbf{0.000} \\
Adam      & \cellcolor[HTML]{5FBC89}\textbf{0.024} & \cellcolor[HTML]{57BB8A}\textbf{0.001} & \cellcolor[HTML]{F6B66A}0.681 & \cellcolor[HTML]{57BB8A}\textbf{0.000} & \cellcolor[HTML]{57BB8A}\textbf{0.000} \\
RMSProp   & \cellcolor[HTML]{57BB8A}\textbf{0.000} & \cellcolor[HTML]{FBC568}0.598          & \cellcolor[HTML]{ED956F}0.866 & \cellcolor[HTML]{57BB8A}\textbf{0.001} & \cellcolor[HTML]{58BB8A}\textbf{0.005} \\
SGD       & \cellcolor[HTML]{E67C73}1.000          & \cellcolor[HTML]{E77D72}1.000          & \cellcolor[HTML]{E67C73}1.000 & \cellcolor[HTML]{57BB8A}\textbf{0.000} & \cellcolor[HTML]{57BB8A}\textbf{0.000} \\
Mom (0.9) & \cellcolor[HTML]{E67C73}1.000          & \cellcolor[HTML]{ED956F}0.862          & \cellcolor[HTML]{E67C73}1.000 & \cellcolor[HTML]{E77D72}0.995          & \cellcolor[HTML]{57BB8A}\textbf{0.001} \\ 
\bottomrule
\end{tabular}
    \label{tbl:c10_reg_low_lr}
  \end{subtable}
  \begin{subtable}{0.45\textwidth}\centering
    \caption{Different Regularization Methods - High learning rate (0.1)}
    \begin{tabular}{lrrrr}
\toprule
          & \multicolumn{1}{l}{BN}                 & \multicolumn{1}{l}{DA\_BN}             & \multicolumn{1}{l}{L2\_BN}             & \multicolumn{1}{l}{SC\_BN}             \\
\midrule
Adagrad   & \cellcolor[HTML]{57BB8A}\textbf{0.001} & \cellcolor[HTML]{6CBE86}0.063          & \cellcolor[HTML]{E67C73}0.953          & \cellcolor[HTML]{57BB8A}\textbf{0.000} \\
Adam      & \cellcolor[HTML]{5ABB8A}\textbf{0.009} & \cellcolor[HTML]{57BB8A}\textbf{0.001} & \cellcolor[HTML]{6FBF85}0.074          & \cellcolor[HTML]{57BB8A}\textbf{0.000} \\
RMSProp   & \cellcolor[HTML]{A3C77A}0.227          & \cellcolor[HTML]{D5CF6F}0.375          & \cellcolor[HTML]{A0C67B}0.219          & \cellcolor[HTML]{63BC88}\textbf{0.037} \\
SGD       & \cellcolor[HTML]{57BB8A}\textbf{0.000} & \cellcolor[HTML]{FFD666}0.500          & \cellcolor[HTML]{B8CA76}0.289          & \cellcolor[HTML]{57BB8A}\textbf{0.000} \\
Mom (0.9) & \cellcolor[HTML]{57BB8A}\textbf{0.000} & \cellcolor[HTML]{5FBC89}\textbf{0.024} & \cellcolor[HTML]{58BB8A}\textbf{0.004} & \cellcolor[HTML]{57BB8A}\textbf{0.000} \\
\bottomrule
\end{tabular}

    \label{tbl:c10_reg_high_lr}
  \end{subtable}
  \begin{subtable}{\textwidth}\centering
    \begin{center}
      \caption{Effect of Different Activation Functions - High learning rate (0.001)}
      \begin{tabular}{lrrrr}
\toprule
          & \multicolumn{1}{l}{ReLU}               & \multicolumn{1}{l}{SReLU}              & \multicolumn{1}{l}{Swish}              & \multicolumn{1}{l}{PReLU}              \\
\midrule
Adam      & \cellcolor[HTML]{63BC88}\textbf{0.004} & \cellcolor[HTML]{6ABE86}\textbf{0.006} & \cellcolor[HTML]{5BBB89}\textbf{0.002} & \cellcolor[HTML]{66BD87}\textbf{0.005} \\
Mom (0.9) & \cellcolor[HTML]{57BB8A}\textbf{0.000} & \cellcolor[HTML]{FFD666}0.053          & \cellcolor[HTML]{5BBB89}\textbf{0.002} & \cellcolor[HTML]{58BB8A}\textbf{0.001} \\ 
\bottomrule
\end{tabular}
      \label{tbl:activation_c10}
    \end{center}
  \end{subtable}
  \vspace{1ex}
  \caption{\textbf{Wilcoxon Signed Rank Test Results for MLPs with Four Hidden Layers, trained on CIFAR-10.} Wilcoxon Signed Rank Test Results for CIFAR-10, using a $p$-value of 0.05, with the bold values indicating where we can be statistically confident that sparse networks perform better than dense (reject $H_0$ from \ref{txt:wil_test}). We also use a continuous colour scale to make the results more interpretable. This scale ranges from green (0 - likely that sparse networks perform better than dense) to yellow (0.5 - 50\% chance that sparse networks perform better than dense) to red (1 - highly likely that sparse networks do not outperform dense - cannot reject $H_0$ from \ref{txt:wil_test}).}
  \label{tbl:c10_reg_wilcoxon}
\end{table}

\begin{table}[ht]
  \small
  \begin{subtable}{\linewidth}\centering
    \begin{center}
      \caption{Effect of Different Regularization Methods}
      \resizebox{0.95\linewidth}{!}{%
        \begin{tabular}{lrrrrrrrrrr}
\toprule
          & \multicolumn{1}{l}{NR}                 & \multicolumn{1}{l}{DA}                 & \multicolumn{1}{l}{L2}        & \multicolumn{1}{l}{SC}                 & \multicolumn{1}{l}{BN}                 & \multicolumn{1}{l}{DA\_BN}             & \multicolumn{1}{l}{L2\_BN}             & \multicolumn{1}{l}{SC\_BN}             & \multicolumn{1}{l}{DA\_SC\_BN}         & \multicolumn{1}{l}{DA\_L2\_SC\_BN}     \\
\midrule
Adagrad   & \cellcolor[HTML]{E67C73}1.000          & \cellcolor[HTML]{E67C73}1.000          & \cellcolor[HTML]{E77D72}0.998 & \cellcolor[HTML]{A7C779}0.239          & \cellcolor[HTML]{58BB8A}\textbf{0.006} & \cellcolor[HTML]{57BB8A}\textbf{0.002} & \cellcolor[HTML]{57BB8A}\textbf{0.001} & \cellcolor[HTML]{57BB8A}\textbf{0.003} & \cellcolor[HTML]{57BB8A}\textbf{0.001} & \cellcolor[HTML]{58BB8A}\textbf{0.004} \\
Adam      & \cellcolor[HTML]{57BB8A}\textbf{0.000} & \cellcolor[HTML]{69BD87}0.055          & \cellcolor[HTML]{99C57C}0.198 & \cellcolor[HTML]{57BB8A}\textbf{0.003} & \cellcolor[HTML]{71BF85}0.079          & \cellcolor[HTML]{68BD87}0.051          & \cellcolor[HTML]{ACC878}0.254          & \cellcolor[HTML]{8EC37F}0.166          & \cellcolor[HTML]{64BD88}\textbf{0.039} & \cellcolor[HTML]{7EC182}0.118          \\
RMSProp   & \cellcolor[HTML]{57BB8A}\textbf{0.001} & \cellcolor[HTML]{57BB8A}\textbf{0.000} & \cellcolor[HTML]{BBCB75}0.300 & \cellcolor[HTML]{8EC37F}0.166          & \cellcolor[HTML]{7EC182}0.117          & \cellcolor[HTML]{5DBC89}\textbf{0.021} & \cellcolor[HTML]{EB8C70}0.914          & \cellcolor[HTML]{FDCF67}0.541          & \cellcolor[HTML]{77C084}0.096          & \cellcolor[HTML]{5EBC89}\textbf{0.023} \\
SGD       & \cellcolor[HTML]{E67C73}1.000          & \cellcolor[HTML]{E67C73}1.000          & \cellcolor[HTML]{E67C73}1.000 & \cellcolor[HTML]{AAC879}0.248          & \cellcolor[HTML]{57BB8A}\textbf{0.000} & \cellcolor[HTML]{57BB8A}\textbf{0.000} & \cellcolor[HTML]{57BB8A}\textbf{0.001} & \cellcolor[HTML]{57BB8A}\textbf{0.003} & \cellcolor[HTML]{57BB8A}\textbf{0.001} & \cellcolor[HTML]{58BB8A}\textbf{0.004} \\
Mom (0.9) & \cellcolor[HTML]{E67C73}1.000          & \cellcolor[HTML]{E67C73}1.000          & \cellcolor[HTML]{E77D72}1.000 & \cellcolor[HTML]{E77D72}0.999          & \cellcolor[HTML]{57BB8A}\textbf{0.001} & \cellcolor[HTML]{57BB8A}\textbf{0.000} & \cellcolor[HTML]{59BB8A}\textbf{0.007} & \cellcolor[HTML]{59BB8A}\textbf{0.008} & \cellcolor[HTML]{57BB8A}\textbf{0.001} & \cellcolor[HTML]{57BB8A}\textbf{0.003} \\
\bottomrule
\end{tabular}

      }
      \label{tbl:reg_c100}
    \end{center}
    \begin{center}
      \caption{Effect of Activation Functions}
      \begin{tabular}{lrrrrrr}
\toprule
          & \multicolumn{1}{c}{relu}               & \multicolumn{1}{c}{swish}              & \multicolumn{1}{c}{srelu}     & \multicolumn{1}{c}{prelu}              & \multicolumn{1}{c}{elu}                & \multicolumn{1}{c}{sigmoid}            \\
\midrule
Adagrad   & \cellcolor[HTML]{57BB8A}\textbf{0.000} & \cellcolor[HTML]{57BB8A}\textbf{0.001} & \cellcolor[HTML]{79C083}0.104 & \cellcolor[HTML]{57BB8A}\textbf{0.001} & \cellcolor[HTML]{57BB8A}\textbf{0.000} & \cellcolor[HTML]{5ABB8A}\textbf{0.011} \\
Adam      & \cellcolor[HTML]{64BD88}\textbf{0.042} & \cellcolor[HTML]{57BB8A}\textbf{0.003} & \cellcolor[HTML]{70BF85}0.076 & \cellcolor[HTML]{57BB8A}\textbf{0.001} & \cellcolor[HTML]{58BB8A}\textbf{0.004} & \cellcolor[HTML]{57BB8A}\textbf{0.000} \\
RMSProp   & \cellcolor[HTML]{79C083}0.104          & \cellcolor[HTML]{8EC37F}0.165          & \cellcolor[HTML]{6CBE86}0.065 & \cellcolor[HTML]{66BD87}\textbf{0.047} & \cellcolor[HTML]{5BBB8A}\textbf{0.013} & \cellcolor[HTML]{57BB8A}\textbf{0.001} \\
SGD       & \cellcolor[HTML]{57BB8A}\textbf{0.002} & \cellcolor[HTML]{57BB8A}\textbf{0.000} & \cellcolor[HTML]{93C47E}0.180 & \cellcolor[HTML]{57BB8A}\textbf{0.000} & \cellcolor[HTML]{57BB8A}\textbf{0.001} & \cellcolor[HTML]{57BB8A}\textbf{0.000} \\
Mom (0.9) & \cellcolor[HTML]{57BB8A}\textbf{0.001} & \cellcolor[HTML]{57BB8A}\textbf{0.000} & \cellcolor[HTML]{68BD87}0.054 & \cellcolor[HTML]{57BB8A}\textbf{0.000} & \cellcolor[HTML]{57BB8A}\textbf{0.000} & \cellcolor[HTML]{68BD87}0.054      \\
\bottomrule
\end{tabular}
      \label{tbl:activation_c100_low_lr}
    \end{center}
  \end{subtable}
  \vspace{1ex}
  \caption{\textbf{Wilcoxon Signed Rank Test Results for MLPs with Four Hidden Layers, with low learning rate (0.001), trained on CIFAR-100.} Wilcoxon Signed Rank Test Results for CIFAR-100, using a low learning rate. We use a $p$-value of 0.05, with the bold values indicating where we can be statistically confident that sparse networks perform better than dense (reject $H_0$ from \ref{txt:wil_test}). We also use a continuous colour scale to make the results more interpretable. This scale ranges from green (0 - likely that sparse networks perform better than dense) to yellow (0.5 - 50\% chance that sparse networks perform better than dense) to red (1 - highly likely that sparse networks do not outperform dense - cannot reject $H_0$ from \ref{txt:wil_test}).  }
  \label{tbl:wilcoxon_low_lr_all}
\end{table}

\end{document}